\documentclass[final,5p,times,twocolumn]{elsarticle}

\usepackage{amssymb}
\usepackage{times} 
\usepackage{amsfonts}
\usepackage{amsthm}
\usepackage{graphics} 
\usepackage{epsfig} 
\usepackage{amsmath} 
\usepackage{subfigure}
\usepackage{booktabs}
\usepackage{multirow}
\usepackage{makecell}
\usepackage{mathtools}
\usepackage{graphicx}
\usepackage{lscape}
\usepackage{multicol} 
\usepackage{graphicx, epstopdf}
\usepackage[table,xcdraw]{xcolor}
\usepackage{caption}
\usepackage{xspace,epsfig,url}
\usepackage{enumerate}
\usepackage{float}
\usepackage{epsf}
\usepackage{psfrag}
\usepackage{verbatim}
\usepackage[all]{xy}
\usepackage{color,soul}
\usepackage{bm}
\usepackage{comment}
\usepackage{cases}
\usepackage{hyperref}
\usepackage{physics}
\usepackage{amssymb}
\usepackage{pifont}
\usepackage{array} 
\usepackage[ruled,vlined,linesnumbered]{algorithm2e}
\usepackage{textcomp}
\usepackage{stfloats}
\usepackage{caption}
\usepackage{nicefrac, xfrac}
\newcommand{\expnumber}[2]{{#1}\mathrm{E}{#2}}

\journal{ArXiv Preprint}

\begin{document}

\begin{frontmatter}



\title{Multiple Global Peaks Big Bang-Big Crunch Algorithm for Multimodal Optimization}

\author[inst1]{Fabio Stroppa}
\author[inst1]{Ahmet Astar}

\affiliation[inst1]{organization={Computer Engineering Department, Kadir Has University},
            addressline={Cibali, Kadir Has Cd., Fatih}, 
            city={Istanbul},
            postcode={34083}, 
            country={Turkey}}

\begin{abstract}
The main challenge of multimodal optimization problems is identifying multiple peaks with high accuracy in multidimensional search spaces with irregular landscapes. 
This work proposes the Multiple Global Peaks Big Bang-Big Crunch (MGP-BBBC) algorithm, which addresses the challenge of multimodal optimization problems by introducing a specialized mechanism for each operator. The algorithm expands the Big Bang-Big Crunch algorithm, a state-of-the-art metaheuristic inspired by the universe's evolution. Specifically, MGP-BBBC groups the best individuals of the population into cluster-based centers of mass and then expands them with a progressively lower disturbance to guarantee convergence. During this process, it (i) applies a distance-based filtering to remove unnecessary elites such that the ones on smaller peaks are not lost, (ii) promotes isolated individuals based on their niche count after clustering, and (iii) balances exploration and exploitation during offspring generation to target specific accuracy levels.
Experimental results on twenty multimodal benchmark test functions show that MGP-BBBC generally performs better or competitively with respect to other state-of-the-art multimodal optimizers.
\end{abstract}


\end{frontmatter}



\section{Introduction}
\label{sec:intro}


Multimodal optimization problems (MMOPs) are characterized by having multiple optimal solutions~\cite{preuss2015multimodal}. Each optimum in MMOPs is considered a peak, and the solvers must identify multiple peaks simultaneously with an appropriate level of accuracy. This scenario is frequently encountered in real-world applications~\cite{zaman2017evolutionary,vidanalage2018multimodal}, in which relying on a single optimum is often insufficient due to several factors. A global optimum might be unrealistic or excessively costly~\cite{cuevas2014cuckoo}. Solutions may need to adapt as resources or conditions change over time~\cite{deb2012multimodal}. The availability of multiple optimal solutions can provide valuable insights into the problem’s characteristics~\cite{deb2008innovization}. Additionally, robust solutions are preferred over global ones because they are less sensitive to small changes, and uncertainties~\cite{nomaguchi2016robust}, and reliable solutions are favored to avoid infeasible outcomes~\cite{dizangian2015reliability}. Consequently, a local optimum with reasonable performance and cost may be more desirable than a marginally better but costly global optimum~\cite{wong2012evolutionary}.

Evolutionary Algorithms (EAs) effectively address complex multimodal optimization problems thanks to their stochastic search and parallel processing to find multiple good solutions simultaneously~\cite{goldberg1989genetic,holland1975adaptation}. Various multimodal evolutionary algorithms (MMEAs) utilize strategies such as: 
promoting diversity via spatial segregation or distribution~\cite{izzo2012generalized,gordon2004visualization}; 
partitioning populations and restricting mating~\cite{kashtiban2016solving,thomsen2000religion}; 
using elitist approaches or conserving genotypes~\cite{liang2011genetic,li2002species}; 
enforcing diversity through fitness sharing~\cite{miller1996genetic}, clearing~\cite{petrowski1996clearing}, niching~\cite{li2016seeking}, crowding~\cite{thomsen2004multimodal}, or clustering~\cite{yenin2023multi,yin1993fast}; 
and applying multi-objective optimization techniques~\cite{deb2012multimodal}. 
However, these methods often face challenges such as tuning parameters, scalability issues, computational overhead, incomplete search space exploration, reliance on gradient information, premature convergence, and failure to balance exploration with exploitation~\cite{chen2009preserving,hong2022balancing}.

\begin{figure}[t!]
    \centering
    \includegraphics[width=8cm]{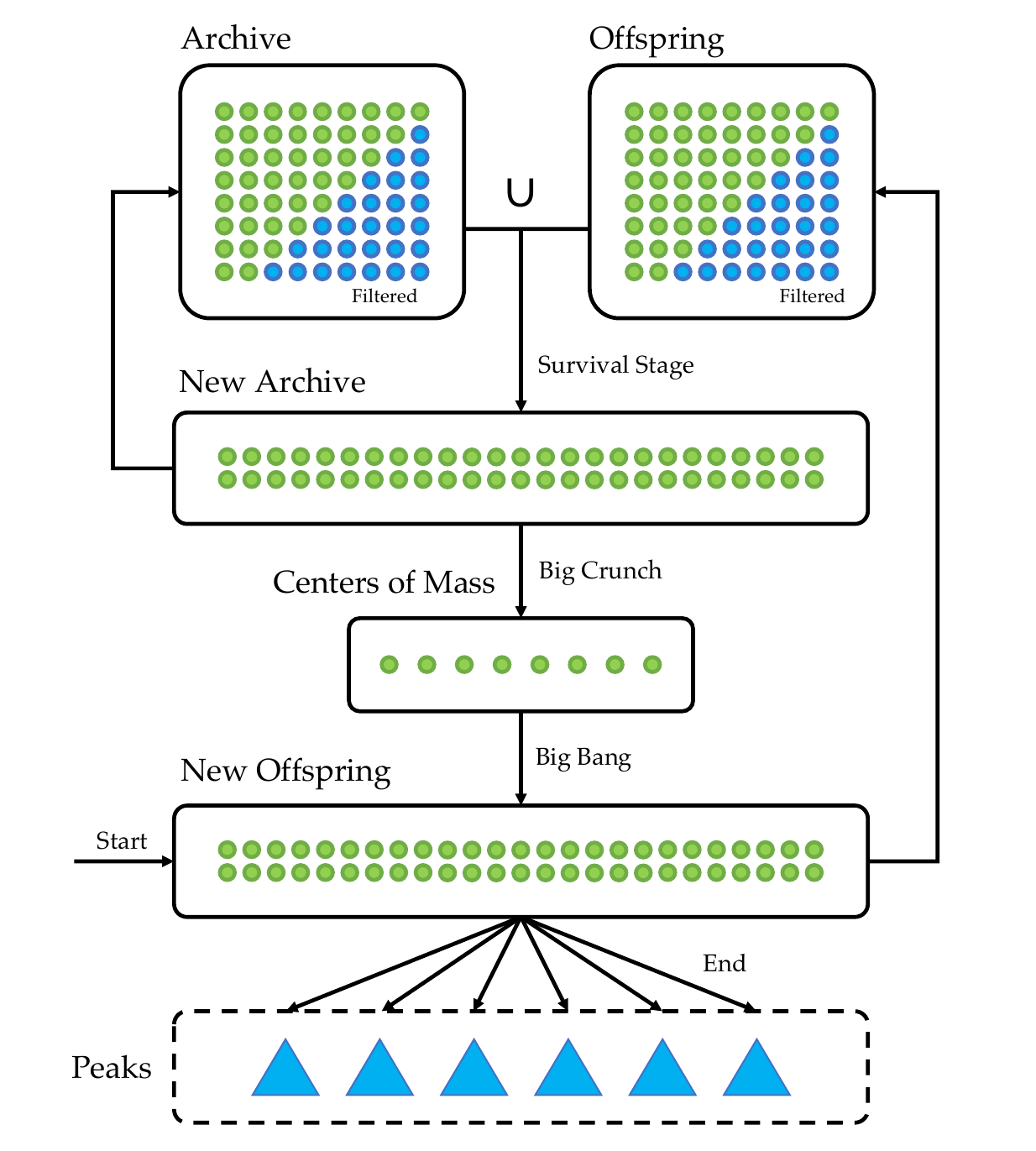}
    \caption{Basic framework of MGP-BBBC.}
    \label{fig:mgpbbbc_framework}
\end{figure}

The main challenge in MMOPs is dealing with search landscapes where peaks have varying sizes of concave regions -- i.e., the interval or span over which the function displays an upward/downward curvature around its peak. Large and wide peaks will have a higher probability of being surrounded by solutions, whereas small and sharp peaks might be easily missed. By promoting elites (i.e., individuals with the best fitness), wide peaks will be overpopulated, leaving no room for exploring unpromising areas that can, in fact, hide a sharp peak. This requires MMEAs to maintain a sufficient degree of diversity in the population. Methods such as niching are often used, which divide the population into several subpopulations around each peak. However, determining the optimal way to divide the population is not trivial. The state-of-the-art offers different efficient strategies, such as the distributed individuals for multiple peaks implemented in DIDE~\cite{chen2019distributed}, the dynamic niching implemented in SSGA~\cite{de2014dynamic}, niching with repelling subpopulations and taboo regions implemented in RS-CMSA~\cite{ahrari2017multimodal, ahrari2021static}, or hill-valley clustering implemented in HillVallEA~\cite{maree2018real, maree2019benchmarking}.

In this work, we propose a novel MMEA that adapts the Big Bang-Big Crunch algorithm (BBBC)\cite{erol2006new} to solve MMOPs. We selected this algorithm based on its significantly better performance in single global optimization with respect to other classic EAs~\cite{akbas2024impact}, and we propose to investigate its performances when extended to multimodal optimization.
Our algorithm, namely the Multiple Global Peaks BBBC algorithm (MGP-BBBC), is a cluster-based method and a niching-based extension of the k-Cluster BBBC algorithm (k-BBBC)\cite{yenin2023multi}. MGP-BBBC overcomes k-BBBC's limitation of knowing the number of optima sought by using a kernel-based clustering method and aims at retrieving only the global peaks rather than the local as well. Besides a termination condition (often based on a maximum number of generations $g$) and the MMOP to be solved $f$, MGP-BBBC only features two user-defined parameters: the size of the population $n$ and the clustering bandwidth $h$.

Fig.~\ref{fig:mgpbbbc_framework} shows the basic framework of MGP-BBBC, in which each component brings a different contribution to solve the challenge of MMOPs. MGP-BBBC proposes:
a \textit{survival stage} that allows isolated individuals with worse fitness to win the environmental selection, such that small and sharp peaks are not missed;
the \textit{big-crunch} operator that reduces the population into different centers of mass and assigns them a number of desired offspring based on the niche count of their respective cluster, to increase the selection pressure of isolated individuals; and
the \textit{big-bang} operator that produces offspring by dynamically turning exploration into exploitation in the second half of the generation loop, targeting specific high levels of accuracy.
We evaluated the performance of MGP-BBBC on the widely used CEC'2013 benchmark set containing twenty multimodal test functions. Compared with the state-of-the-art multimodal optimization algorithms, MGP-BBBC performs better than eleven over thirteen algorithms and performs competitively against the other two.

The rest of the paper is organized as follows: 
Sec.~\ref{sec:bbbc} describes the original BBBC algorithm; 
Sec.~\ref{sec:mgp_bbbc} describes the proposed MGP-BBBC algorithm; 
Sec.~\ref{sec:experiments} presents the comparison with the state-of-the-art algorithms; 
Sec.~\ref{sec:parameter_analysis} presents a parameter analysis and their effect on the performance; and finally,
Sec.~\ref{sec:conclusion} summarizes the work and presents possible future ideas.

\section{Big Bang-Big Crunch Algorithm} \label{sec:bbbc}

The Big Bang-Big Crunch (BBBC) algorithm was proposed by Erol et al.~\cite{erol2006new, gencc2010big} to address some of the biggest disadvantages of genetic algorithms: premature convergence, convergence speed, and execution time. Inspired by the evolution of the universe, it is composed of two phases: (i) explosion, or \textit{big bang}, in which the energy dissipation produces disorder and randomness, and (ii) implosion, or \textit{big crunch}, in which randomness is drawn back into a different order. BBBC creates an initial random population $P$ uniformly spread throughout the search space (bang), evaluates its individuals, and groups them into their center of mass (crunch). These two phases are repeated, generating new solutions progressively closer to the center of mass. Finally, the center of mass converges to the optimal solution of the problem. 

BBBC features different big-crunch operators~\cite{gencc2010big} to calculate the center of mass $c$. The simplest way is choosing the most fitting individual as in Eqn.~(\ref{eq:centerOfMass_best}). 


\begin{equation}
    \label{eq:centerOfMass_best}
    \begin{aligned}
        c = 
        \begin{dcases}
            \mathbf{x}_j : f(\mathbf{x}_j) = \max_{i=1}^{n}{f(\mathbf{x}_i)} \; \text{if max}\\
            \mathbf{x}_j : f(\mathbf{x}_j) = \min_{i=1}^{n}{f(\mathbf{x}_i)} \; \text{if min}
        \end{dcases}, \; \mathbf{x}_i \in P\\
    \end{aligned}
\end{equation}

The big-bang operator generates new individuals $\mathbf{x}^{new}$ for the next population as expressed in Eqn.~(\ref{eq:bang}). The center of mass $\mathbf{c}$ is disturbed with a random number $r$, which is multiplied by the search space upper bound $\mathbf{x}^{U}$ (to ensure the newly generated individual being within the search space) and divided by the current iteration step $i$ (to progressively reduce the \textit{extent of the expansion} in the search space and guaranteeing convergence). Note that the iteration step can also be multiplied by a further parameter to adjust the extent of expansion and quicken convergence~\cite{genc2013big} -- a property that MGP-BBBC uses for exploitation as discussed in Sec.~\ref{sec:big_bang}.

\begin{equation}
    \label{eq:bang}
    \begin{aligned}
        \mathbf{x}^{new} = \mathbf{c} + \frac{(\mathbf{x}^{U}-\mathbf{x}^{L}) \cdot r}{i}  , \quad \mathbf{x} \in P
    \end{aligned}
\end{equation}

\section{Multiple Global Peaks BBBC} \label{sec:mgp_bbbc}

Multiple Global Peaks BBBC (MGP-BBBC) is a multimodal extension of the BBBC algorithm (Sec.~\ref{alg:mgp_bbbc}). Similar to its predecessor k-BBBC~\cite{yenin2023multi}, MGP-BBBC identifies the multiple centers of mass by dividing the population into clusters. However, while k-BBBC relied on k-means~\cite{ahmed2020k} and therefore required a given-fixed number of clusters, MGP-BBBC is based on the mean-shift clustering~\cite{fukunaga1975estimation, cheng1995mean}, which automatically determines the number of clusters by identifying dense regions in the dataset space. Furthermore, unlike k-BBBC, MGP-BBBC amis at localizing only the global peaks, disregarding any local sub-optimal solution. 

\IncMargin{1em}
\begin{algorithm}
    \SetKwFunction{Init}{randomInitialization}
    \SetKwFunction{Eval}{evaluation}
    \SetKwFunction{Bang}{bigBang}
    \SetKwFunction{Crunch}{bigCrunch}
    \SetKwFunction{Sur}{survival}
    \SetKwInOut{Input}{input}
    \SetKwInOut{Output}{output}
    \Input{Population size $n$, number of generations $g$, clustering bandwidth $h$, optimization problem $f$}
    \Output{Archive of elites $A$}
    \Begin{	
        $O \gets \Init(n)$\;
            $A \gets \emptyset$\;
            $th \gets h$\;
            \For{$it \gets 1$  \textbf{to} $g$}
            {
                \If{$it \neq 1$}
                {
                    $O \gets \Bang(COM, it, g, OPC, f)$\;
                }
                $O \gets \Eval(O,f)$\;
                $A, th \gets \Sur(O,A, th)$\;
                $COM, OPC \gets \Crunch(A, h)$\;
            }
        \KwRet{$A$};
    }
    \caption{MGP-BBBC}\label{alg:mgp_bbbc}
\end{algorithm}\DecMargin{1em}

Algorithm~\ref{alg:mgp_bbbc} shows MGP-BBBC's framework.
The input parameters are the population size $n$, the maximum number of generations $g$ (which can also be replaced by a maximum number of function evaluations), the clustering bandwidth for the mean-shift clustering $h$ (defined in Sec.~\ref{sec:big_crunch}, which significantly affects the performance as detailed in Sec.~\ref{sec:parameter_analysis}), and the optimization problem $f$ (which includes the objective function, its dimensionality, search bounds, and constraints if any). 
The multiple global peaks are stored in an archive of elites $A$, separated from the rest of the population, which we will refer to as the population of offspring $O$.  it initializes $O$ with $n$ randomly generated individuals within the problem's bounds (line 2) and $A$ to an empty set (line 3); then, in the generational loop (lines 5-10), it evaluates $O$ (line 8, this procedure is reported in the supplementary material under Algorithm S.1), identifies elites to store in $A$ with a survival stage (line 9), crunches the elites to their centers of mass $C$ through clustering (line 10), and generates a new population from $C$ with a given number of offspring per each center of mass $OPC$ (line 7). The following subsection will describe each of these phases.

In MGP-BBBC, both $O$ and $A$ have the same size $n$. Each individual in the population is an object containing the values of the decision variables (attribute \texttt{.x}), the fitness (attribute \texttt{.fit}), and an additional boolean tag for removal from the population during filtering (attribute \texttt{.tag}, reported only in the supplementary material under Algorithm S.3).

\IncMargin{1em}
\begin{algorithm} [t!]
    \SetKwFunction{Dist}{pairwiseDistances}
    \SetKwFunction{Trunc}{filtering}
    \SetKwFunction{Sort}{sort}
    \SetKwFunction{Sortrows}{sortrows}
    \SetKwRepeat{Do}{do}{while}
    \SetKwInOut{Input}{input}
    \SetKwInOut{Output}{output}
    \Input{Offspring population $O$, archive of elites $A$, filtering threshold $th$}
    \Output{New archive of elites $A$, updated filtering threshold $th$}
    \Begin
    {	
        $n \gets $ size of $ O $\;
        \If{$A \neq \emptyset$}
        {
            $D_A \gets \Dist(A)$\;
            $D_O \gets \Dist(O)$\;
            \Do
            {$n' < n$} 
            {
                $A' \gets \Trunc(A,D_A,th)$\;
                $O' \gets \Trunc(O,D_O,th)$\;
                $A \gets O' \cup A' $\;
                $n' \gets $ size of $ A $\;
                \If{$n' < n$}
                {
                    $th \gets th \cdot 0.9 $\;
                }
            }
            $\Sort(A)$ by fitness\;
            resize $A$ by keeping the first $n$ individuals\;
        }
        \Else
        {
            $A \gets O$\;
        }
        \KwRet{$A,th$};
    }
    \caption{Survival Stage}\label{alg:survival}
\end{algorithm}\DecMargin{1em}
\IncMargin{1em}
\begin{algorithm}
    \SetKwFunction{Dist}{pairwiseDistances}
    \SetKwFunction{Trunc}{filtering}
    \SetKwFunction{Sort}{sort}
    \SetKwFunction{Sortrows}{sortrows}
    \SetKwRepeat{Do}{do}{while}
    \SetKwInOut{Input}{input}
    \SetKwInOut{Output}{output}
    \Input{Population $P$}
    \Output{Array of pairwise distances $D$ (having $\frac{n-1}{2}$ rows, each of them storing (1) pairwise distance, (2) index of first individual in $P$, and (3) index of second individual in $P$ }
    \Begin
    {	
        $n \gets $ size of $ P $\;
        $n \gets n \cdot \frac{n-1}{2} $\;        
        $D \gets \emptyset$\; 
        \For{$k \gets 1$ \textbf{to} $n$}
        {
            \For{$j \gets i+1$ \textbf{to} $|n|$}
            {
                $d.\texttt{dist} \gets \norm{P_i,P_j}^2$\; 
                $d.\texttt{ind1} \gets i$\; 
                $d.\texttt{ind2} \gets j$\; 
                $D \gets D \cup \{d\} $\;
            }
        }
        \KwRet{$D$};
    }
    \caption{Pairwise Distances}\label{alg:pairwise_distances}
\end{algorithm}\DecMargin{1em}
\IncMargin{1em}
\begin{algorithm}
    \SetKwFunction{Dist}{pairwiseDistances}
    \SetKwFunction{Trunc}{filtering}
    \SetKwFunction{Sort}{sort}
    \SetKwRepeat{Do}{do}{while}
    \SetKwInOut{Input}{input}
    \SetKwInOut{Output}{output}
    \Input{Population $P$, array of pairwise distances $D$, proximity threshold $th$}
    \Output{Filtered population $P'$}
    \Begin
    {	
        $P' \gets \emptyset$\;
        \For{$i \gets 1$ \textbf{to} $|P|$}
        {
            $P_i.\texttt{tag} \gets 0 $\;
        }
        \For{$k \gets 1$ \textbf{to} $|D|$}
        {
            \If{$D_k.\texttt{dist} < th$}
            {
                $i \gets D_k.\texttt{ind1} $\;
                $j \gets D_k.\texttt{ind2} $\;
                \If{$P_i.\texttt{tag} = 0 \wedge P_j.\texttt{tag} = 0$}
                {
                    /* for maximization problems */ \newline
                    \If{$P_i.\texttt{fit} \geq P_j.\texttt{fit}$}
                    {
                        $P_j.\texttt{tag} \gets 1 $\;
                    } 
                    \Else
                    {
                        $P_i.\texttt{tag} \gets 1 $\;
                    }
                }
            }
        }
        \For{$i \gets 1$ \textbf{to} $|P|$}
        {
            \If{$P_i.\texttt{tag} = 0$}
            {
                $P' \gets P' \cup \{P_i\} $\;
            }
            
        }
        \KwRet{$P'$};
    }
    \caption{Filtering Operation}\label{alg:filtering}
\end{algorithm}\DecMargin{1em}

\subsection{Survival Stage} \label{sec:survival}

\begin{figure}[t!]
    \centering
    \subfigure[\protect\url{}\label{fig:vincent_fun}Vincent with 2D]
    {\includegraphics[height=4.5cm]{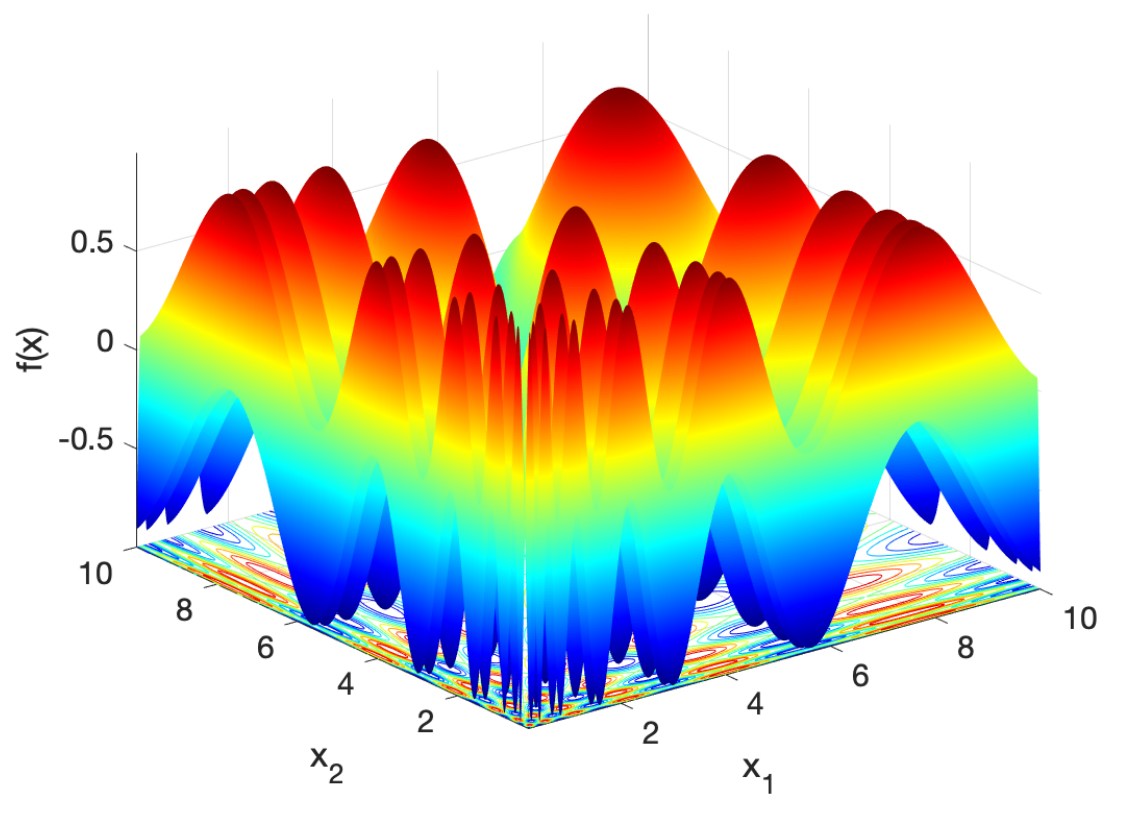}}
    \subfigure[\protect\url{}\label{fig:vincent_elites}Population Filtering]
    {\includegraphics[height=4.5cm]{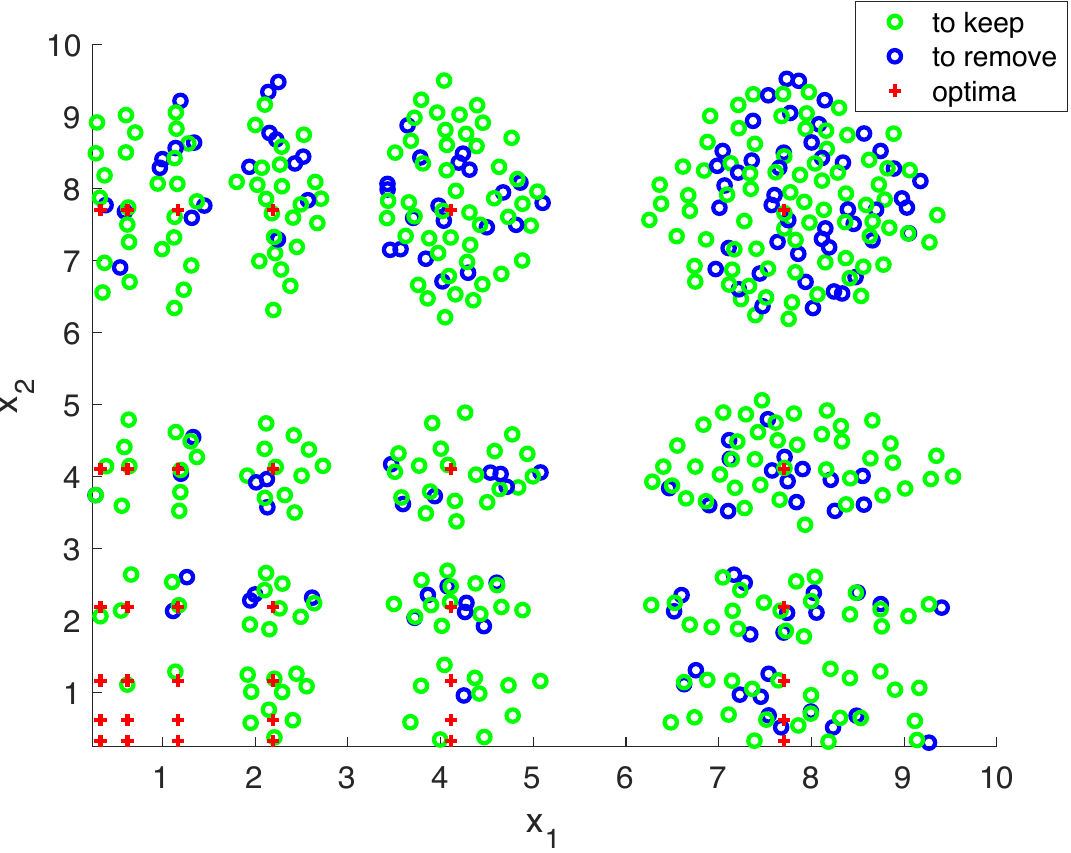}}
    \caption{Distribution of elites (green and blue circles) around the global peaks (red crosses) for Vincent with 2D function ($F_7$ of CEC'2013 benchmark set). The plot shows that more elites tend to surround the prominent peaks (up-right area) due to the higher density of high fitness points rather than the smaller sharper ones (down-left area) -- and many peaks in the down-left area are missed. The filtering procedure will remove points close to each other (blue circles), allowing MGP-BBBC to explore surroundings in low-density areas when producing new offspring.
    }
    \label{fig:filtering}
\end{figure}

Algorithm~\ref{alg:survival} shows the survival stage, which identifies the elites in the population and stores them in an archive $A$ having the same size as the population $O$. An elite is defined as an individual with high fitness, and MGP-BBBC filters out non-elites with a $\mu + \lambda$ schema~\cite{beyer2002evolution} (lines 14-15). However, by doing so, two suboptimal individuals close to the same peak might filter out a third isolated suboptimal individual with worse fitness but close to another peak -- therefore, missing that peak. This is even more exacerbated in problems featuring peaks with different sizes of concave regions, such as the Vincent function shown in Fig.~\ref{fig:vincent_fun}. It is more likely that a uniform exploration of the search space would generate elites around the large peak rather than the small one. 

Therefore, before non-elite removal, MGP-BBBC applies a distance-based filtering operation: it calculates all the pairwise distances in the population $A$ and $O$ (lines 4-5), and then applies the filtering procedure shown in Fig.~\ref{fig:vincent_elites} on both populations separately (lines 7-8). For each pair of individuals, this operation removes the one with worse fitness if their distance is smaller than a threshold $th$, allowing far away individuals to dominate over individuals in high-density regions. The pseudocode of the pairwise distances and the filtering operation are reported in Algorithm~\ref{alg:pairwise_distances} and \ref{alg:filtering}, respectively. 
The new archive of elites $A$ is then composed of the union of the filtered populations (line 9). 

However, the size of the new archive $A$ must be greater or equal to the desired population size $n$ for the $\mu + \lambda$ schema to work. This might not always be the case, depending on the extent of filtering performed: the larger the threshold $th$, the more individuals will be filtered out. Therefore, MGP-BBBC dynamically tunes the value of the threshold starting from the same value as the clustering bandwidth $h$ (line 4 of Algorithm~\ref{alg:mgp_bbbc}) and decreasing it by a small amount (i.e., by $10\%$) until the size of the archive is greater or equal to $n$ (lines 6-13). As the generations increase and the population converges to the peaks, the value of $th$ will also decrease.

Note that the first time MGP-BBBC calls the survival stage, $A$ will be empty (line 3) and, therefore, assigned directly to be filled with the individuals in the current population $O$ (lines 16-17). Furthermore, sorting (line 14) is based on the individuals' fitness, and it is in descending or ascending order based on maximization or minimization problems, respectively.

With the time complexity of the pairwise distance calculation and filtering operation being both $\mathcal{O}(n^2)$, the time complexity of the survival stage of MGP-BBBC is $\mathcal{O}(\delta_{th} \cdot n^2 + n\log{n})$ -- where the term $\delta_{th}$ indicates the number of decrements of $th$, which at most depends on its data type resolution until data underflow. If data underflow occurs, then the remainder in $A$ is filled with old elites randomly picked\footnote{This procedure is not reported in the pseudocode for sake of clarity.}.

\subsection{Big Crunch} \label{sec:big_crunch}

 Algorithm~\ref{alg:big_crunch} shows the big-crunch operator of MGP-BBBC, which identifies the centers of mass $COM$ within the archive $A$. This operation relies on mean-shift clustering~\cite{fukunaga1975estimation, cheng1995mean}, which iteratively shifts each point to a higher-density position until convergence. The points converging to the same position are assigned to the same cluster. The procedure is reported in the supplementary material under Algorithm S.2, and it uses a flat kernel function requiring a bandwidth parameter $h$. The kernel bandwidth defines the neighborhood size for density estimation, influencing how points are grouped into clusters: a smaller bandwidth results in more compact clusters, while a larger one creates broader clusters. However, the bandwidth's impact on clustering is influenced by data density rather than directly by the space's size, making it crucial to choose a bandwidth suited to the dataset's scale for meaningful results (see Sec.~\ref{sec:parameter_analysis}).
Due to the bandwidth's geometrical role, MGP-BBBC uses its value $h$ to initialize the threshold $th$ of the filtering in the survival stage (see Sec.~\ref{sec:survival}).

Once the big-bang operator retrieves a set of clusters $C$ (line 2), it identifies the individual with the best fitness for each cluster (lines 6-9), which is added into the set $COM$ (line 10). Lastly, the big-bang operator defines the number of desired offspring that will be generated for each center of mass: ultimately, the overall number of offspring must be equal to $n$; however, the number of clusters (and thus, centers of mass) retrieved by the mean-shift clustering varies based on the input dataset. In early generations, it is likely that the big-bang operator will retrieve many centers of mass, whereas in late generations, there will be fewer. Therefore, MGP-BBBC assigns a desired number of offspring per center of mass $OPC$ (which total sums up to $n$) based on the niche count $NC$ for each cluster, stored at line 11. Specifically, it assigns more offspring to centers of mass having a niche count smaller than the average of $NC$, to promote isolated individuals and prevent high-density regions from being overcrowded.
The procedure is reported in Algorithm~\ref{alg:offspring_per_com} and executed at line 12.

With the time complexity of the mean-shift clustering being $\mathcal{O}(n^2)$ (the number of iterations for convergence is not an input, and the bandwidth $h$ influences the algorithm's behavior but does not change the fundamental complexity), a number of clusters that is upper bounded by $n$ and therefore $\mathcal{O}(n)$ for both finding the best individual for each cluster and calculating $OPC$, the time complexity of the big-crunch operator of MGP-BBBC is dominated by the clustering: $\mathcal{O}(n^2)$.
 


\IncMargin{1em}
\begin{algorithm} [b!]
    \SetKwFunction{Clust}{meanShiftClustering}
    \SetKwFunction{Max}{max}
    \SetKwFunction{Off}{offspringPerCOM}
    \SetKwInOut{Input}{input}
    \SetKwInOut{Output}{output}
    \Input{Archive $A$, clustering bandwidth $h$}
    \Output{Array of centers of mass $COM$, array containing number of desired offspring per each center of mass $OPC$}
    \Begin
    {	
        $C \gets \Clust(A.\texttt{x}, h)$\;
        $COM \gets \emptyset$\;
        $NC \gets \emptyset$\;
        \For{$i \gets 1$ \textbf{to} $|C|$}
        {
            $c_{\text{best}} \gets (C_{i_1})$\;
            \For{$c_j \in C_i$}{
                /* for maximization problems */ \newline
                \If{$c_{\text{best}}.\texttt{fit} < c_j.\texttt{fit}$}
                {
                    $c_{\text{best}} \gets c_j$\;
                }
            }
               
            $COM \gets COM \cup \{c_{\text{best}}\} $\;
            $NC \gets NC \cup \{|C_i|\} $\;
        }
        $OPC \gets \Off(|A|,NC)$\;
    \KwRet{$COM, OPC$};
    }
    \caption{Big Crunch}\label{alg:big_crunch}
\end{algorithm}\DecMargin{1em}
\IncMargin{1em}
\begin{algorithm}
    \SetKwFunction{Clust}{meanShiftClustering}
    \SetKwFunction{Eval}{evaluation}
    \SetKwFunction{Max}{max}
    \SetKwFunction{Mod}{mod}
    \SetKwFunction{Sum}{sum}
    \SetKwFunction{Mean}{mean}
    \SetKwFunction{Floor}{floor}
    \SetKwFunction{Round}{round}
    \SetKwFunction{Rand}{randi}
    \SetKwFunction{Bang}{bigBang}
    \SetKwFunction{Crunch}{bigCrunch}
    \SetKwFunction{Sur}{survival}
    \SetKwRepeat{Do}{do}{while}
    \SetKwInOut{Input}{input}
    \SetKwInOut{Output}{output}
    \Input{Size of desired offspring population $n$, niche count per center of mass $NC$}
    \Output{Array containing number of desired offspring per each center of mass $OPC$}
    \Begin
    {	
        $OPC \gets \emptyset$\;
        $avg \gets \Mean(NC)$\;
        \For{$i \gets 1$ \textbf{to} $|NC|$}
        {
            $OPC_i \gets \Round(avg)$\;
        }
        $n' \gets \Sum(OPC)$\;
        \If{$n' < n$}
        {
            \For{$i \gets 1$ \textbf{to} $\Mod(n, n')$}
            {
                \Do
                {$NC_j> \Floor{avg}$} 
                {
                    $j \gets \Rand(1,|NC|)$\;
                }
                $OPC_j \gets OPC_j+1$\;
            }
        }
        \ElseIf{$n' > n$}
        {
            \For{$i \gets 1$ \textbf{to} $n' - n$}
            {
                \Do
                {$NC_j < \Floor{avg}$} 
                {
                    $j \gets \Rand(1,|NC|)$\;
                }
                $OPC_j \gets OPC_j-1$\;
            }
        }
        \KwRet{$OPC$};
    }
    \caption{Offspring per Center of Mass}\label{alg:offspring_per_com}
\end{algorithm}\DecMargin{1em}


\IncMargin{1em}
\begin{algorithm} [t!]
    \SetKwFunction{Expexp}{getExtent}
    \SetKwFunction{Rand}{rand}
    \SetKwFunction{Min}{min}
    \SetKwFunction{Max}{max}
    \SetKwInOut{Input}{input}
    \SetKwInOut{Output}{output}
    \Input{Array of centers of mass $COM$, generation number $it$, max number of generations $g$, array containing the number of desired offspring per each center of mass $OPC$, optimization problem $f$}
    \Output{Offspring population $O$}
    \Begin
    {	
        $O \gets \emptyset$\;
        $e \gets \Expexp(it,g,f)$\;
        \For{$i \gets 1$ \textbf{to} $|COM|$}
        {
            \For{$j \gets 1$ \textbf{to} $OPC_i$}
            {
                $X \gets COM_i.\texttt{x} + e\cdot\Rand(-1,1)$\;
                $X \gets \Min(\Max(X,f.\texttt{lowB}),f.\texttt{uppB})$\;
                
                
                $O \gets O \cup \{X\}$\;
            }
        }
        
    \KwRet{$O$};
    }
    \caption{Big Bang}\label{alg:big_bang}
\end{algorithm}\DecMargin{1em}
\IncMargin{1em}
\begin{algorithm}
    \SetKwFunction{Dist}{pairwiseDistances}
    \SetKwFunction{Floor}{floor}
    \SetKwFunction{Sort}{sort}
    \SetKwRepeat{Do}{do}{while}
    \SetKwInOut{Input}{input}
    \SetKwInOut{Output}{output}
    \Input{Current iteration counter $it$, max number of generations $g$, optimization problem $f$}
    \Output{Extent of expansion $e$}
    \Begin
    {	
        \If{$it < g \cdot 0.6$}{
            $e_1 \gets \frac{f.\texttt{uppB} - f.\texttt{lowB}}{4}$\;
            $a \gets \frac{e_1 - 1.0 \cdot 10^{-1}}{\log\left(g \cdot 0.6\right)}$\;
            $e \gets e_1 - a \cdot \log\left(it + 1\right)$\;
        }
        \Else{
            $I \gets \left\{10^{-1}, 10^{-2}, 10^{-3}, 1.0 10^{-4}, 10^{-5}\right\}$\;
            $z \gets \Floor(\frac{g - it}{5})$\;
            \For{$i \gets 1$ \textbf{to} $5$}{
                \If{$z \cdot (i-1) \leq it \wedge it < z \cdot i$}{
                    $e \gets I_i$\;
                }
            }
        }
        \KwRet{$e$};
    }
    \caption{Get Bang Extent of Expansion}\label{alg:extent}
\end{algorithm}\DecMargin{1em}

\begin{figure}[h!]
    \centering
    \includegraphics[width=\columnwidth]{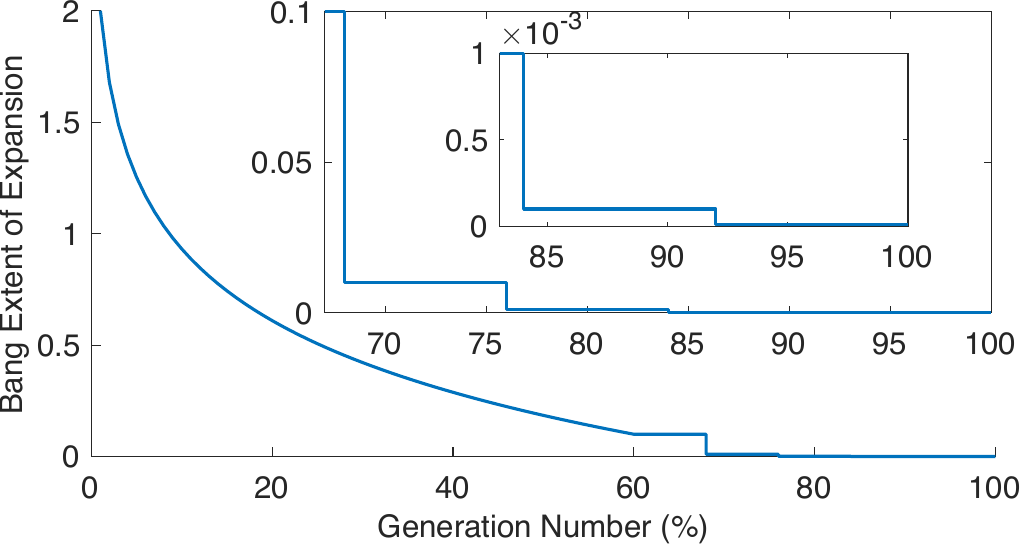}
    \caption{Reduction of the \textit{bang extent of expansion} over generations. In the example, the extent starts from $2.0$ and decreases logarithmically until the $60\%$ of the total number of generations (exploration phase) when it reaches a value of $1.0E-01$. After that, it decreases with constant steps to target the other levels of accuracy ($1.0E-02$, $1.0E-03$, $1.0E-04$, and $1.0E-05$) in an equal number of generations (exploitation phase). Zoomed-in plots allow readers to appreciate the details of the exploitation phase.
    }
    \label{fig:expexp_conv}
\end{figure}

\subsection{Big Bang} \label{sec:big_bang}

Algorithm~\ref{alg:big_bang} shows the big-bang operator of MGP-BBBC, which generates new offspring $O$ from the centers of mass $COM$ retrieved by the big-crunch operator (see Sec.~\ref{alg:big_crunch}). While generating offspring, it is essential to properly balance exploration and exploitation to retrieve peaks with high accuracy. To achieve this balance, MGP-BBBC dynamically tunes the extent of expansion of the big-bang operator (line 2). While the standard BBBC algorithm decrements the extent linearly as specified in Eqn.~(\ref{eq:bang}), MGP-BBBC divides it into two trends based on the total number of generations $g$ -- as shown in Fig.~\ref{fig:expexp_conv}. During the first $60\%$ of generations, the extent is decreased logarithmically from a starting value (empirically set to be one-fourth of the problem's bound when it was observed a population distributed on the search rather than on the bounds) to $1.0E-01$: in this phase, MGP-BBBC will \textit{explore} the surroundings of the centers of mass, localizing as many global peaks as possible. During the last $40\%$ of generations, the extent will follow a constant trend with five uniformly distributed steps, going from $1.0E-01$ to $1.0E-05$: in this phase, MGP-BBBC will \textit{exploit} the centers of mass, converging with high accuracy to the respective global peak. The pseudocode of the procedure is reported in Algorithm~\ref{alg:extent}.

Once the extent of expansion $e$ is set for its current iteration, the big-bang operator applies it on every center of mass (line 4), producing a number of offspring defined by the respective element in $OPC$ (line 5). Each offspring is produced by applying a random disturbance with, at most, a magnitude equal to $e$ (lines 6-8). Note that the new individual $X$ is generated within the bounds of the search space with the filtering formula at line 7, which is applied to all decision variables and their respective bounds.

With the time complexity of calculating the extent being constant, the complexity of the big-bang operator is only determined linearly by the amount of offspring to be produced:  $\mathcal{O}(nd)$, where $d$ is the dimensionality of the MMOP.

\section{Comparative Study} \label{sec:experiments}

\begin{table*}[]
\centering
\caption{Properties of the twenty multimodal test functions in the CEC'2013 benchmark set}
\label{tab:functions_CEC13}
\resizebox{\linewidth}{!}{%
\begin{tabular}{|c|c|c|c|c|c|}
\hline
\textbf{Function} & \textbf{Name} & \textbf{Dimension Size} & \textbf{No. of Global Optima} & \textbf{No. of Local Optima} & \textbf{Max Function Evaluations} \\ \hline
$F_{1}$ & Five-Uneven-Peak Trap & 1 & 2 & 3 & $\expnumber{5.0}{+04}$ \\ \hline
$F_{2}$ & Equal Maxima & 1 & 5 & 0 & $\expnumber{5.0}{+04}$ \\ \hline
$F_{3}$ & Uneven Decreasing Maxima & 1 & 1 & 4 & $\expnumber{5.0}{+04}$ \\ \hline
$F_{4}$ & Himmelblau & 2 & 4 & 0 & $\expnumber{5.0}{+04}$ \\ \hline
$F_{5}$ & Six-Hump Camel Back & 2 & 2 & 2 & $\expnumber{5.0}{+04}$ \\ \hline
$F_{6}$ & Shubert with 2D & 2 & 18 & many & $\expnumber{2.0}{+05}$ \\ \hline
$F_{7}$ & Vincent with 2D & 2 & 36 & 0 & $\expnumber{2.0}{+05}$ \\ \hline
$F_{8}$ & Shubert with 3D & 3 & 81 & many & $\expnumber{4.0}{+05}$ \\ \hline
$F_{9}$ & Vincent with 3D & 3 & 216 & 0 & $\expnumber{4.0}{+05}$ \\ \hline
$F_{10}$ & Modified Rastrigin & 2 & 12 & 0 & $\expnumber{2.0}{+05}$ \\ \hline
$F_{11}$ & Composition Function 1 with 2D & 2 & 6 & many & $\expnumber{2.0}{+05}$ \\ \hline
$F_{12}$ & Composition Function 2 with 2D & 2 & 8 & many & $\expnumber{2.0}{+05}$ \\ \hline
$F_{13}$ & Composition Function 3 with 2D & 2 & 6 & many & $\expnumber{2.0}{+05}$ \\ \hline
$F_{14}$ & Composition Function 3 with 3D & 3 & 6 & many & $\expnumber{4.0}{+05}$ \\ \hline
$F_{15}$ & Composition Function 4 with 3D & 3 & 8 & many & $\expnumber{4.0}{+05}$ \\ \hline
$F_{16}$ & Composition Function 3 with 5D & 5 & 6 & many & $\expnumber{4.0}{+05}$ \\ \hline
$F_{17}$ & Composition Function 4 with 5D & 5 & 8 & many & $\expnumber{4.0}{+05}$ \\ \hline
$F_{18}$ & Composition Function 3 with 10D & 10 & 6 & many & $\expnumber{4.0}{+05}$ \\ \hline
$F_{19}$ & Composition Function 4 with 10D & 10 & 8 & many & $\expnumber{4.0}{+05}$ \\ \hline
$F_{20}$ & Composition Function 4 with 20D & 20 & 8 & many & $\expnumber{4.0}{+05}$ \\ \hline
\end{tabular}%
}
\end{table*}

\subsection{Benchmark Functions and Performance Metrics} \label{sec:benchmark_metrics}

We used the CEC'2013 benchmark set~\cite{li2013benchmark} to evaluate MGP-BBBC performances. The CEC'2013 benchmark set contains twenty multimodal maximization problems, listed in Table~\ref{tab:functions_CEC13} with their properties. Specifically, $F_1-F_3$ are simple mono-dimensional functions, $F_4-F_7$ and $F_{10}$ are scalable two-dimensional functions, $F_8-F_9$ are scalable three-dimensional functions, and $F_{11}-F_{20}$ are complex composition functions with dimensionality ranging from two to twenty. 


We executed each problem by setting a maximum number of function evaluations (MaxFEs) based on the procedure established in the literature~\cite{li2013benchmark}, which we report in the rightmost column of Table~\ref{tab:functions_CEC13}. MaxFEs sets the total number of runs (NR) for each problem, given the population size. Based on this, we evaluated the algorithm's performance with the following two metrics:
\begin{itemize}
    \item \textit{peak ratio (PR)}, the average percentage of global peaks found over multiple runs, defined in Eqn.~(\ref{eq:pr}) as the sum of the number of global peaks for each run (NPF) divided by the total number of global peaks (TPN) times the total number of runs (NR); and

    \begin{equation}
        \label{eq:pr}
        \text{PR} = \frac{\sum^{\text{NR}}_{i=1}{\text{NPF}_i}}{\text{TNP} \cdot \text{NR}}
    \end{equation}
    
    \item \textit{success ratio (SR)}, the percentage of successful runs among multiple runs (i.e., runs that correctly retrieved all global optima), defined in Eqn.~(\ref{eq:sr}) as the ratio between the number of successful (NSR) and the total number of runs (NR).

    \begin{equation}
        \label{eq:sr}
        \text{SR} = \frac{\text{NSR}}{\text{NR}}
    \end{equation}
    
\end{itemize}

\begin{table}[]
\centering
\caption{Experimental results of peak ratio and success rate on the CEC’2013 benchmark set at five accuracy levels}
\label{tab:mgpbbbc_results}
\resizebox{\columnwidth}{!}{
\begin{tabular}{|c|cc|cc|cc|cc|cc|}
\hline
\multirow{2}{*}{Func.} & \multicolumn{2}{c|}{$\expnumber{1.0}{-01}$} & \multicolumn{2}{c|}{$\expnumber{1.0}{-02}$} & \multicolumn{2}{c|}{$\expnumber{1.0}{-03}$} & \multicolumn{2}{c|}{$\expnumber{1.0}{-04}$} & \multicolumn{2}{c|}{$\expnumber{1.0}{-05}$} \\ \cline{2-11} 
 & \multicolumn{1}{c|}{PR} & SR & \multicolumn{1}{c|}{PR} & SR & \multicolumn{1}{c|}{PR} & SR & \multicolumn{1}{c|}{PR} & SR & \multicolumn{1}{c|}{PR} & SR \\ \hline
$F_{1}$ & \multicolumn{1}{c|}{1.000} & 1.000 & \multicolumn{1}{c|}{1.000} & 1.000 & \multicolumn{1}{c|}{1.000} & 1.000 & \multicolumn{1}{c|}{1.000} & 1.000 & \multicolumn{1}{c|}{1.000} & 1.000 \\ \hline
$F_{2}$ & \multicolumn{1}{c|}{1.000} & 1.000 & \multicolumn{1}{c|}{1.000} & 1.000 & \multicolumn{1}{c|}{1.000} & 1.000 & \multicolumn{1}{c|}{1.000} & 1.000 & \multicolumn{1}{c|}{1.000} & 1.000 \\ \hline
$F_{3}$ & \multicolumn{1}{c|}{1.000} & 1.000 & \multicolumn{1}{c|}{1.000} & 1.000 & \multicolumn{1}{c|}{1.000} & 1.000 & \multicolumn{1}{c|}{1.000} & 1.000 & \multicolumn{1}{c|}{1.000} & 1.000 \\ \hline
$F_{4}$ & \multicolumn{1}{c|}{1.000} & 1.000 & \multicolumn{1}{c|}{1.000} & 1.000 & \multicolumn{1}{c|}{1.000} & 1.000 & \multicolumn{1}{c|}{1.000} & 1.000 & \multicolumn{1}{c|}{1.000} & 1.000 \\ \hline
$F_{5}$ & \multicolumn{1}{c|}{1.000} & 1.000 & \multicolumn{1}{c|}{1.000} & 1.000 & \multicolumn{1}{c|}{1.000} & 1.000 & \multicolumn{1}{c|}{1.000} & 1.000 & \multicolumn{1}{c|}{1.000} & 1.000 \\ \hline
$F_{6}$ & \multicolumn{1}{c|}{1.000} & 1.000 & \multicolumn{1}{c|}{1.000} & 1.000 & \multicolumn{1}{c|}{1.000} & 1.000 & \multicolumn{1}{c|}{1.000} & 1.000 & \multicolumn{1}{c|}{1.000} & 1.000 \\ \hline
$F_{7}$ & \multicolumn{1}{c|}{0.999} & 0.980 & \multicolumn{1}{c|}{0.999} & 0.960 & \multicolumn{1}{c|}{0.999} & 0.960 & \multicolumn{1}{c|}{0.998} & 0.960 & \multicolumn{1}{c|}{0.998} & 0.960 \\ \hline
$F_{8}$ & \multicolumn{1}{c|}{1.000} & 1.000 & \multicolumn{1}{c|}{1.000} & 1.000 & \multicolumn{1}{c|}{1.000} & 1.000 & \multicolumn{1}{c|}{1.000} & 1.000 & \multicolumn{1}{c|}{1.000} & 1.000 \\ \hline
$F_{9}$ & \multicolumn{1}{c|}{0.570} & 0.000 & \multicolumn{1}{c|}{0.540} & 0.000 & \multicolumn{1}{c|}{0.503} & 0.000 & \multicolumn{1}{c|}{0.478} & 0.000 & \multicolumn{1}{c|}{0.448} & 0.000 \\ \hline
$F_{10}$ & \multicolumn{1}{c|}{1.000} & 1.000 & \multicolumn{1}{c|}{1.000} & 1.000 & \multicolumn{1}{c|}{1.000} & 1.000 & \multicolumn{1}{c|}{1.000} & 1.000 & \multicolumn{1}{c|}{1.000} & 1.000 \\ \hline
$F_{11}$ & \multicolumn{1}{c|}{1.000} & 1.000 & \multicolumn{1}{c|}{1.000} & 1.000 & \multicolumn{1}{c|}{1.000} & 1.000 & \multicolumn{1}{c|}{1.000} & 1.000 & \multicolumn{1}{c|}{1.000} & 1.000 \\ \hline
$F_{12}$ & \multicolumn{1}{c|}{1.000} & 1.000 & \multicolumn{1}{c|}{1.000} & 1.000 & \multicolumn{1}{c|}{1.000} & 1.000 & \multicolumn{1}{c|}{1.000} & 1.000 & \multicolumn{1}{c|}{1.000} & 1.000 \\ \hline
$F_{13}$ & \multicolumn{1}{c|}{1.000} & 1.000 & \multicolumn{1}{c|}{1.000} & 1.000 & \multicolumn{1}{c|}{1.000} & 1.000 & \multicolumn{1}{c|}{1.000} & 1.000 & \multicolumn{1}{c|}{1.000} & 1.000 \\ \hline
$F_{14}$ & \multicolumn{1}{c|}{0.943} & 0.660 & \multicolumn{1}{c|}{0.943} & 0.660 & \multicolumn{1}{c|}{0.940} & 0.640 & \multicolumn{1}{c|}{0.930} & 0.580 & \multicolumn{1}{c|}{0.913} & 0.480 \\ \hline
$F_{15}$ & \multicolumn{1}{c|}{0.723} & 0.000 & \multicolumn{1}{c|}{0.720} & 0.000 & \multicolumn{1}{c|}{0.720} & 0.000 & \multicolumn{1}{c|}{0.720} & 0.000 & \multicolumn{1}{c|}{0.720} & 0.000 \\ \hline
$F_{16}$ & \multicolumn{1}{c|}{0.730} & 0.000 & \multicolumn{1}{c|}{0.723} & 0.000 & \multicolumn{1}{c|}{0.710} & 0.000 & \multicolumn{1}{c|}{0.707} & 0.000 & \multicolumn{1}{c|}{0.707} & 0.000 \\ \hline
$F_{17}$ & \multicolumn{1}{c|}{0.683} & 0.000 & \multicolumn{1}{c|}{0.598} & 0.000 & \multicolumn{1}{c|}{0.598} & 0.000 & \multicolumn{1}{c|}{0.598} & 0.000 & \multicolumn{1}{c|}{0.598} & 0.000 \\ \hline
$F_{18}$ & \multicolumn{1}{c|}{0.667} & 0.000 & \multicolumn{1}{c|}{0.667} & 0.000 & \multicolumn{1}{c|}{0.667} & 0.000 & \multicolumn{1}{c|}{0.667} & 0.000 & \multicolumn{1}{c|}{0.620} & 0.000 \\ \hline
$F_{19}$ & \multicolumn{1}{c|}{0.380} & 0.000 & \multicolumn{1}{c|}{0.373} & 0.000 & \multicolumn{1}{c|}{0.373} & 0.000 & \multicolumn{1}{c|}{0.373} & 0.000 & \multicolumn{1}{c|}{0.368} & 0.000 \\ \hline
$F_{20}$ & \multicolumn{1}{c|}{0.363} & 0.000 & \multicolumn{1}{c|}{0.363} & 0.000 & \multicolumn{1}{c|}{0.363} & 0.000 & \multicolumn{1}{c|}{0.358} & 0.000 & \multicolumn{1}{c|}{0.000} & 0.000 \\ \hline
\end{tabular}%
}
\end{table}

Lastly, we analyzed the results based on five accuracy levels: from $\varepsilon = \expnumber{1.0}{-01}$ to $\varepsilon = \expnumber{1.0}{-05}$. Based on the procedure established in the literature~\cite{wang2017dual, deb2012multimodal}, we mainly discuss the results with accuracy $\varepsilon = \expnumber{1.0}{-04}$, whereas the results with other levels are reported in Tab.~\ref{tab:mgpbbbc_results}.

\begin{table}[]
\centering
\caption{MGP-BBBC Best Parameter Settings}
\label{tab:best_param_settings}
\resizebox{\columnwidth}{!}{
\begin{tabular}{|c|c|c|c|c|c|c|}
\cline{1-3} \cline{5-7}
\textbf{Func.} & \textbf{Pop. Size ($n$)} & \textbf{$h$} & \textbf{} & \textbf{Func.} & \textbf{Pop. Size ($n$)} & \textbf{$h$} \\ \cline{1-3} \cline{5-7} 
$F_{1}$ & 1000 & 0.80 &  & $F_{11}$ & 1000 & 0.40 \\ \cline{1-3} \cline{5-7} 
$F_{2}$ & 1000 & 0.08 &  & $F_{12}$ & 1000 & 0.60 \\ \cline{1-3} \cline{5-7} 
$F_{3}$ & 1000 & 0.80 &  & $F_{13}$ & 1000 & 0.40 \\ \cline{1-3} \cline{5-7} 
$F_{4}$ & 1000 & 0.80 &  & $F_{14}$ & 1000 & 1.40 \\ \cline{1-3} \cline{5-7} 
$F_{5}$ & 1000 & 0.80 &  & $F_{15}$ & 500 & 2.00 \\ \cline{1-3} \cline{5-7} 
$F_{6}$ & 1000 & 0.20 &  & $F_{16}$ & 1000 & 3.60 \\ \cline{1-3} \cline{5-7} 
$F_{7}$ & 500 & 0.20 &  & $F_{17}$ & 500 & 4.00 \\ \cline{1-3} \cline{5-7} 
$F_{8}$ & 1000 & 0.60 &  & $F_{18}$ & 500 & 4.00 \\ \cline{1-3} \cline{5-7} 
$F_{9}$ & 1000 & 0.40 &  & $F_{19}$ & 500 & 6.00 \\ \cline{1-3} \cline{5-7} 
$F_{10}$ & 1000 & 0.40 &  & $F_{20}$ & 500 & 10.00 \\ \cline{1-3} \cline{5-7} 
\end{tabular}%
}
\end{table}

\subsection{Compared Algorithms and Parameter Configurations} \label{sec:algorithm_list}

We compared MGP-BBBC with its base version k-BBBC~\cite{yenin2023multi} and twelve other MMOEAs from the state-of-the-art: 
SPEA-DE/rand/1~\cite{xia2024learning},
SSGA~\cite{de2014dynamic},
CDE~\cite{thomsen2004multimodal},
RS-CMSA-EAII~\cite{ahrari2021static},
LBP-ADE~\cite{zhao2019local},
HillVallEA19~\cite{maree2019benchmarking},
ADE-DDE~\cite{wang2020guardhealth},
ANDE~\cite{wang2019automatic},
NCD-DE~\cite{jiang2021optimizing},
CFNDE~\cite{ma2023coarse},
DIDE~\cite{chen2019distributed},
MOMMOP~\cite{wang2014mommop},
EMO-MMO~\cite{cheng2017evolutionary}	
-- most of these algorithms participated in competitions for either CEC or GECCO throughout 2017 and 2020 (the latest at the time of this work), others are taken from the latest published articles in the literature.
The settings of MGP-BBBC were set according to the analysis described in Sec.~\ref{sec:parameter_analysis} and reported on Table~\ref{tab:best_param_settings}, whereas the results of the state-of-the-art algorithms were collected from the CEC repository or according to their corresponding references. 

We ran fifty independent runs (NR$=50$) and averaged the results for a fair comparison. We used the Wilcoxon rank-sum test~\cite{demvsar2006statistical} to statistically evaluate the results, with significance level $\alpha=0.05$. In the tables reporting the results, we added a $+$ sign to identify when MGP-BBBC is significantly better, a $-$ sign when it is significantly worse, and a $\approx$ sign when results are not significantly different.

\begin{table*}[]
\centering
\caption{Comparative results of peak ratio and success rate on the CEC'2013 benchmark set at the accuracy level $\varepsilon=1.0E-04$}
\label{tab:comparison_4}
\resizebox{\textwidth}{!}{
\begin{tabular}{|ccc|cc|cc|cc|cc|cc|cc|}
\hline
\multicolumn{1}{|c|}{\multirow{2}{*}{Func.}} & \multicolumn{2}{c|}{MGP-BBBC} & \multicolumn{2}{c|}{kBBBC} & \multicolumn{2}{c|}{SSGA} & \multicolumn{2}{c|}{SPEA-DE/rand/1} & \multicolumn{2}{c|}{RS-CMSA-EAII} & \multicolumn{2}{c|}{LBP-ADE} & \multicolumn{2}{c|}{HillVallEA19} \\  \cline{2-15} 
\multicolumn{1}{|c|}{} & \multicolumn{1}{c|}{PR} & SR & \multicolumn{1}{c|}{PR} & SR & \multicolumn{1}{c|}{PR} & SR & \multicolumn{1}{c|}{PR} & SR & \multicolumn{1}{c|}{PR} & SR & \multicolumn{1}{c|}{PR} & SR & \multicolumn{1}{c|}{PR} & SR \\ \hline
\multicolumn{1}{|c|}{$F_{1}$} & \multicolumn{1}{c|}{\textbf{1.000}} & \textbf{1.000} & \multicolumn{1}{c|}{\textbf{1.000($\approx$)}} & \textbf{1.000} & \multicolumn{1}{c|}{\textbf{1.000($\approx$)}} & \textbf{1.000} & \multicolumn{1}{c|}{\textbf{1.000($\approx$)}} & \textbf{1.000} & \multicolumn{1}{c|}{\textbf{1.000($\approx$)}} & \textbf{1.000} & \multicolumn{1}{c|}{\textbf{1.000($\approx$)}} & \textbf{1.000} & \multicolumn{1}{c|}{\textbf{1.000($\approx$)}} & \textbf{1.000} \\ \hline
\multicolumn{1}{|c|}{$F_{2}$} & \multicolumn{1}{c|}{\textbf{1.000}} & \textbf{1.000} & \multicolumn{1}{c|}{\textbf{1.000($\approx$)}} & \textbf{1.000} & \multicolumn{1}{c|}{\textbf{1.000($\approx$)}} & \textbf{1.000} & \multicolumn{1}{c|}{\textbf{1.000($\approx$)}} & \textbf{1.000} & \multicolumn{1}{c|}{\textbf{1.000($\approx$)}} & \textbf{1.000} & \multicolumn{1}{c|}{\textbf{1.000($\approx$)}} & \textbf{1.000} & \multicolumn{1}{c|}{\textbf{1.000($\approx$)}} & \textbf{1.000} \\ \hline
\multicolumn{1}{|c|}{$F_{3}$} & \multicolumn{1}{c|}{\textbf{1.000}} & \textbf{1.000} & \multicolumn{1}{c|}{\textbf{1.000($\approx$)}} & \textbf{1.000} & \multicolumn{1}{c|}{\textbf{1.000($\approx$)}} & \textbf{1.000} & \multicolumn{1}{c|}{\textbf{1.000($\approx$)}} & \textbf{1.000} & \multicolumn{1}{c|}{\textbf{1.000($\approx$)}} & \textbf{1.000} & \multicolumn{1}{c|}{\textbf{1.000($\approx$)}} & \textbf{1.000} & \multicolumn{1}{c|}{\textbf{1.000($\approx$)}} & \textbf{1.000} \\ \hline
\multicolumn{1}{|c|}{$F_{4}$} & \multicolumn{1}{c|}{\textbf{1.000}} & \textbf{1.000} & \multicolumn{1}{c|}{\textbf{1.000($\approx$)}} & \textbf{1.000} & \multicolumn{1}{c|}{0.155($+$)} & 0.000 & \multicolumn{1}{c|}{\textbf{1.000($\approx$)}} & \textbf{1.000} & \multicolumn{1}{c|}{\textbf{1.000($\approx$)}} & \textbf{1.000} & \multicolumn{1}{c|}{\textbf{1.000($\approx$)}} & \textbf{1.000} & \multicolumn{1}{c|}{\textbf{1.000($\approx$)}} & \textbf{1.000} \\ \hline
\multicolumn{1}{|c|}{$F_{5}$} & \multicolumn{1}{c|}{\textbf{1.000}} & \textbf{1.000} & \multicolumn{1}{c|}{\textbf{1.000($\approx$)}} & \textbf{1.000} & \multicolumn{1}{c|}{0.460($+$)} & 0.200 & \multicolumn{1}{c|}{\textbf{1.000($\approx$)}} & \textbf{1.000} & \multicolumn{1}{c|}{\textbf{1.000($\approx$)}} & \textbf{1.000} & \multicolumn{1}{c|}{\textbf{1.000($\approx$)}} & \textbf{1.000} & \multicolumn{1}{c|}{\textbf{1.000($\approx$)}} & \textbf{1.000} \\ \hline
\multicolumn{1}{|c|}{$F_{6}$} & \multicolumn{1}{c|}{\textbf{1.000}} & \textbf{1.000} & \multicolumn{1}{c|}{0.943($+$)} & 0.280 & \multicolumn{1}{c|}{\textbf{1.000($\approx$)}} & \textbf{1.000} & \multicolumn{1}{c|}{\textbf{1.000($\approx$)}} & \textbf{1.000} & \multicolumn{1}{c|}{\textbf{1.000($\approx$)}} & \textbf{1.000} & \multicolumn{1}{c|}{\textbf{1.000($\approx$)}} & \textbf{1.000} & \multicolumn{1}{c|}{\textbf{1.000($\approx$)}} & \textbf{1.000} \\ \hline
\multicolumn{1}{|c|}{$F_{7}$} & \multicolumn{1}{c|}{0.998} & 0.960 & \multicolumn{1}{c|}{0.744($+$)} & 0.000 & \multicolumn{1}{c|}{0.901($+$)} & 0.000 & \multicolumn{1}{c|}{0.972($+$)} & 0.240 & \multicolumn{1}{c|}{\textbf{1.000($\approx$)}} & \textbf{1.000} & \multicolumn{1}{c|}{0.889($+$)} & 0.000 & \multicolumn{1}{c|}{\textbf{1.000($\approx$)}} & \textbf{1.000} \\ \hline
\multicolumn{1}{|c|}{$F_{8}$} & \multicolumn{1}{c|}{\textbf{1.000}} & \textbf{1.000} & \multicolumn{1}{c|}{0.000($+$)} & 0.000 & \multicolumn{1}{c|}{\textbf{1.000($\approx$)}} & \textbf{1.000} & \multicolumn{1}{c|}{0.533($+$)} & 0.000 & \multicolumn{1}{c|}{0.975($+$)} & 0.000 & \multicolumn{1}{c|}{0.575($+$)} & 0.000 & \multicolumn{1}{c|}{0.975($+$)} & 0.000 \\ \hline
\multicolumn{1}{|c|}{$F_{9}$} & \multicolumn{1}{c|}{0.478} & 0.000 & \multicolumn{1}{c|}{0.005($+$)} & 0.000 & \multicolumn{1}{c|}{0.518($-$)} & 0.000 & \multicolumn{1}{c|}{0.723($-$)} & 0.000 & \multicolumn{1}{c|}{0.990($-$)} & 0.000 & \multicolumn{1}{c|}{0.476($+$)} & 0.000 & \multicolumn{1}{c|}{0.972($-$)} & 0.000 \\ \hline
\multicolumn{1}{|c|}{$F_{10}$} & \multicolumn{1}{c|}{\textbf{1.000}} & \textbf{1.000} & \multicolumn{1}{c|}{\textbf{1.000($\approx$)}} & \textbf{1.000} & \multicolumn{1}{c|}{\textbf{1.000($\approx$)}} & \textbf{1.000} & \multicolumn{1}{c|}{\textbf{1.000($\approx$)}} & \textbf{1.000} & \multicolumn{1}{c|}{\textbf{1.000($\approx$)}} & \textbf{1.000} & \multicolumn{1}{c|}{\textbf{1.000($\approx$)}} & \textbf{1.000} & \multicolumn{1}{c|}{\textbf{1.000($\approx$)}} & \textbf{1.000} \\ \hline
\multicolumn{1}{|c|}{$F_{11}$} & \multicolumn{1}{c|}{\textbf{1.000}} & \textbf{1.000} & \multicolumn{1}{c|}{\textbf{1.000($\approx$)}} & \textbf{1.000} & \multicolumn{1}{c|}{\textbf{1.000($\approx$)}} & \textbf{1.000} & \multicolumn{1}{c|}{\textbf{1.000($\approx$)}} & \textbf{1.000} & \multicolumn{1}{c|}{\textbf{1.000($\approx$)}} & \textbf{1.000} & \multicolumn{1}{c|}{0.674($+$)} & 0.000 & \multicolumn{1}{c|}{\textbf{1.000($\approx$)}} & \textbf{1.000} \\ \hline
\multicolumn{1}{|c|}{$F_{12}$} & \multicolumn{1}{c|}{\textbf{1.000}} & \textbf{1.000} & \multicolumn{1}{c|}{0.920($+$)} & 0.360 & \multicolumn{1}{c|}{\textbf{1.000($\approx$)}} & \textbf{1.000} & \multicolumn{1}{c|}{\textbf{1.000($\approx$)}} & \textbf{1.000} & \multicolumn{1}{c|}{\textbf{1.000($\approx$)}} & \textbf{1.000} & \multicolumn{1}{c|}{0.750($+$)} & 0.000 & \multicolumn{1}{c|}{\textbf{1.000($\approx$)}} & \textbf{1.000} \\ \hline
\multicolumn{1}{|c|}{$F_{13}$} & \multicolumn{1}{c|}{\textbf{1.000}} & \textbf{1.000} & \multicolumn{1}{c|}{0.993($+$)} & 0.960 & \multicolumn{1}{c|}{0.957($+$)} & 0.760 & \multicolumn{1}{c|}{\textbf{1.000($\approx$)}} & \textbf{1.000} & \multicolumn{1}{c|}{0.993($+$)} & 0.980 & \multicolumn{1}{c|}{0.667($+$)} & 0.000 & \multicolumn{1}{c|}{\textbf{1.000($\approx$)}} & \textbf{1.000} \\ \hline
\multicolumn{1}{|c|}{$F_{14}$} & \multicolumn{1}{c|}{\textbf{0.930}} & \textbf{0.580} & \multicolumn{1}{c|}{0.857($+$)} & 0.140 & \multicolumn{1}{c|}{0.727($+$)} & 0.020 & \multicolumn{1}{c|}{0.839($+$)} & 0.000 & \multicolumn{1}{c|}{0.850($+$)} & 0.060 & \multicolumn{1}{c|}{0.667($+$)} & 0.000 & \multicolumn{1}{c|}{0.923($+$)} & 0.560 \\ \hline
\multicolumn{1}{|c|}{$F_{15}$} & \multicolumn{1}{c|}{0.720} & 0.000 & \multicolumn{1}{c|}{0.658($+$)} & 0.000 & \multicolumn{1}{c|}{0.563($+$)} & 0.000 & \multicolumn{1}{c|}{\textbf{0.750($-$)}} & 0.000 & \multicolumn{1}{c|}{\textbf{0.750($-$)}} & 0.000 & \multicolumn{1}{c|}{0.654($+$)} & 0.000 & \multicolumn{1}{c|}{\textbf{0.750($-$)}} & 0.000 \\ \hline
\multicolumn{1}{|c|}{$F_{16}$} & \multicolumn{1}{c|}{0.707} & 0.000 & \multicolumn{1}{c|}{0.710($-$)} & 0.000 & \multicolumn{1}{c|}{0.673($+$)} & 0.000 & \multicolumn{1}{c|}{0.663($+$)} & 0.000 & \multicolumn{1}{c|}{\textbf{0.883($-$)}} & 0.000 & \multicolumn{1}{c|}{0.667($+$)} & 0.000 & \multicolumn{1}{c|}{0.723($-$)} & 0.000 \\ \hline
\multicolumn{1}{|c|}{$F_{17}$} & \multicolumn{1}{c|}{0.598} & 0.000 & \multicolumn{1}{c|}{0.543($+$)} & 0.000 & \multicolumn{1}{c|}{0.485($+$)} & 0.000 & \multicolumn{1}{c|}{0.505($+$)} & 0.000 & \multicolumn{1}{c|}{\textbf{0.750($-$)}} & 0.000 & \multicolumn{1}{c|}{0.532($+$)} & 0.000 & \multicolumn{1}{c|}{\textbf{0.750($-$)}} & 0.000 \\ \hline
\multicolumn{1}{|c|}{$F_{18}$} & \multicolumn{1}{c|}{\textbf{0.667}} & 0.000 & \multicolumn{1}{c|}{0.440($+$)} & 0.000 & \multicolumn{1}{c|}{0.307($+$)} & 0.000 & \multicolumn{1}{c|}{0.170($+$)} & 0.000 & \multicolumn{1}{c|}{\textbf{0.667($\approx$)}} & 0.000 & \multicolumn{1}{c|}{\textbf{0.667($\approx$)}} & 0.000 & \multicolumn{1}{c|}{\textbf{0.667($\approx$)}} & 0.000 \\ \hline
\multicolumn{1}{|c|}{$F_{19}$} & \multicolumn{1}{c|}{0.373} & 0.000 & \multicolumn{1}{c|}{0.375($\approx$)} & 0.000 & \multicolumn{1}{c|}{0.023($+$)} & 0.000 & \multicolumn{1}{c|}{0.000($+$)} & 0.000 & \multicolumn{1}{c|}{\textbf{0.703($-$)}} & 0.000 & \multicolumn{1}{c|}{0.475($-$)} & 0.000 & \multicolumn{1}{c|}{0.593($-$)} & 0.000 \\ \hline
\multicolumn{1}{|c|}{$F_{20}$} & \multicolumn{1}{c|}{0.358} & 0.000 & \multicolumn{1}{c|}{0.000($+$)} & 0.000 & \multicolumn{1}{c|}{0.000($+$)} & 0.000 & \multicolumn{1}{c|}{0.025($+$)} & 0.000 & \multicolumn{1}{c|}{\textbf{0.618($-$)}} & 0.000 & \multicolumn{1}{c|}{0.275($+$)} & 0.000 & \multicolumn{1}{c|}{0.480($-$)} & 0.000 \\ \hline
\multicolumn{3}{|l|}{Total $+$   (Significantly better)} & \multicolumn{2}{c|}{11} & \multicolumn{2}{c|}{11} & \multicolumn{2}{c|}{8} & \multicolumn{2}{c|}{3} & \multicolumn{2}{c|}{11} & \multicolumn{2}{c|}{2} \\ \hline
\multicolumn{3}{|l|}{Total $\approx$ (Not   sig. different)} & \multicolumn{2}{c|}{8} & \multicolumn{2}{c|}{8} & \multicolumn{2}{c|}{10} & \multicolumn{2}{c|}{11} & \multicolumn{2}{c|}{8} & \multicolumn{2}{c|}{12} \\ \hline
\multicolumn{3}{|l|}{Total $-$   (Significantly worse)} & \multicolumn{2}{c|}{1} & \multicolumn{2}{c|}{1} & \multicolumn{2}{c|}{2} & \multicolumn{2}{c|}{6} & \multicolumn{2}{c|}{1} & \multicolumn{2}{c|}{6} \\ \hline
\multicolumn{1}{|c|}{\multirow{2}{*}{Func.}} & \multicolumn{2}{c|}{ADE-DDE} & \multicolumn{2}{c|}{ANDE} & \multicolumn{2}{c|}{NCD-DE} & \multicolumn{2}{c|}{CFNDE} & \multicolumn{2}{c|}{DIDE} & \multicolumn{2}{c|}{MOMMOP} & \multicolumn{2}{c|}{EMO-MMO} \\  \cline{2-15} 
\multicolumn{1}{|c|}{} & \multicolumn{1}{c|}{PR} & SR & \multicolumn{1}{c|}{PR} & SR & \multicolumn{1}{c|}{PR} & SR & \multicolumn{1}{c|}{PR} & SR & \multicolumn{1}{c|}{PR} & SR & \multicolumn{1}{c|}{PR} & SR & \multicolumn{1}{c|}{PR} & SR \\ \hline
\multicolumn{1}{|c|}{$F_{1}$} & \multicolumn{1}{c|}{\textbf{1.000($\approx$)}} & \textbf{1.000} & \multicolumn{1}{c|}{\textbf{1.000($\approx$)}} & \textbf{1.000} & \multicolumn{1}{c|}{\textbf{1.000($\approx$)}} & \textbf{1.000} & \multicolumn{1}{c|}{\textbf{1.000($\approx$)}} & \textbf{1.000} & \multicolumn{1}{c|}{\textbf{1.000($\approx$)}} & \textbf{1.000} & \multicolumn{1}{c|}{\textbf{1.000($\approx$)}} & \textbf{1.000} & \multicolumn{1}{c|}{\textbf{1.000($\approx$)}} & \textbf{1.000} \\ \hline
\multicolumn{1}{|c|}{$F_{2}$} & \multicolumn{1}{c|}{\textbf{1.000($\approx$)}} & \textbf{1.000} & \multicolumn{1}{c|}{\textbf{1.000($\approx$)}} & \textbf{1.000} & \multicolumn{1}{c|}{\textbf{1.000($\approx$)}} & \textbf{1.000} & \multicolumn{1}{c|}{\textbf{1.000($\approx$)}} & \textbf{1.000} & \multicolumn{1}{c|}{\textbf{1.000($\approx$)}} & \textbf{1.000} & \multicolumn{1}{c|}{\textbf{1.000($\approx$)}} & \textbf{1.000} & \multicolumn{1}{c|}{\textbf{1.000($\approx$)}} & \textbf{1.000} \\ \hline
\multicolumn{1}{|c|}{$F_{3}$} & \multicolumn{1}{c|}{\textbf{1.000($\approx$)}} & \textbf{1.000} & \multicolumn{1}{c|}{\textbf{1.000($\approx$)}} & \textbf{1.000} & \multicolumn{1}{c|}{\textbf{1.000($\approx$)}} & \textbf{1.000} & \multicolumn{1}{c|}{\textbf{1.000($\approx$)}} & \textbf{1.000} & \multicolumn{1}{c|}{\textbf{1.000($\approx$)}} & \textbf{1.000} & \multicolumn{1}{c|}{\textbf{1.000($\approx$)}} & \textbf{1.000} & \multicolumn{1}{c|}{\textbf{1.000($\approx$)}} & \textbf{1.000} \\ \hline
\multicolumn{1}{|c|}{$F_{4}$} & \multicolumn{1}{c|}{\textbf{1.000($\approx$)}} & \textbf{1.000} & \multicolumn{1}{c|}{\textbf{1.000($\approx$)}} & \textbf{1.000} & \multicolumn{1}{c|}{\textbf{1.000($\approx$)}} & \textbf{1.000} & \multicolumn{1}{c|}{\textbf{1.000($\approx$)}} & \textbf{1.000} & \multicolumn{1}{c|}{\textbf{1.000($\approx$)}} & \textbf{1.000} & \multicolumn{1}{c|}{0.985($+$)} & 0.940 & \multicolumn{1}{c|}{\textbf{1.000($\approx$)}} & \textbf{1.000} \\ \hline
\multicolumn{1}{|c|}{$F_{5}$} & \multicolumn{1}{c|}{\textbf{1.000($\approx$)}} & \textbf{1.000} & \multicolumn{1}{c|}{\textbf{1.000($\approx$)}} & \textbf{1.000} & \multicolumn{1}{c|}{\textbf{1.000($\approx$)}} & \textbf{1.000} & \multicolumn{1}{c|}{\textbf{1.000($\approx$)}} & \textbf{1.000} & \multicolumn{1}{c|}{\textbf{1.000($\approx$)}} & \textbf{1.000} & \multicolumn{1}{c|}{\textbf{1.000($\approx$)}} & \textbf{1.000} & \multicolumn{1}{c|}{\textbf{1.000($\approx$)}} & \textbf{1.000} \\ \hline
\multicolumn{1}{|c|}{$F_{6}$} & \multicolumn{1}{c|}{\textbf{1.000($\approx$)}} & \textbf{1.000} & \multicolumn{1}{c|}{\textbf{1.000($\approx$)}} & \textbf{1.000} & \multicolumn{1}{c|}{\textbf{1.000($\approx$)}} & \textbf{1.000} & \multicolumn{1}{c|}{\textbf{1.000($\approx$)}} & \textbf{1.000} & \multicolumn{1}{c|}{\textbf{1.000($\approx$)}} & \textbf{1.000} & \multicolumn{1}{c|}{\textbf{1.000($\approx$)}} & \textbf{1.000} & \multicolumn{1}{c|}{\textbf{1.000($\approx$)}} & \textbf{1.000} \\ \hline
\multicolumn{1}{|c|}{$F_{7}$} & \multicolumn{1}{c|}{0.838($+$)} & 0.039 & \multicolumn{1}{c|}{0.933($+$)} & 0.176 & \multicolumn{1}{c|}{0.913($+$)} & 0.078 & \multicolumn{1}{c|}{0.893($+$)} & 0.058 & \multicolumn{1}{c|}{0.921($+$)} & 0.040 & \multicolumn{1}{c|}{\textbf{1.000($\approx$)}} & \textbf{1.000} & \multicolumn{1}{c|}{\textbf{1.000($\approx$)}} & \textbf{1.000} \\ \hline
\multicolumn{1}{|c|}{$F_{8}$} & \multicolumn{1}{c|}{0.747($+$)} & 0.000 & \multicolumn{1}{c|}{0.944($+$)} & 0.078 & \multicolumn{1}{c|}{0.965($+$)} & 0.118 & \multicolumn{1}{c|}{0.984($+$)} & 0.784 & \multicolumn{1}{c|}{0.692($+$)} & 0.000 & \multicolumn{1}{c|}{\textbf{1.000($\approx$)}} & \textbf{1.000} & \multicolumn{1}{c|}{\textbf{1.000($\approx$)}} & \textbf{1.000} \\ \hline
\multicolumn{1}{|c|}{$F_{9}$} & \multicolumn{1}{c|}{0.384($+$)} & 0.000 & \multicolumn{1}{c|}{0.512($-$)} & 0.000 & \multicolumn{1}{c|}{0.572($-$)} & 0.000 & \multicolumn{1}{c|}{0.385($+$)} & 0.000 & \multicolumn{1}{c|}{0.571($-$)} & 0.000 & \multicolumn{1}{c|}{\textbf{1.000($-$)}} & \textbf{0.920} & \multicolumn{1}{c|}{0.950($-$)} & 0.000 \\ \hline
\multicolumn{1}{|c|}{$F_{10}$} & \multicolumn{1}{c|}{\textbf{1.000($\approx$)}} & \textbf{1.000} & \multicolumn{1}{c|}{\textbf{1.000($\approx$)}} & \textbf{1.000} & \multicolumn{1}{c|}{\textbf{1.000($\approx$)}} & \textbf{1.000} & \multicolumn{1}{c|}{\textbf{1.000($\approx$)}} & \textbf{1.000} & \multicolumn{1}{c|}{\textbf{1.000($\approx$)}} & \textbf{1.000} & \multicolumn{1}{c|}{\textbf{1.000($\approx$)}} & \textbf{1.000} & \multicolumn{1}{c|}{\textbf{1.000($\approx$)}} & \textbf{1.000} \\ \hline
\multicolumn{1}{|c|}{$F_{11}$} & \multicolumn{1}{c|}{\textbf{1.000($\approx$)}} & \textbf{1.000} & \multicolumn{1}{c|}{\textbf{1.000($\approx$)}} & \textbf{1.000} & \multicolumn{1}{c|}{\textbf{1.000($\approx$)}} & \textbf{1.000} & \multicolumn{1}{c|}{0.981($+$)} & 0.818 & \multicolumn{1}{c|}{\textbf{1.000($\approx$)}} & \textbf{1.000} & \multicolumn{1}{c|}{0.710($+$)} & 0.040 & \multicolumn{1}{c|}{\textbf{1.000($\approx$)}} & \textbf{1.000} \\ \hline
\multicolumn{1}{|c|}{$F_{12}$} & \multicolumn{1}{c|}{\textbf{1.000($\approx$)}} & \textbf{1.000} & \multicolumn{1}{c|}{\textbf{1.000($\approx$)}} & \textbf{1.000} & \multicolumn{1}{c|}{0.993($+$)} & 0.941 & \multicolumn{1}{c|}{0.977($+$)} & 0.727 & \multicolumn{1}{c|}{\textbf{1.000($\approx$)}} & \textbf{1.000} & \multicolumn{1}{c|}{0.955($+$)} & 0.660 & \multicolumn{1}{c|}{\textbf{1.000($\approx$)}} & \textbf{1.000} \\ \hline
\multicolumn{1}{|c|}{$F_{13}$} & \multicolumn{1}{c|}{0.686($+$)} & 0.000 & \multicolumn{1}{c|}{0.686($+$)} & 0.000 & \multicolumn{1}{c|}{0.902($+$)} & 0.471 & \multicolumn{1}{c|}{0.774($+$)} & 0.000 & \multicolumn{1}{c|}{0.987($+$)} & 0.920 & \multicolumn{1}{c|}{0.667($+$)} & 0.000 & \multicolumn{1}{c|}{0.997($+$)} & 0.980 \\ \hline
\multicolumn{1}{|c|}{$F_{14}$} & \multicolumn{1}{c|}{0.667($+$)} & 0.000 & \multicolumn{1}{c|}{0.667($+$)} & 0.000 & \multicolumn{1}{c|}{0.683($+$)} & 0.000 & \multicolumn{1}{c|}{0.667($+$)} & 0.000 & \multicolumn{1}{c|}{0.773($+$)} & 0.020 & \multicolumn{1}{c|}{0.667($+$)} & 0.000 & \multicolumn{1}{c|}{0.733($+$)} & 0.060 \\ \hline
\multicolumn{1}{|c|}{$F_{15}$} & \multicolumn{1}{c|}{0.637($+$)} & 0.000 & \multicolumn{1}{c|}{0.632($+$)} & 0.000 & \multicolumn{1}{c|}{0.654($+$)} & 0.000 & \multicolumn{1}{c|}{\textbf{0.750($-$)}} & 0.000 & \multicolumn{1}{c|}{0.748($-$)} & 0.000 & \multicolumn{1}{c|}{0.618($+$)} & 0.000 & \multicolumn{1}{c|}{0.595($+$)} & 0.000 \\ \hline
\multicolumn{1}{|c|}{$F_{16}$} & \multicolumn{1}{c|}{0.667($+$)} & 0.000 & \multicolumn{1}{c|}{0.667($+$)} & 0.000 & \multicolumn{1}{c|}{0.667($+$)} & 0.000 & \multicolumn{1}{c|}{0.667($+$)} & 0.000 & \multicolumn{1}{c|}{0.667($+$)} & 0.000 & \multicolumn{1}{c|}{0.630($+$)} & 0.000 & \multicolumn{1}{c|}{0.657($+$)} & 0.000 \\ \hline
\multicolumn{1}{|c|}{$F_{17}$} & \multicolumn{1}{c|}{0.375($+$)} & 0.000 & \multicolumn{1}{c|}{0.397($+$)} & 0.000 & \multicolumn{1}{c|}{0.522($+$)} & 0.000 & \multicolumn{1}{c|}{0.545($+$)} & 0.000 & \multicolumn{1}{c|}{0.593($+$)} & 0.000 & \multicolumn{1}{c|}{0.505($+$)} & 0.000 & \multicolumn{1}{c|}{0.335($+$)} & 0.000 \\ \hline
\multicolumn{1}{|c|}{$F_{18}$} & \multicolumn{1}{c|}{0.654($+$)} & 0.000 & \multicolumn{1}{c|}{0.654($+$)} & 0.000 & \multicolumn{1}{c|}{\textbf{0.667($\approx$)}} & 0.000 & \multicolumn{1}{c|}{\textbf{0.667($\approx$)}} & 0.000 & \multicolumn{1}{c|}{\textbf{0.667($\approx$)}} & 0.000 & \multicolumn{1}{c|}{0.497($+$)} & 0.000 & \multicolumn{1}{c|}{0.327($+$)} & 0.000 \\ \hline
\multicolumn{1}{|c|}{$F_{19}$} & \multicolumn{1}{c|}{0.375($\approx$)} & 0.000 & \multicolumn{1}{c|}{0.363($+$)} & 0.000 & \multicolumn{1}{c|}{0.505($-$)} & 0.000 & \multicolumn{1}{c|}{0.505($-$)} & 0.000 & \multicolumn{1}{c|}{0.543($-$)} & 0.000 & \multicolumn{1}{c|}{0.230($+$)} & 0.000 & \multicolumn{1}{c|}{0.135($+$)} & 0.000 \\ \hline
\multicolumn{1}{|c|}{$F_{20}$} & \multicolumn{1}{c|}{0.250($+$)} & 0.000 & \multicolumn{1}{c|}{0.248($+$)} & 0.000 & \multicolumn{1}{c|}{0.255($+$)} & 0.000 & \multicolumn{1}{c|}{0.302($+$)} & 0.000 & \multicolumn{1}{c|}{0.355($+$)} & 0.000 & \multicolumn{1}{c|}{0.125($+$)} & 0.000 & \multicolumn{1}{c|}{0.080($+$)} & 0.000 \\ \hline
\multicolumn{1}{|c|}{$+$} & \multicolumn{2}{c|}{10} & \multicolumn{2}{c|}{10} & \multicolumn{2}{c|}{9} & \multicolumn{2}{c|}{10} & \multicolumn{2}{c|}{7} & \multicolumn{2}{c|}{11} & \multicolumn{2}{c|}{8} \\ \hline
\multicolumn{1}{|c|}{$\approx$} & \multicolumn{2}{c|}{10} & \multicolumn{2}{c|}{9} & \multicolumn{2}{c|}{9} & \multicolumn{2}{c|}{8} & \multicolumn{2}{c|}{10} & \multicolumn{2}{c|}{8} & \multicolumn{2}{c|}{11} \\ \hline
\multicolumn{1}{|c|}{$-$} & \multicolumn{2}{c|}{0} & \multicolumn{2}{c|}{1} & \multicolumn{2}{c|}{2} & \multicolumn{2}{c|}{2} & \multicolumn{2}{c|}{3} & \multicolumn{2}{c|}{1} & \multicolumn{2}{c|}{1} \\ \hline
\end{tabular}%
}
\end{table*}

\subsection{Comparison With State-of-the-Art Algorithms} \label{sec:comparison}

Table~\ref{tab:comparison_4} reports the experimental results of PR and SR on the CEC'2013 benchmark set with accuracy level $\varepsilon = \expnumber{1.0}{-04}$.
The values are the best retrieved by MGP-BBBC and the other state-of-the-art algorithms -- best PR and SR for each function are bolded. MGP-BBBC features the highest PR on thirteen out of twenty functions, non-significantly different than the algorithms retrieving the highest for two functions, and competitively on the other six functions. 

MGP-BBBC features a performance of both PF and SR $=1.000$ in the functions $F_1-F_6$, $F_8$, $F_{10}-F_{13}$ -- no other algorithm performs a perfect PR and SR past this point. The performance on $F_7$ (Vincent 2D), which features a sine function with a decreasing frequency, is also high: MGP-BBBC features the highest PR ($0.998$) and SR ($0.960$) among nine over thirteen algorithms, except for RS-CMSA-EAII, HillVallEA19, MOMMOP, and EMO-MMO (SR=$1.000$). In its three-dimensional counterpart $F_9$ (Vincent 3D), however, MGP-BBBC features a lower PR ($0.478$) than most of the other algorithms -- this is possibly due to the high number of global peaks ($216$) and the respective MGP-BBBC's requirement of a large population because of its cluster-based nature.
Remarkably, MGP-BBBC features the best PR $=0.930$ among every algorithm on $F_{14}$ (a 3D composite function), competing only with HillVallEA19, which features a significantly lower PR.
In the low-dimensionality composite functions $F_{15}$ (3D) and $F_{17}$ (5D), MGP-BBBC features a PR $=0.720$ and $=0.598$, respectively; $F_{15}$ is significantly worse than five out of thirteen algorithms, whereas $F_{17}$ only two. Interestingly, in the 5D composite function $F_{16}$, MGP-BBBC performs similarly ($0.707$) than its simpler counterpart k-BBBC ($0.710$), which scores the third highest PR followed by  HillVallEA19 ($0.723$) and first best RS-CMSA-EAII ($0.883$): this indicates that BBBC algorithm is a good metaheuristic for this function.
In the 10D composite functions $F_{18}-F_{19}$, MGP-BBBC performs inconsistently: on $F_{18}$ with PR $=0.667$, it ranks the best and non-significantly different than RS-CMSA-EAII, LBP-ADE, and HillVallEA19, NCD-DE, CFNDE, and DIDE (demonstrating MGP-BBBC's ability to handle high dimensionality); however, it ranks far from the best algorithms on $F_{19}$ with PR $=0.355$ against RS-CMSA-EAII ($0.703$) or HillVallEA19 ($0.593$).
Lastly, on the 20D function $F_{20}$, MGP-BBBC features a PR value of $0.358$, far from the best $0.618$ of RS-CMSA-EAII but still outperforming eleven over thirteen algorithms on this high-dimensional function. 

At the bottom of each column, we reported the numbers of functions for which MGP-BBBC performed significantly better ($+$), worse ($-$), or not significantly different ($\approx$) according to the Wilcoxon rank-sum test. These numbers show that, in terms of PR, MGP-BBBC performs competitively compared to all the state-of-the-art algorithms (i.e., the number of $+$ and $\approx$ is higher than the number of $-$) and performs better than eleven out of thirteen algorithms (i.e., the number of $+$ is higher than the number of $-$), with only RS-CMSA-EAII and HillVallEA19 outperforming it. Specifically, there are four out of thirteen cases in which the number of $+$ is higher than the sum of $-$ and $\approx$ (kBBBC, SSGA, LBP-ADE, and MOMMOP).


\section{Parameter Analysis} \label{sec:parameter_analysis}

MGP-BBBC features two user-defined parameters: the population size $n$ (same value for both offspring and archive) and the clustering kernel bandwidth $h$ (see Sec.~\ref{sec:big_crunch}). While the first one is a common parameter of EAs, the second is specific to the clustering algorithm and depends on the optimization problem (i.e., bounds, landscape, number of optima). 

In terms of space geometry, the kernel bandwidth defines the radius of a hypersphere (in a multidimensional space) within which points are considered neighbors. The choice of kernel bandwidth significantly impacts the clustering result. \textit{Smaller} bandwidths lead to tighter, more compact clusters and can result in more clusters, as points need to be closer to each other to be considered part of the same cluster. This makes the clustering more sensitive to noise and outliers, as small changes in position can lead to different cluster assignments. On the other hand, \textit{larger} bandwidths lead to more diffuse clusters, with the risk of merging two nearby clusters into a single cluster. However, this makes the clustering more robust to noise and outliers, as points farther apart can still be grouped together. Unfortunately, the relationship between kernel bandwidth and the size of the space is not linear. While the choice of bandwidth can be influenced by the scale of the data or the density of points, it is not directly proportional to the size of the space. Instead, it is related to the density of points and the desired level of granularity in the clustering result. This makes it essential to choose a bandwidth that is appropriate for the scale of the features in the dataset to achieve meaningful clustering results and requires us to select a proper value for each problem.

For a comprehensive parameter analysis, we first ran a preliminary study to observe which values to test on a large scale for both parameters. We observed that population sizes larger than 1000 individuals significantly reduce the performance of MGP-BBBC, except for problem $F_{17}$; however, this large number did not allow us to perform a full experiment due to high computational time: we set the tests for the population size to $n \in \{50, 100, 500, 1000\}$. For the kernel bandwidth, we observed different ranges for different problems: the first thirteen problems of CEC'2013 required small bandwidth $h<1.0$, whereas the more complex composite functions required a large bandwidth $h<18.00$. 

In this section, we present three parameter analyses to observe the effect of the clustering kernel bandwidth on MGP-BBBC. The first analysis (Sec.~\ref{sec:paramAnalysis_specific}) was a preliminary observation with specific ranges for each of the twenty problems in the CEC'2013 benchmark set. However, choosing different bandwidths for each problem (i) prevented us from performing a statistical analysis of the interactions between the population size and the bandwidth and (ii) made the parameter setting too dependent on the specific problem. Therefore, we conducted two more analyses that would allow us to set a unique range of bandwidth for all problems: the second analysis relates the bandwidth to the volume of the entire search space (Sec.~\ref{sec:paramAnalysis_volume}), and the third one relates the bandwidth to the real-time extent of expansion at each iteration (Sec.~\ref{sec:paramAnalysis_exploration}).
Once selected the range for each parameter, we choose a $4 \times 10$ set of combinations (i.e., four values for $n$ and ten for $h$). 
The best settings for each problem of CEC'2013, adopted in the comparison of Sec.~\ref{sec:comparison}, are reported in Table~\ref{tab:best_param_settings}: we kept both $n$ and $h$ consistent when multiple combinations yielded the same best performance.

\subsection{Specific Kernel Bandwidth Values For Each Function}
\label{sec:paramAnalysis_specific}

\begin{figure*}[h!]
    \centering
    \subfigure[\protect\url{}\label{fig:f01_h}$F_1$]
    {\includegraphics[height=3.5cm]{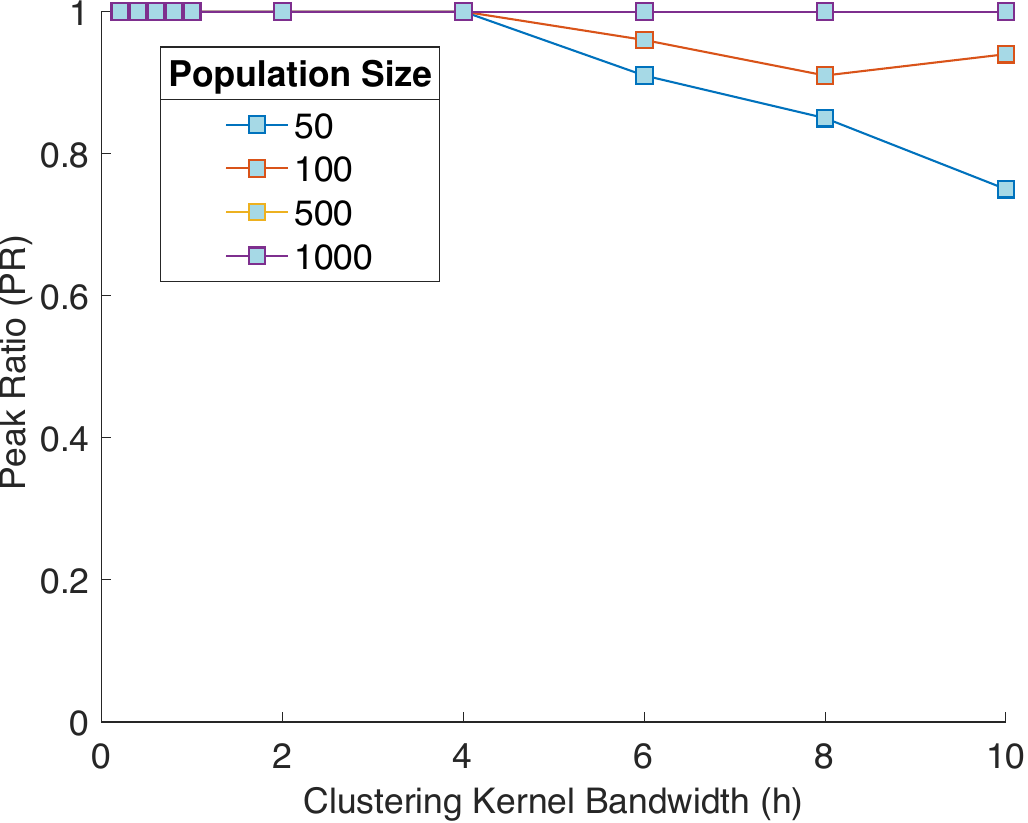}}
    \subfigure[\protect\url{}\label{fig:f02_h}$F_2$]
    {\includegraphics[height=3.5cm]{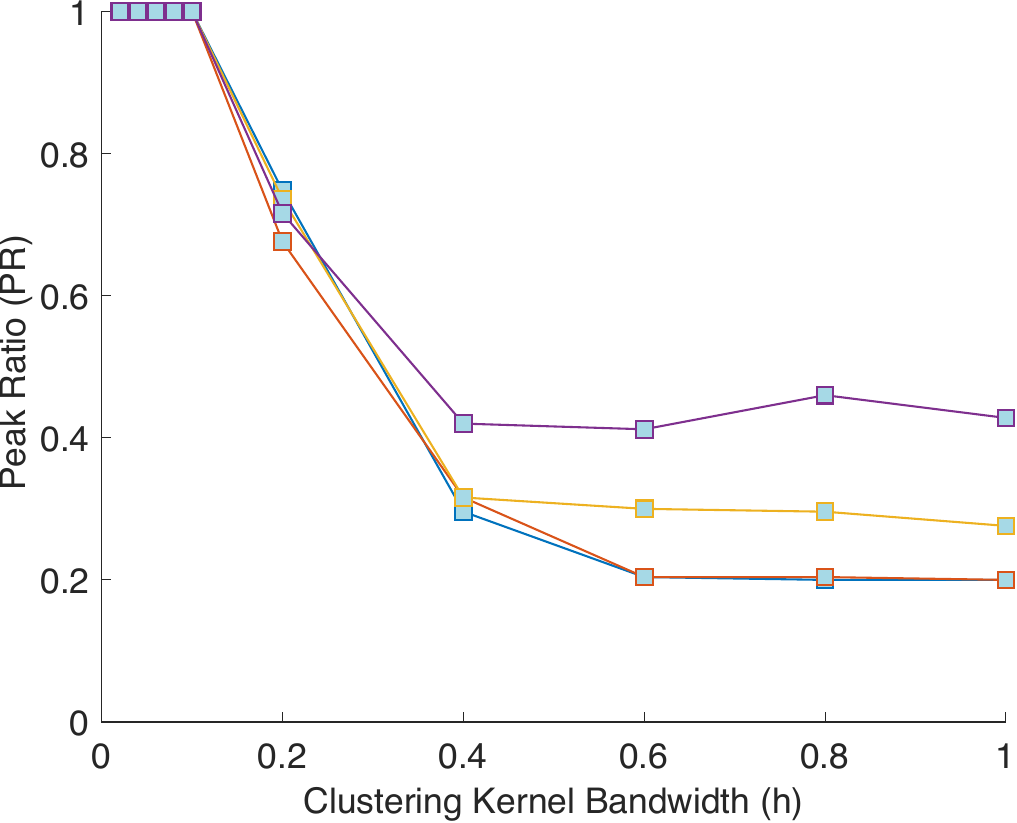}}
    \subfigure[\protect\url{}\label{fig:f03_h}$F_3$]
    {\includegraphics[height=3.5cm]{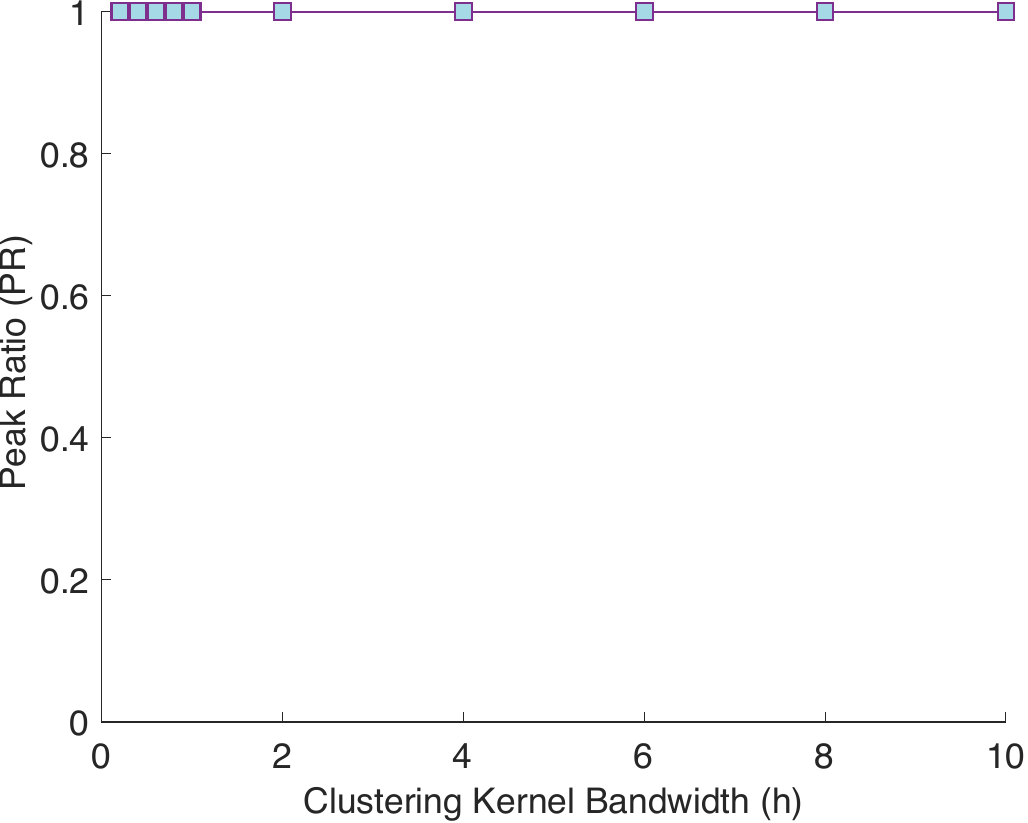}}
    \subfigure[\protect\url{}\label{fig:f04_h}$F_4$]
    {\includegraphics[height=3.5cm]{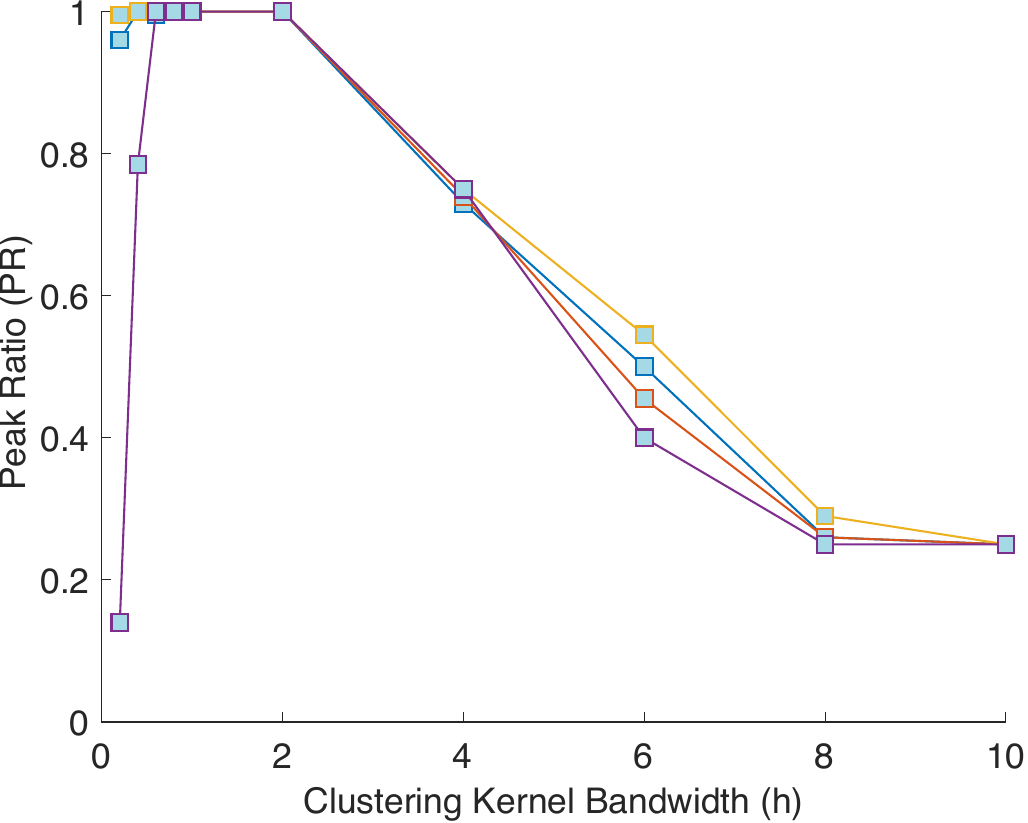}}

    \subfigure[\protect\url{}\label{fig:f05_h}$F_5$]
    {\includegraphics[height=3.5cm]{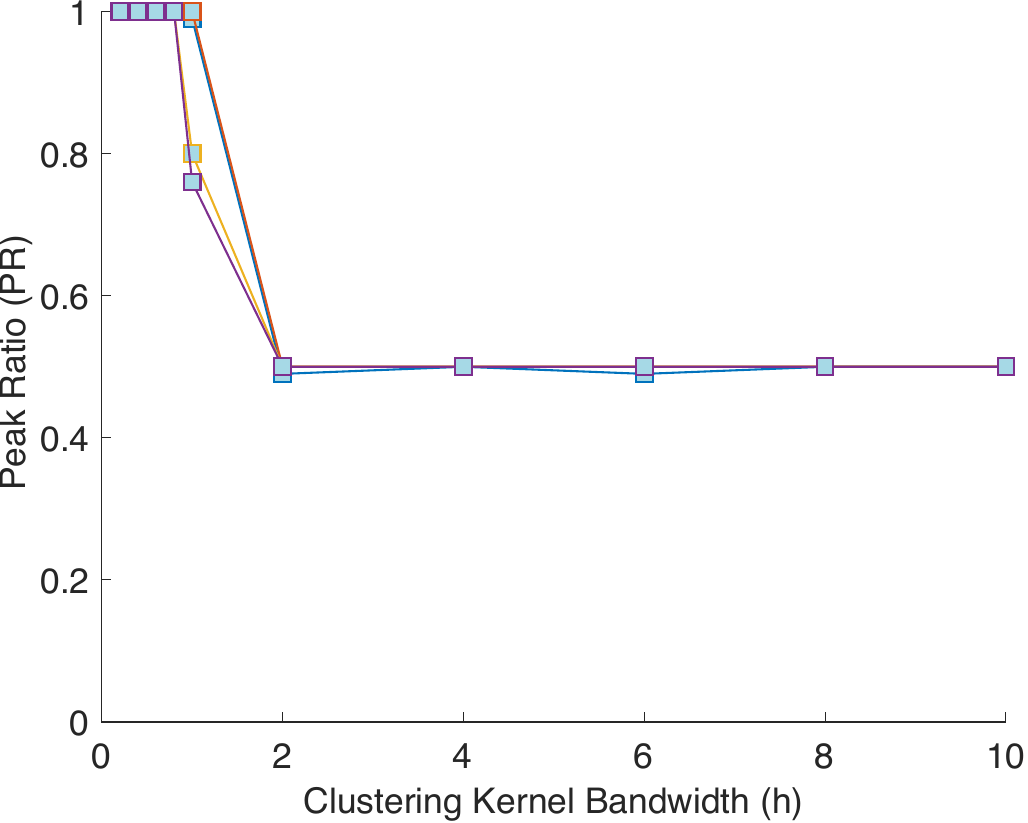}}
    \subfigure[\protect\url{}\label{fig:f06_h}$F_6$]
    {\includegraphics[height=3.5cm]{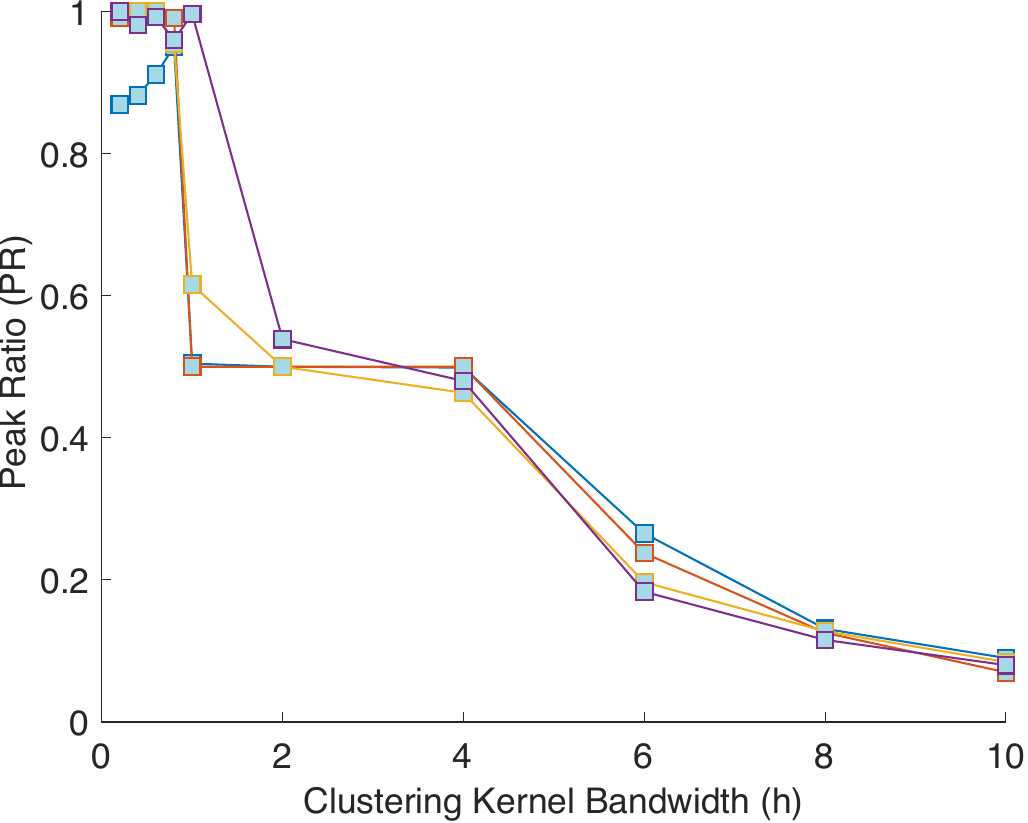}}
    \subfigure[\protect\url{}\label{fig:f07_h}$F_7$]
    {\includegraphics[height=3.5cm]{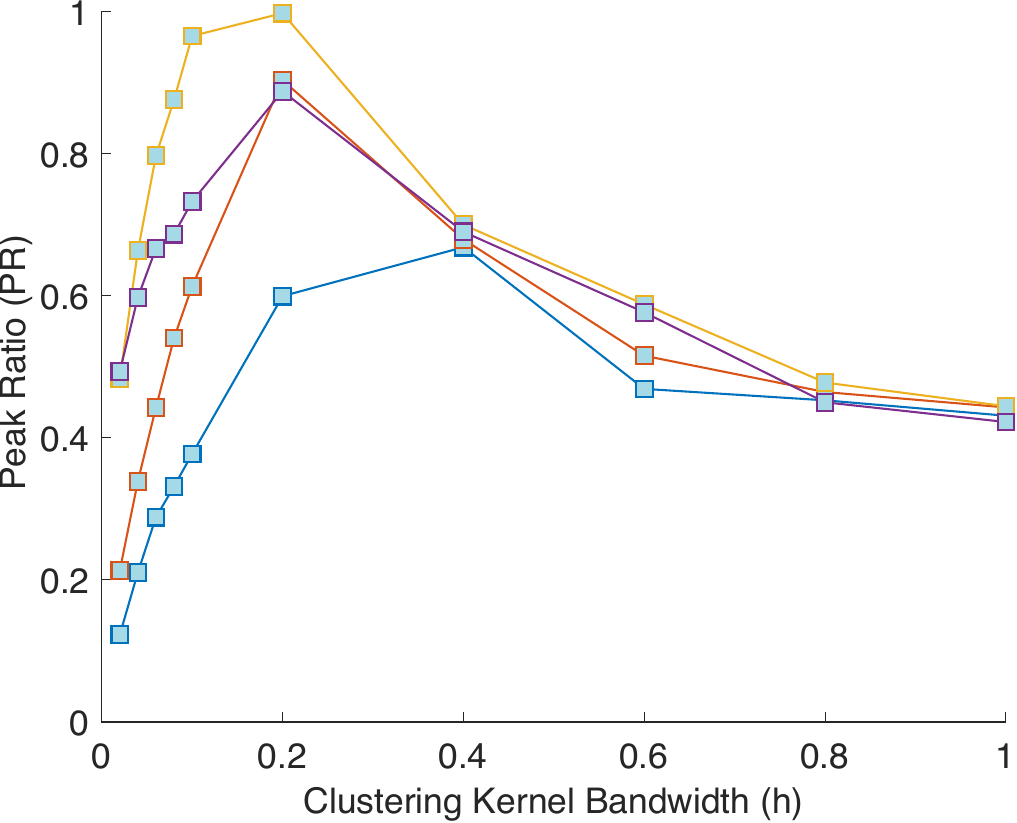}}
    \subfigure[\protect\url{}\label{fig:f08_h}$F_8$]
    {\includegraphics[height=3.5cm]{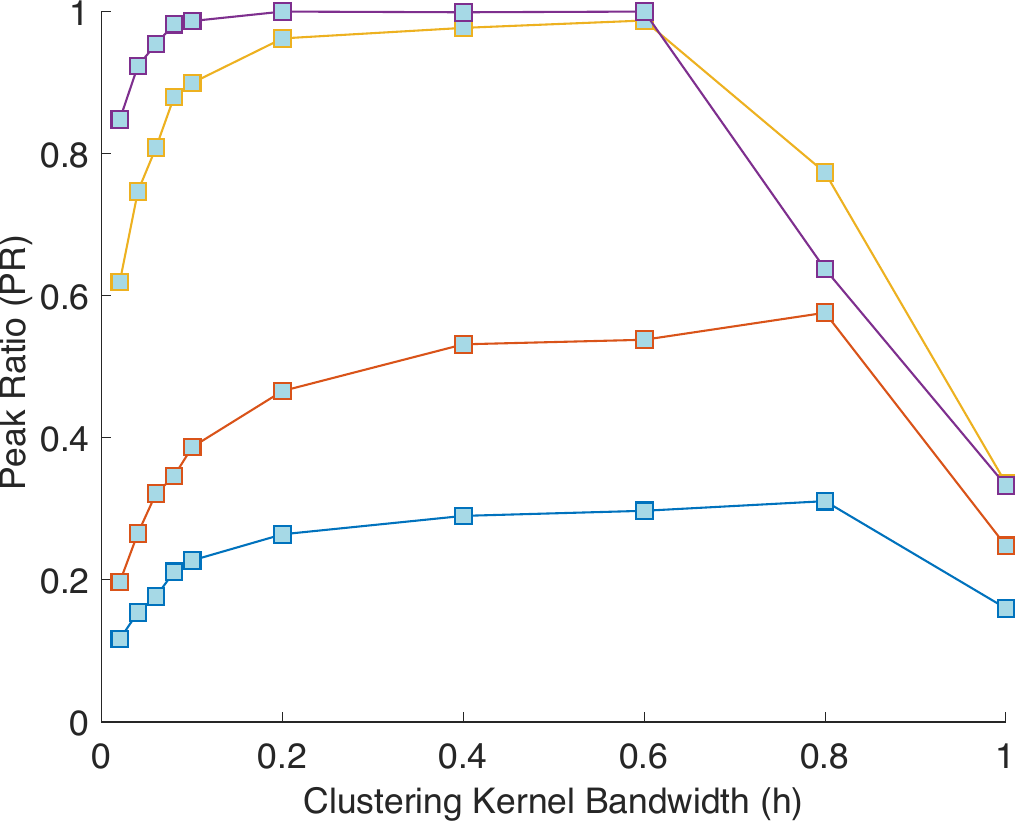}}

    \subfigure[\protect\url{}\label{fig:f09_h}$F_9$]
    {\includegraphics[height=3.5cm]{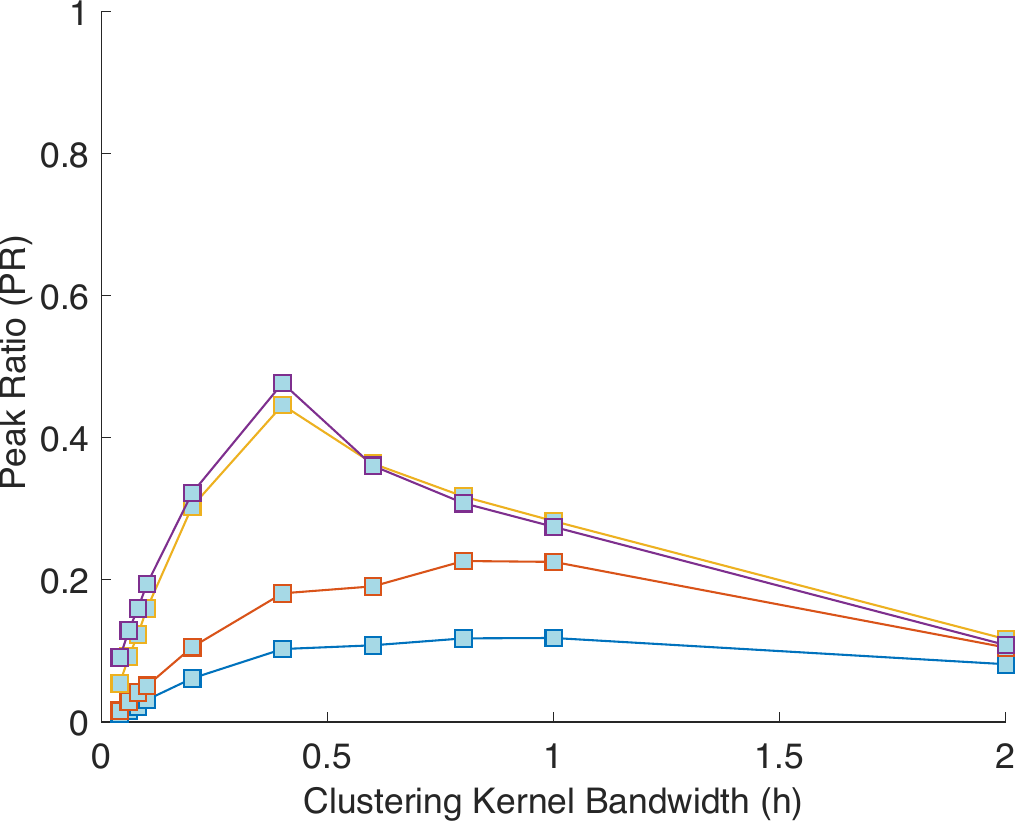}}
    \subfigure[\protect\url{}\label{fig:f10_h}$F_{10}$]
    {\includegraphics[height=3.5cm]{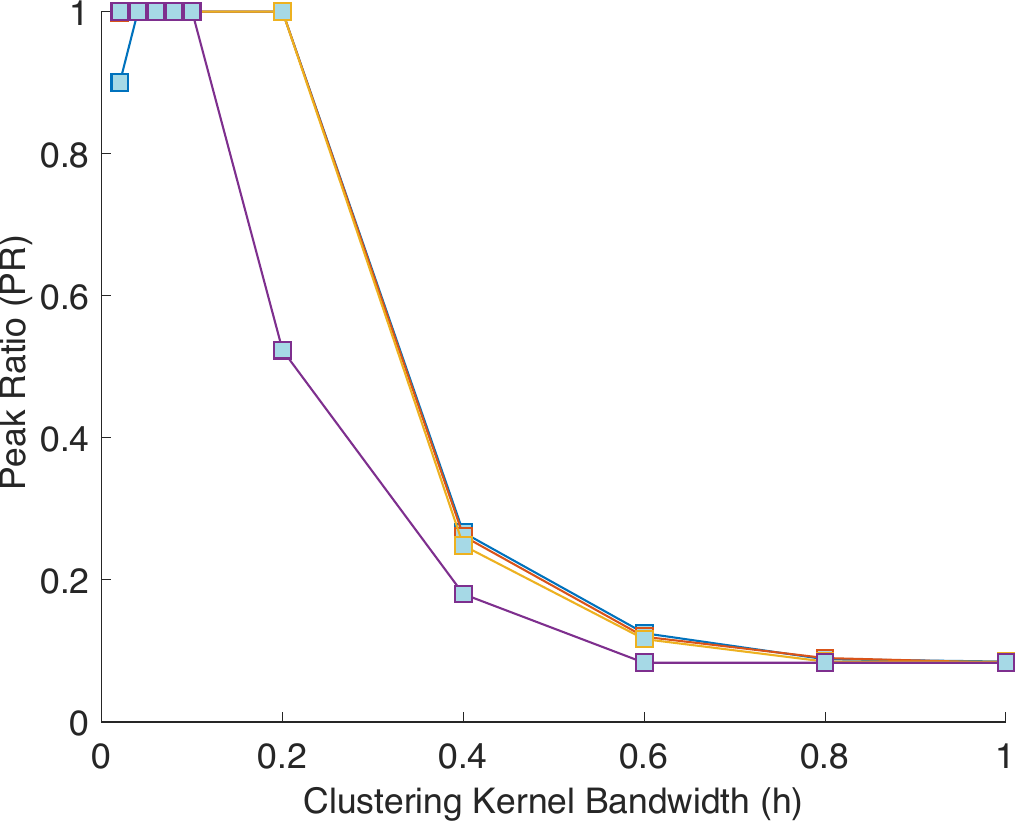}}
    \subfigure[\protect\url{}\label{fig:f11_h}$F_{11}$]
    {\includegraphics[height=3.5cm]{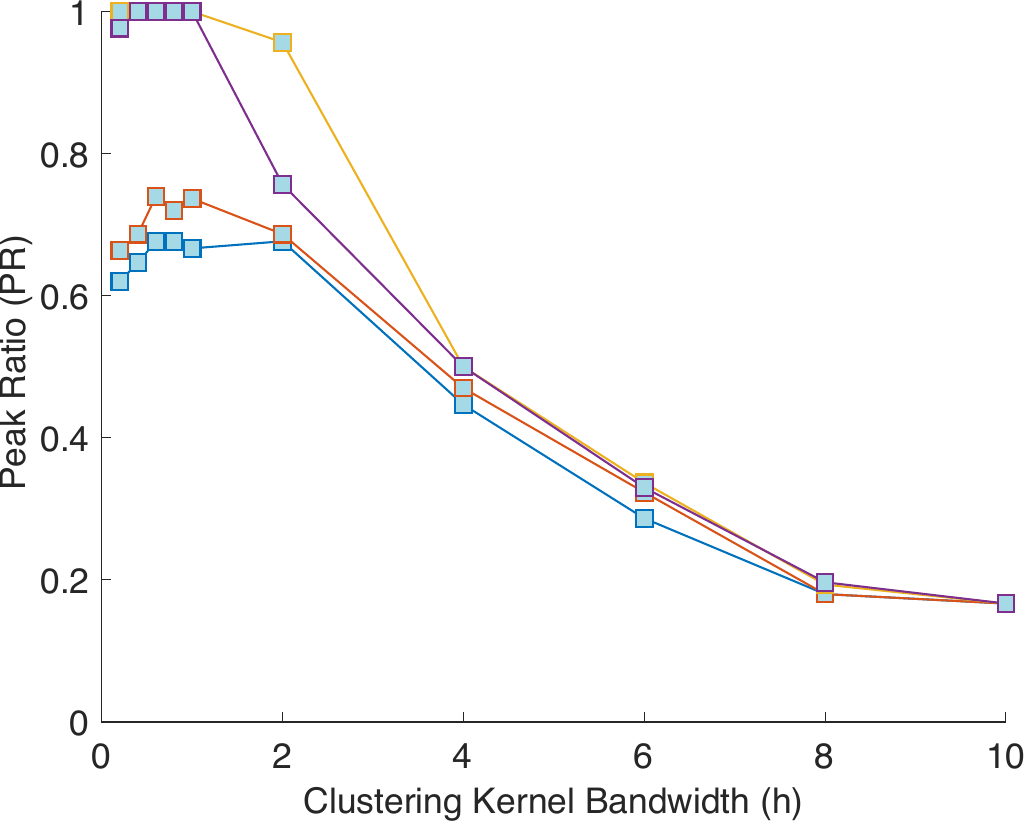}}
    \subfigure[\protect\url{}\label{fig:f12_h}$F_{12}$]
    {\includegraphics[height=3.5cm]{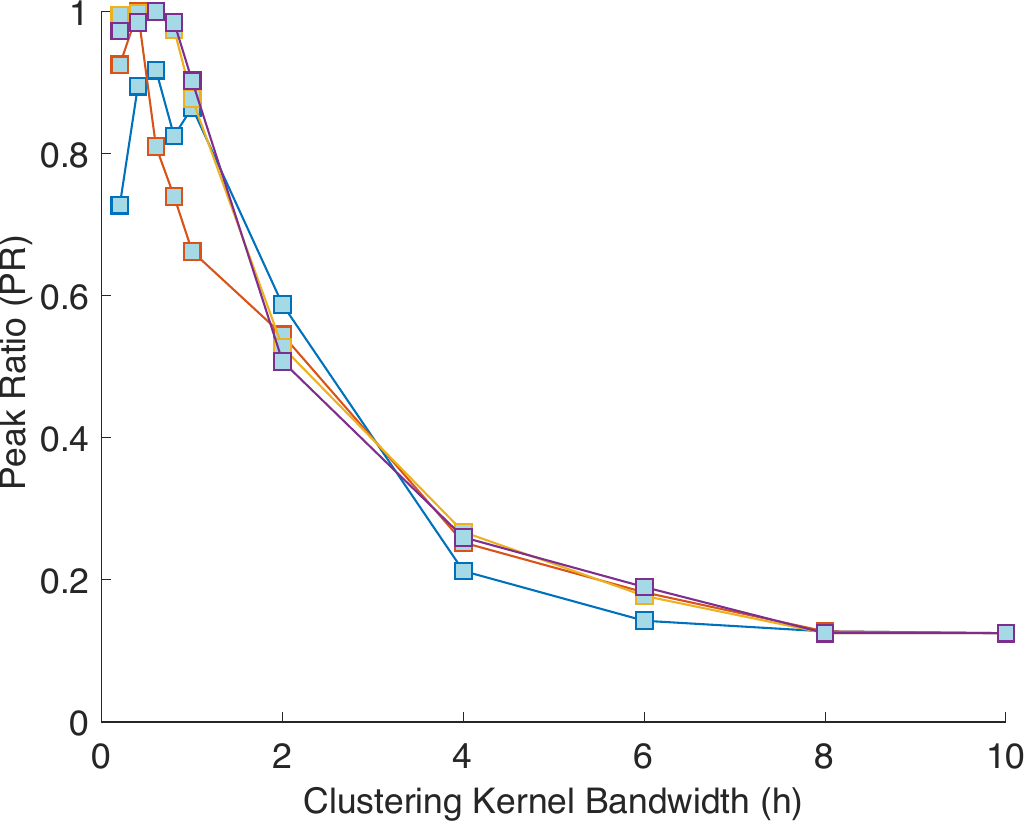}}

    \subfigure[\protect\url{}\label{fig:f13_h}$F_{13}$]
    {\includegraphics[height=3.5cm]{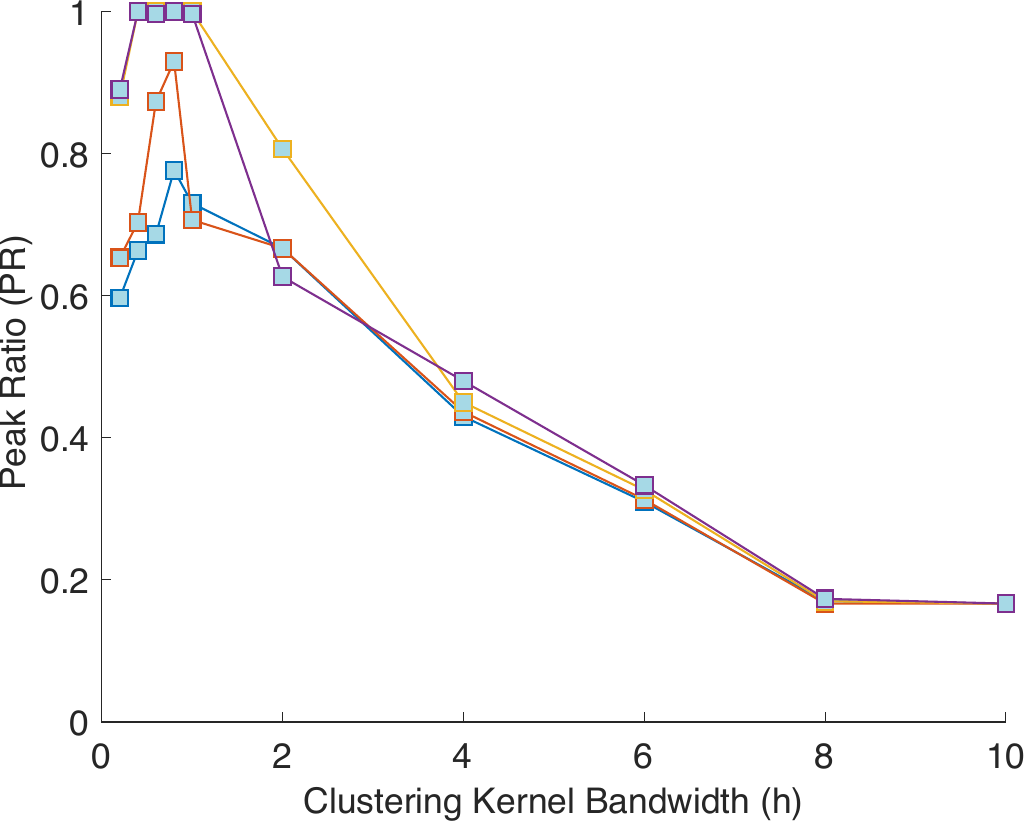}}
    \subfigure[\protect\url{}\label{fig:f14_h}$F_{14}$]
    {\includegraphics[height=3.5cm]{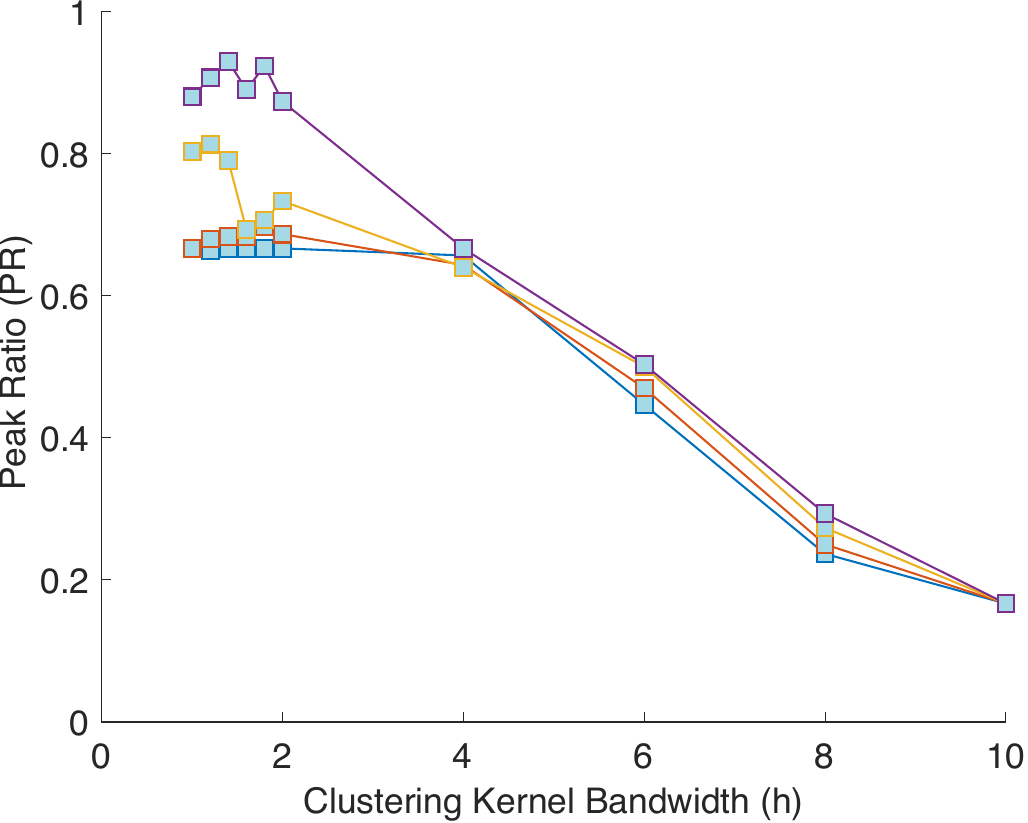}}
    \subfigure[\protect\url{}\label{fig:f15_h}$F_{15}$]
    {\includegraphics[height=3.5cm]{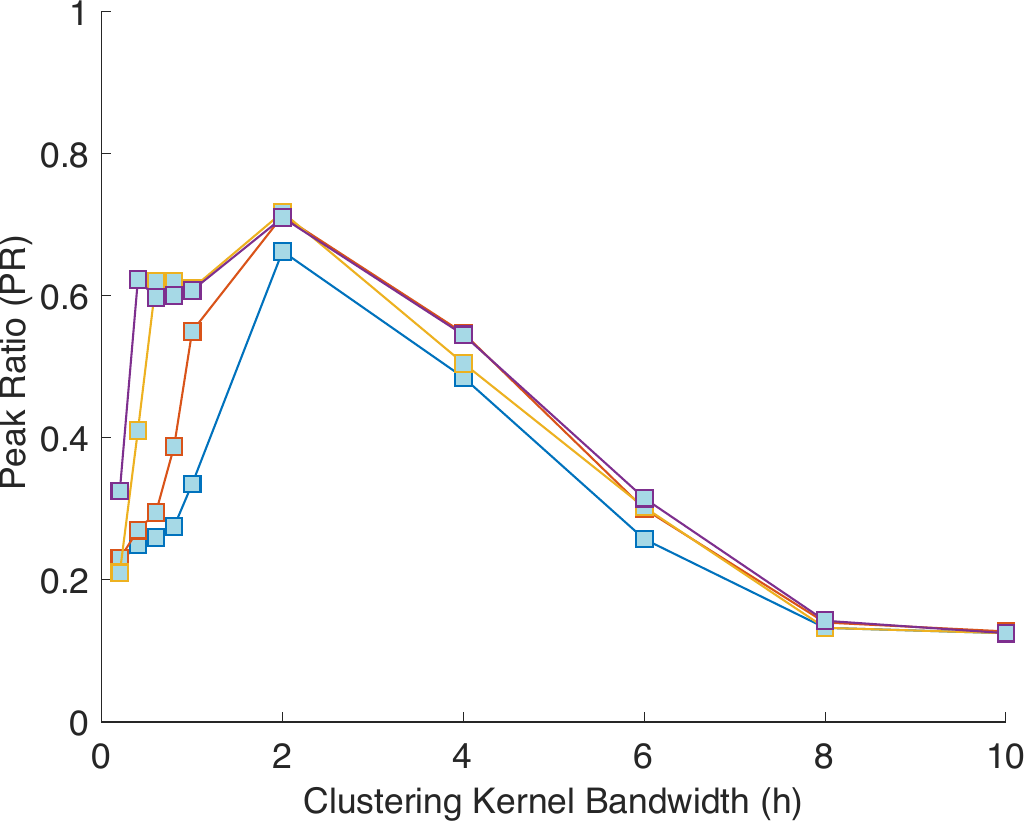}}
    \subfigure[\protect\url{}\label{fig:f16_h}$F_{16}$]
    {\includegraphics[height=3.5cm]{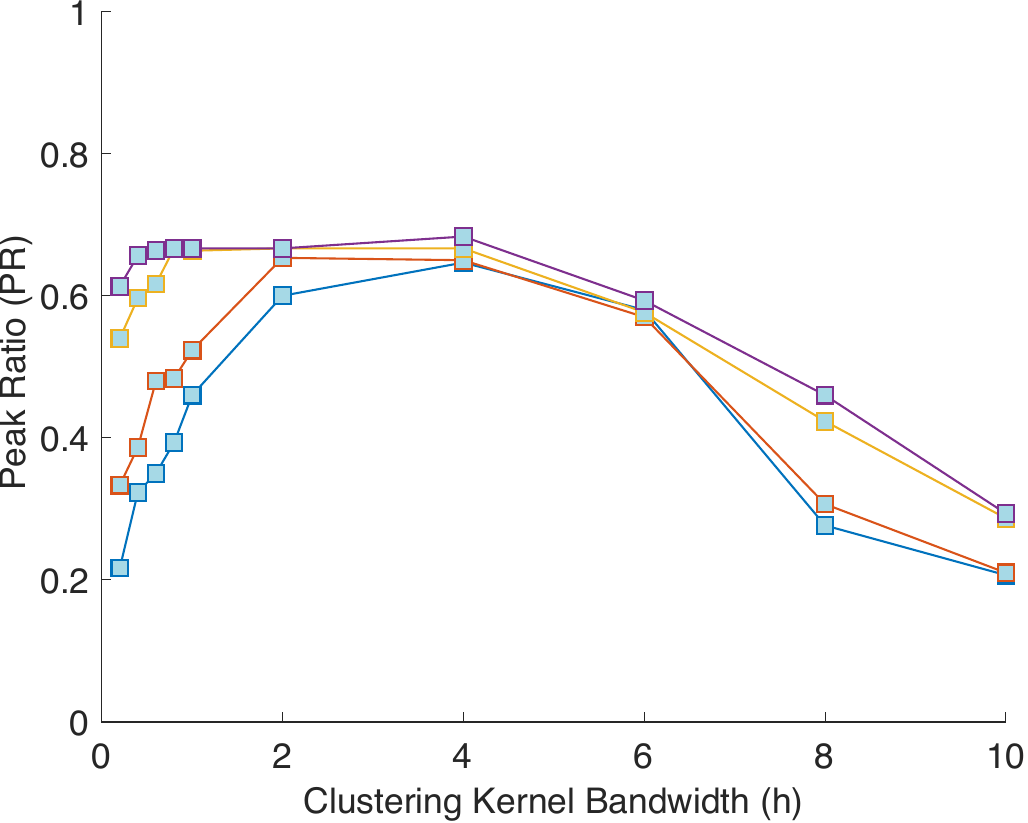}}

    \subfigure[\protect\url{}\label{fig:f17_h}$F_{17}$]
    {\includegraphics[height=3.5cm]{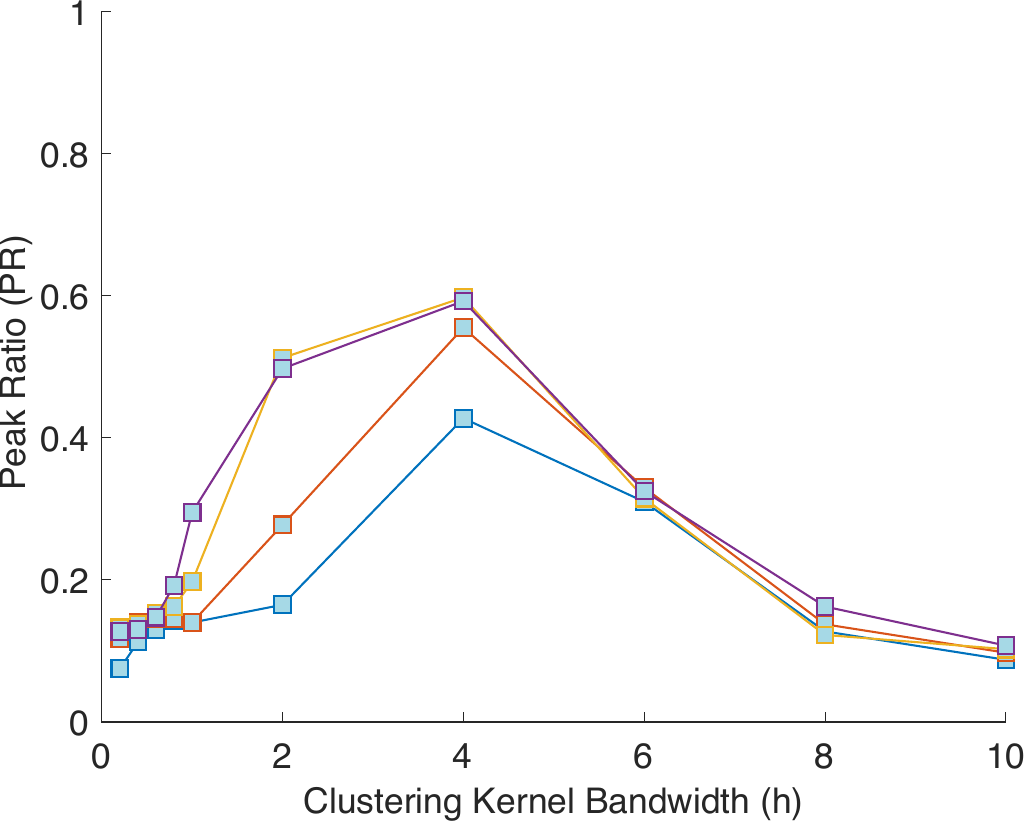}}
    \subfigure[\protect\url{}\label{fig:f18_h}$F_{18}$]
    {\includegraphics[height=3.5cm]{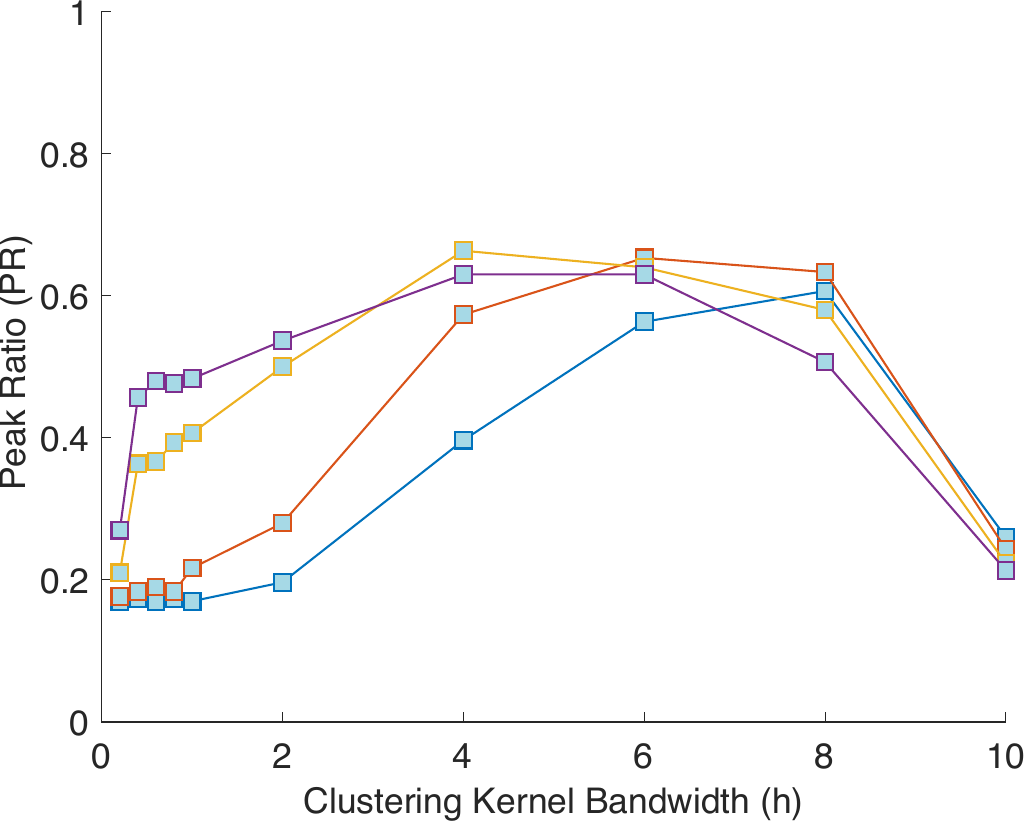}}
    \subfigure[\protect\url{}\label{fig:f19_h}$F_{19}$]
    {\includegraphics[height=3.5cm]{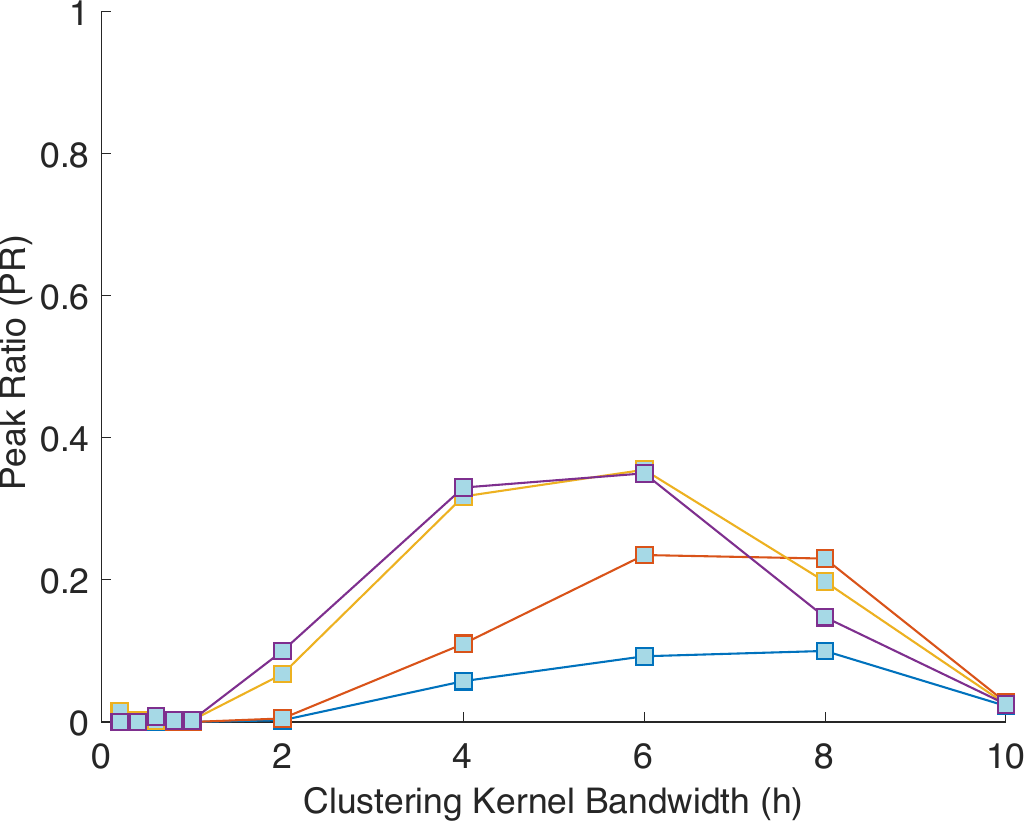}}
    \subfigure[\protect\url{}\label{fig:f20_h}$F_{20}$]
    {\includegraphics[height=3.5cm]{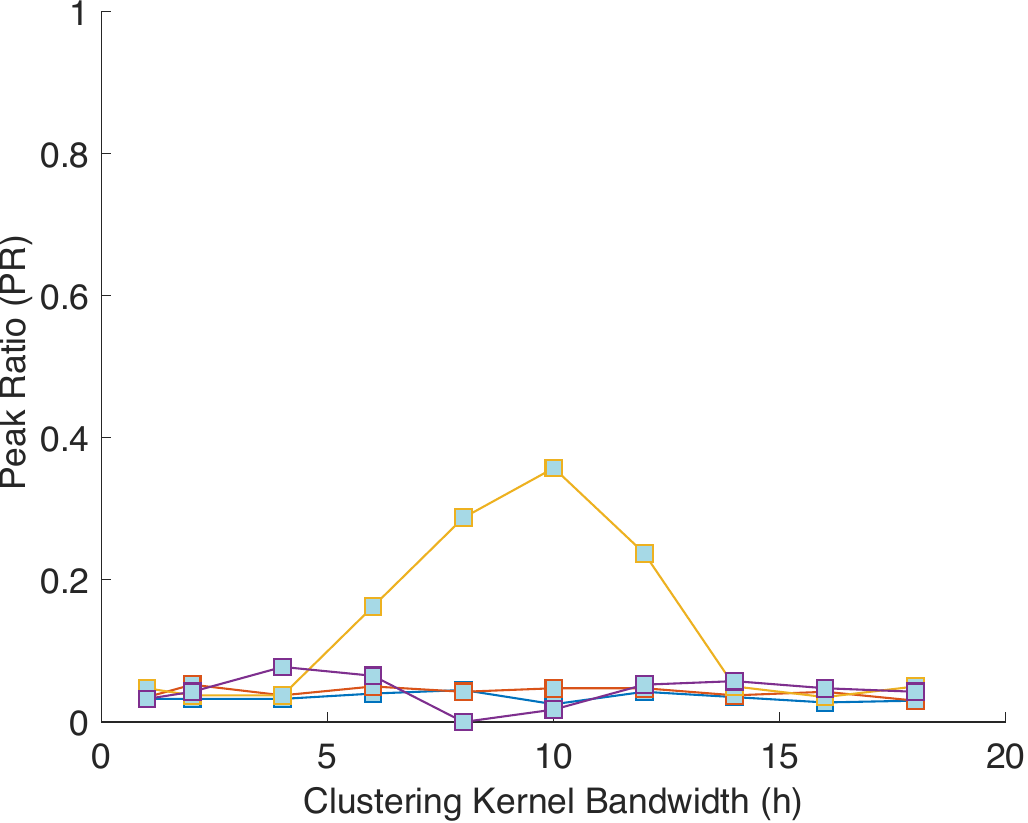}}
    
    \caption{Changes in the PR values (at the accuracy level $\varepsilon=1.0E-04$) with different Population Size (n) and Clustering Kernel Bandwidth (h) on CEC'2013 benchmark set.}
    \label{fig:pr_values}
\end{figure*}

To observe how the values of the parameters affect the results of MGP-BBBC, we compared the best result for each function against each combination with the Wilcoxon rank-sum test at the significance level $\alpha = 0.05$ (results are reported in the supplementary material under Tables S.I and S.II for accuracy $\varepsilon = 1.0E-03$, S.III and S.IV for $\varepsilon = 1.0E-04$, and S.V and S.VI for $\varepsilon = 1.0E-05$). The tables indicate the number of settings that performed significantly worse ($-$) or non-significantly different ($\approx$) than the best setting in terms of both population size (rows) and bandwidth (columns). The trend for each problem is graphically depicted in Fig.~\ref{fig:pr_values} for accuracy $\varepsilon = 1.0E-04$ only. 

Lastly, we adopted Friedman's test with the Bonferroni-Dunn procedure~\cite{demvsar2006statistical} to test the robustness of MGP-BBBC. Since the selected values of $h$ differ from problem to problem, Friedman’s test can analyze the results for solving each problem separately; however, considering that the population size was set to be the same for all problems, we ran the Friedman's test to analyze the effect of the population size on all twenty problems as well. Table~\ref{tab:bonferroni_pop} shows the results of the population size among all the problems, whereas Tables from S.VII to S.XXVI of the supplementary material show the results of both population size and bandwidth for each problem. The p-value shows the result of the Bonferroni-Dunn procedure; ``Y'' indicates that there is a significant difference (at the significance level $\alpha = 0.05$) between the corresponding pairwise settings, whereas ``N'' indicates no significant difference.

The results show that MGP-BBBC's performance improves as the population size increases; however, there are no significant differences between $n=500$ and $n=1000$ overall. Beyond this point, it is not worth investigating -- although there might be some good results for $n=800$. Fig.~\ref{fig:pr_values} also show similar trends for $n=500$ and $n=1000$, with the only exception of $F_{20}$. In this problem, $n=500$ is the only setting that provided good results, whereas the other population sizes show flat low trends; however, according to the Bonferroni-Dunn procedure (Tab. S.XXVI), there is no statistical difference between $n=500$ and $n=1000$ for $F_{20}$ -- this is possibly due to the large observation range of $h \in [2.0, 18.0]$. 

\begin{table*}[t!]
\centering
\caption{Experimental results of Friedman Test with Bonferroni-Dunn Procedure on Population Size ($n$) over all the functions of CEC'2013}
\label{tab:bonferroni_pop}
\resizebox{\textwidth}{!}{%
\begin{tabular}{|cc|cc|cc|cc|cc|cc|}
\hline
\multicolumn{2}{|c|}{\textbf{pair}} & \multicolumn{2}{c|}{\textbf{1.00E-03}} & \multicolumn{2}{c|}{\textbf{pair}} & \multicolumn{2}{c|}{\textbf{1.00E-04}} & \multicolumn{2}{c|}{\textbf{pair}} & \multicolumn{2}{c|}{\textbf{1.00E-05}} \\ \hline
\multicolumn{1}{|c|}{\textbf{n1}} & \textbf{n2} & \multicolumn{1}{c|}{\textbf{p-value}} & \textbf{$\alpha=0.05$} & \multicolumn{1}{c|}{\textbf{n1}} & \textbf{n2} & \multicolumn{1}{c|}{\textbf{p-value}} & \textbf{$\alpha=0.05$} & \multicolumn{1}{c|}{\textbf{n1}} & \textbf{n2} & \multicolumn{1}{c|}{\textbf{p-value}} & \textbf{$\alpha=0.05$} \\ \hline
\multicolumn{1}{|c|}{50.00} & 100.00 & \multicolumn{1}{c|}{0.003596} & Y & \multicolumn{1}{c|}{50.00} & 100.00 & \multicolumn{1}{c|}{0.002889} & Y & \multicolumn{1}{c|}{50.00} & 100.00 & \multicolumn{1}{c|}{0.001775} & Y \\ \hline
\multicolumn{1}{|c|}{50.00} & 500.00 & \multicolumn{1}{c|}{1.70E-23} & Y & \multicolumn{1}{c|}{50.00} & 500.00 & \multicolumn{1}{c|}{6.78E-22} & Y & \multicolumn{1}{c|}{50.00} & 500.00 & \multicolumn{1}{c|}{1.71E-20} & Y \\ \hline
\multicolumn{1}{|c|}{50.00} & 1000.00 & \multicolumn{1}{c|}{2.74E-25} & Y & \multicolumn{1}{c|}{50.00} & 1000.00 & \multicolumn{1}{c|}{2.18E-22} & Y & \multicolumn{1}{c|}{50.00} & 1000.00 & \multicolumn{1}{c|}{3.97E-20} & Y \\ \hline
\multicolumn{1}{|c|}{100.00} & 500.00 & \multicolumn{1}{c|}{9.94E-11} & Y & \multicolumn{1}{c|}{100.00} & 500.00 & \multicolumn{1}{c|}{1.69E-09} & Y & \multicolumn{1}{c|}{100.00} & 500.00 & \multicolumn{1}{c|}{2.96E-08} & Y \\ \hline
\multicolumn{1}{|c|}{100.00} & 1000.00 & \multicolumn{1}{c|}{6.11E-12} & Y & \multicolumn{1}{c|}{100.00} & 1000.00 & \multicolumn{1}{c|}{8.04E-10} & Y & \multicolumn{1}{c|}{100.00} & 1000.00 & \multicolumn{1}{c|}{5.03E-08} & Y \\ \hline
\multicolumn{1}{|c|}{500.00} & 1000.00 & \multicolumn{1}{c|}{1} & N & \multicolumn{1}{c|}{500.00} & 1000.00 & \multicolumn{1}{c|}{1} & N & \multicolumn{1}{c|}{500.00} & 1000.00 & \multicolumn{1}{c|}{1} & N \\ \hline
\multicolumn{2}{|c|}{Not Significantly   Different} & \multicolumn{2}{c|}{1} & \multicolumn{2}{c|}{Not Significantly Different} & \multicolumn{2}{c|}{1} & \multicolumn{2}{c|}{Not Significantly Different} & \multicolumn{2}{c|}{1} \\ \hline
\multicolumn{2}{|c|}{Significantly   Different} & \multicolumn{2}{c|}{5} & \multicolumn{2}{c|}{Significantly Different} & \multicolumn{2}{c|}{5} & \multicolumn{2}{c|}{Significantly Different} & \multicolumn{2}{c|}{5} \\ \hline
\end{tabular}%
}
\end{table*}

Regarding the clustering kernel bandwidth $h$, there is no significant impact on the performance of the problems $F_{1}-F_{5}$, $F_{10}$, and $F_{19}-F_{20}$, whereas for the other problems observe significant distance. From the trends shown in Fig.~\ref{fig:pr_values}, it is observable that smaller bandwidths are usually more promising than large bandwidths, resulting in a higher number of clusters and, therefore, centers of mass. However, determining the appropriate size for a ``small'' bandwidth varies depending on the specific problem. We observe that problems with high dimensionality require higher bandwidths; this can also depend on how far the peaks are in the space, which cannot be estimated without knowing the landscape. 

\subsection{Kernel Bandwidth Values to Volume of the Search Space}
\label{sec:paramAnalysis_volume}

To relate the kernel bandwidth to the size of a multidimensional search space, we can compare the volume of the bandwidth region (which is a hypersphere or hyperellipse) to the volume of the entire search space (which is a hypercube or hyperrectangle). We use the following definitions: 

\begin{itemize}
    \item $V_s$ as the volume of the entire search space (hyperrectangle or hypercube), calculated based on the bounds $[\mathbf{x}_i^L, \mathbf{x}_i^U]$ for each dimension $i \in 1...D$ as per Eqn.~(\ref{eq:volume_search_space}).

    \begin{equation}
        \label{eq:volume_search_space}
        \begin{aligned}
            V_s = \prod_{i=1}^{D}{\mathbf{x}_i^U - \mathbf{x}_i^L}
        \end{aligned}
    \end{equation}

    \item $V_h$ as the volume of the bandwidth region (hypersphere), which represents the region around each point within the kernel bandwidth, defined as in Eqn.~(\ref{eq:volume_bandwith}) where the bandwidth $h$ is the radius of the hypersphere, and $\Gamma$ is the Gamma function~\cite{philip1959leonhard}.

    \begin{equation}
        \label{eq:volume_bandwith}
        \begin{aligned}
            V_h = \frac{\pi^{\frac{D}{2}h^D}}{\Gamma\left( \frac{D}{2}+1\right)}
        \end{aligned}
    \end{equation}
    
\end{itemize}
To relate the bandwidth to the size of the search space, we computed the (inverted) ratio of the bandwidth region to the entire search space $\frac{V_s}{V_h}$. This ratio shows how much of the search space is covered by the bandwidth region around each point. The higher the ratio, the more significant the bandwidth is relative to the overall space.

We evaluated MGP-BBBC by setting the kernel bandwidth $h$ based on its ratio with the search space volume $V_s$, such that each function is tested on the same range of values -- i.e., the bandwidth is tuned based on the problem to solve. We set ten equispaced ratio values, from $2000$ to $10$, and applied the value of bandwidth retrieved with Eqn.~(\ref{eq:bandwidth_fromVolRatio}). These ratio values were chosen empirically based on the values of Tables S.I--S.VI in the range of values having non-significant differences against the best-retrieved PR. This analysis produced 400 observations: 4 values for population size (50, 100, 500, and 1000), 10 values for ratio, and 20 functions in the CEC'2013 benchmark set. 

\begin{equation}
    \label{eq:bandwidth_fromVolRatio}
    \begin{aligned}
        h = \left( \frac{\Gamma \left( \frac{D}{2} + 1 \right) \cdot \prod_{i=1}^{D}{\left(\mathbf{x}_i^U - \mathbf{x}_i^L\right)}}{\pi^{\frac{D}{2}} \cdot \text{ratio}} \right)^{\frac{1}{D}}
    \end{aligned}
\end{equation}

\begin{figure*}[h!]
    \centering
    \subfigure[\protect\url{}\label{fig:f01_r}$F_1$]
    {\includegraphics[height=3.5cm]{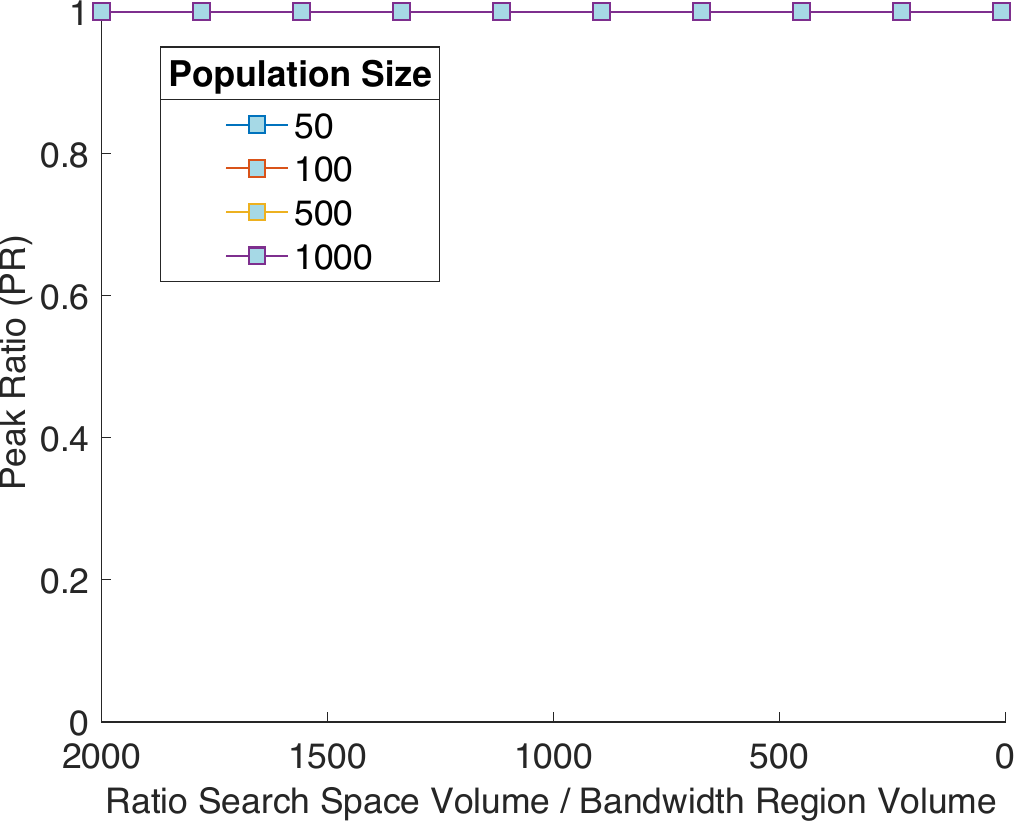}}
    \subfigure[\protect\url{}\label{fig:f02_r}$F_2$]
    {\includegraphics[height=3.5cm]{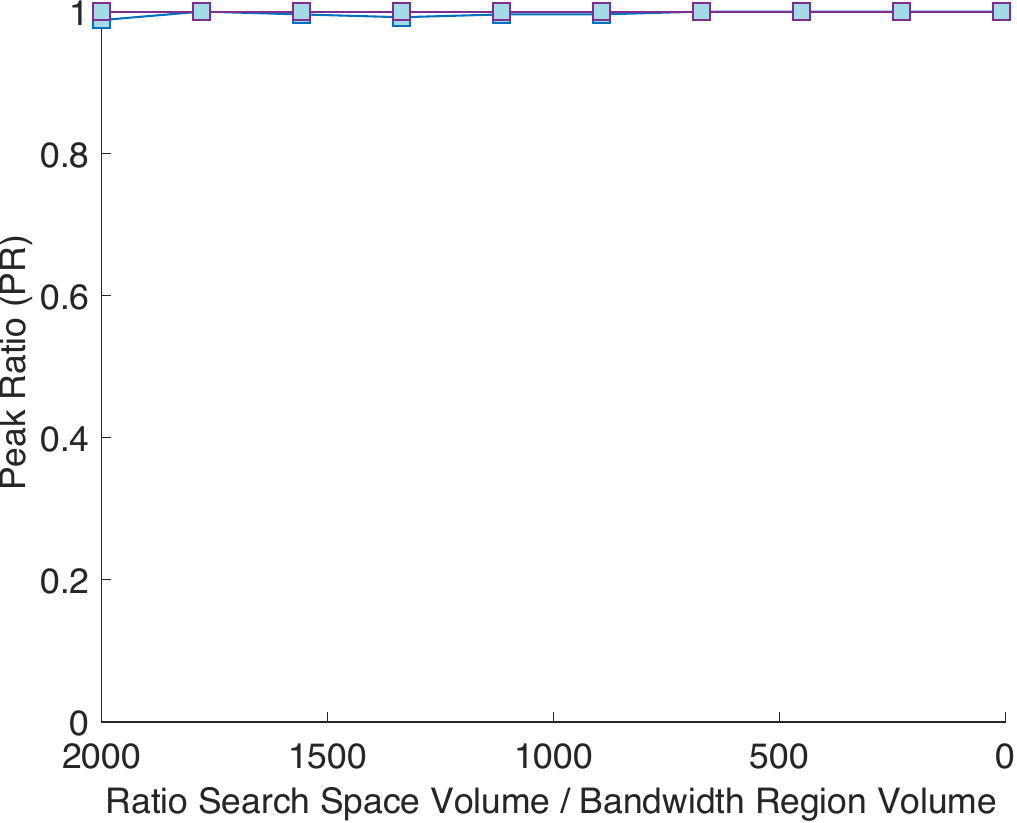}}
    \subfigure[\protect\url{}\label{fig:f03_r}$F_3$]
    {\includegraphics[height=3.5cm]{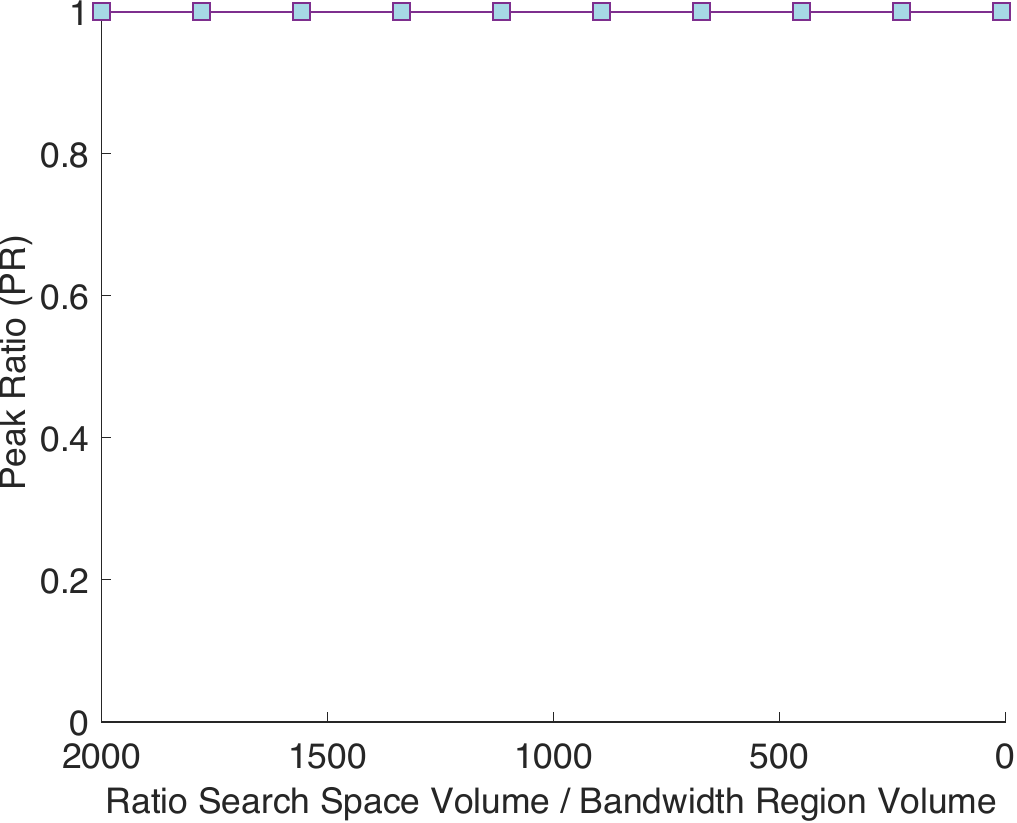}}
    \subfigure[\protect\url{}\label{fig:f04_r}$F_4$]
    {\includegraphics[height=3.5cm]{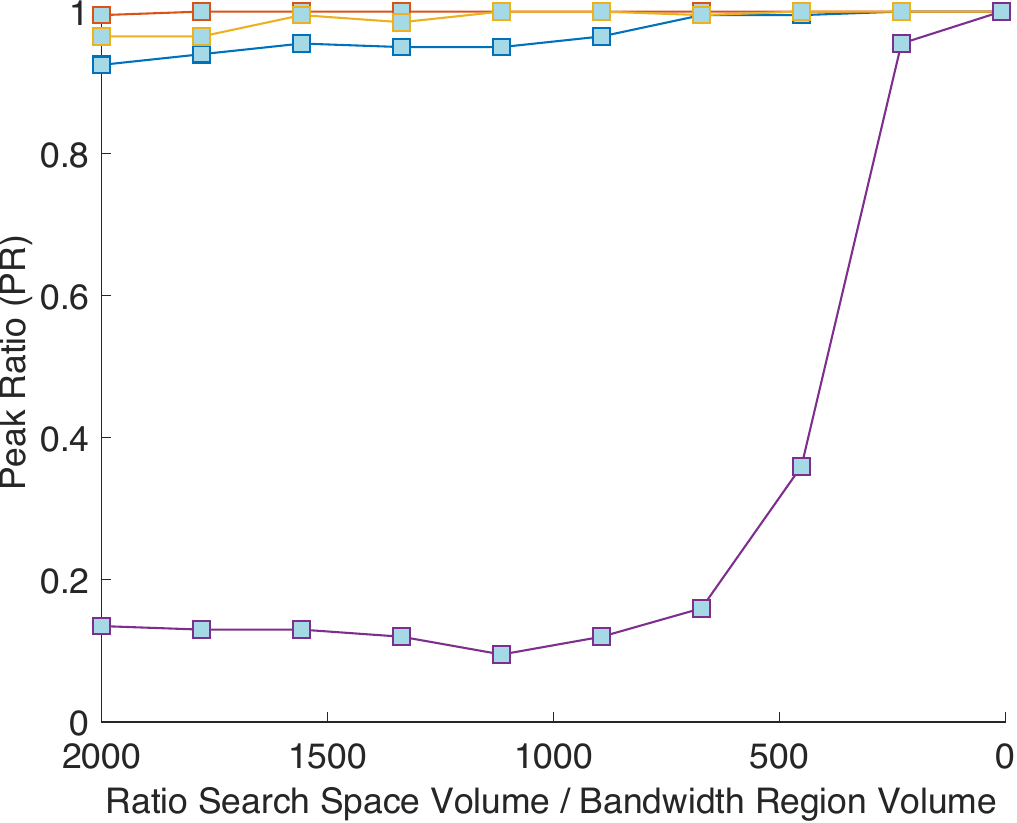}}

    \subfigure[\protect\url{}\label{fig:f05_r}$F_5$]
    {\includegraphics[height=3.5cm]{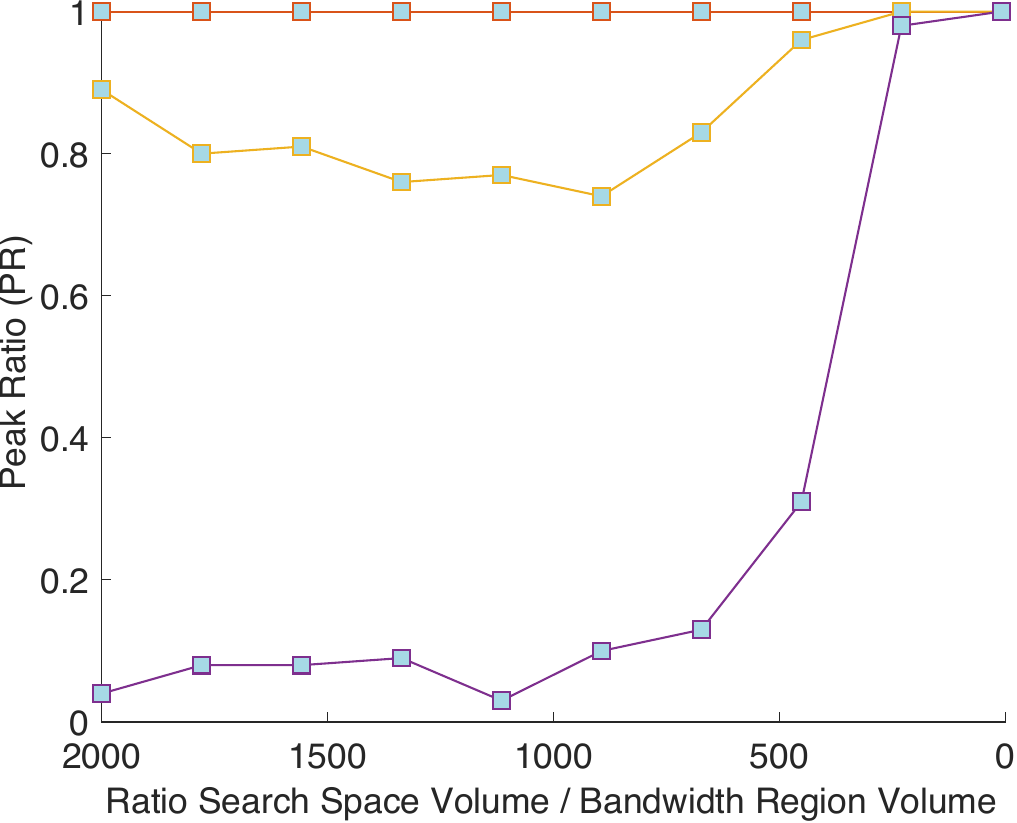}}
    \subfigure[\protect\url{}\label{fig:f06_r}$F_6$]
    {\includegraphics[height=3.5cm]{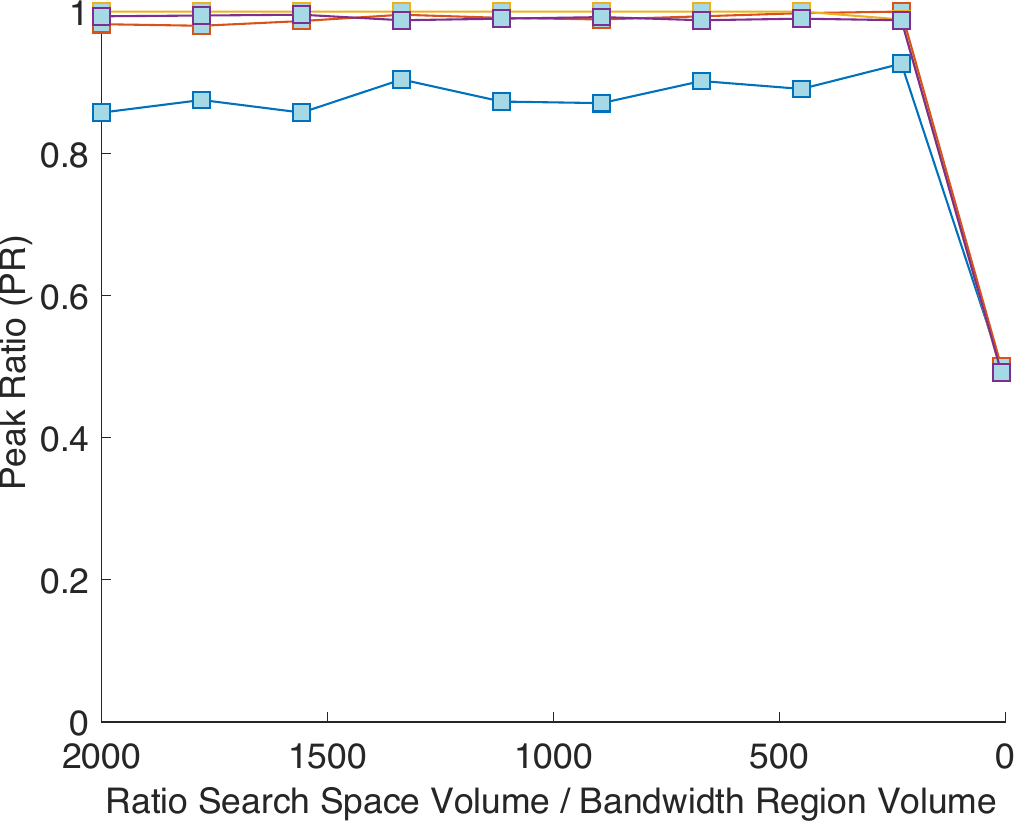}}
    \subfigure[\protect\url{}\label{fig:f07_r}$F_7$]
    {\includegraphics[height=3.5cm]{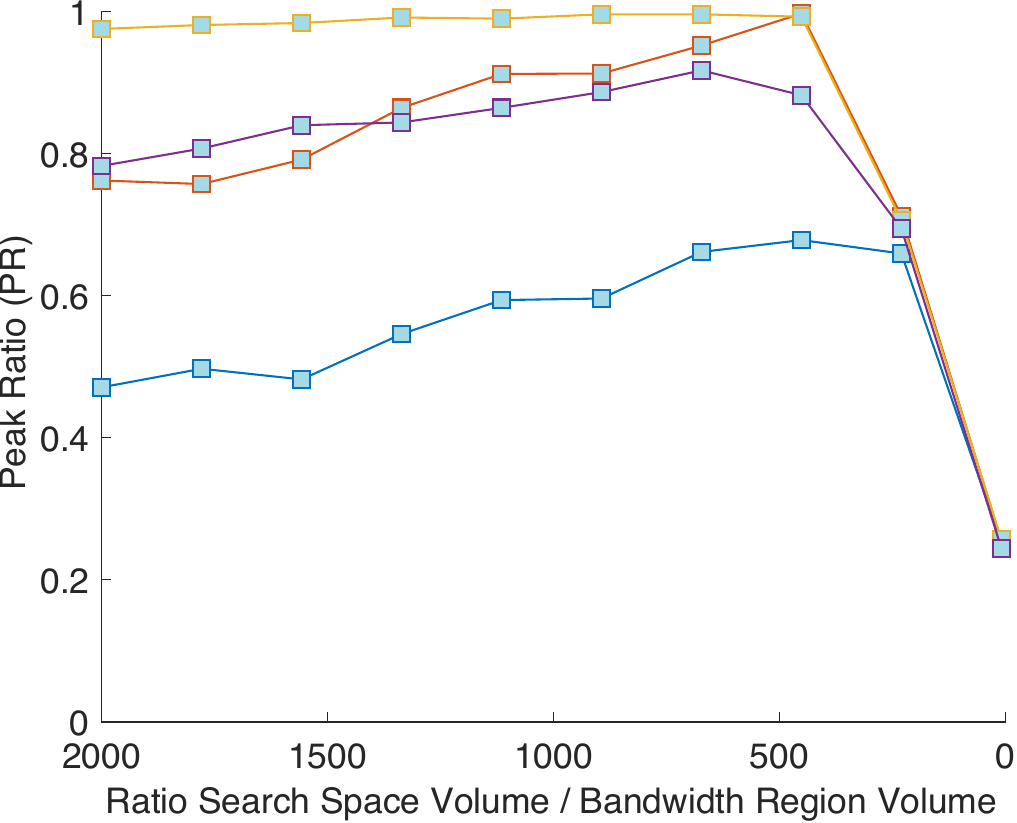}}
    \subfigure[\protect\url{}\label{fig:f08_r}$F_8$]
    {\includegraphics[height=3.5cm]{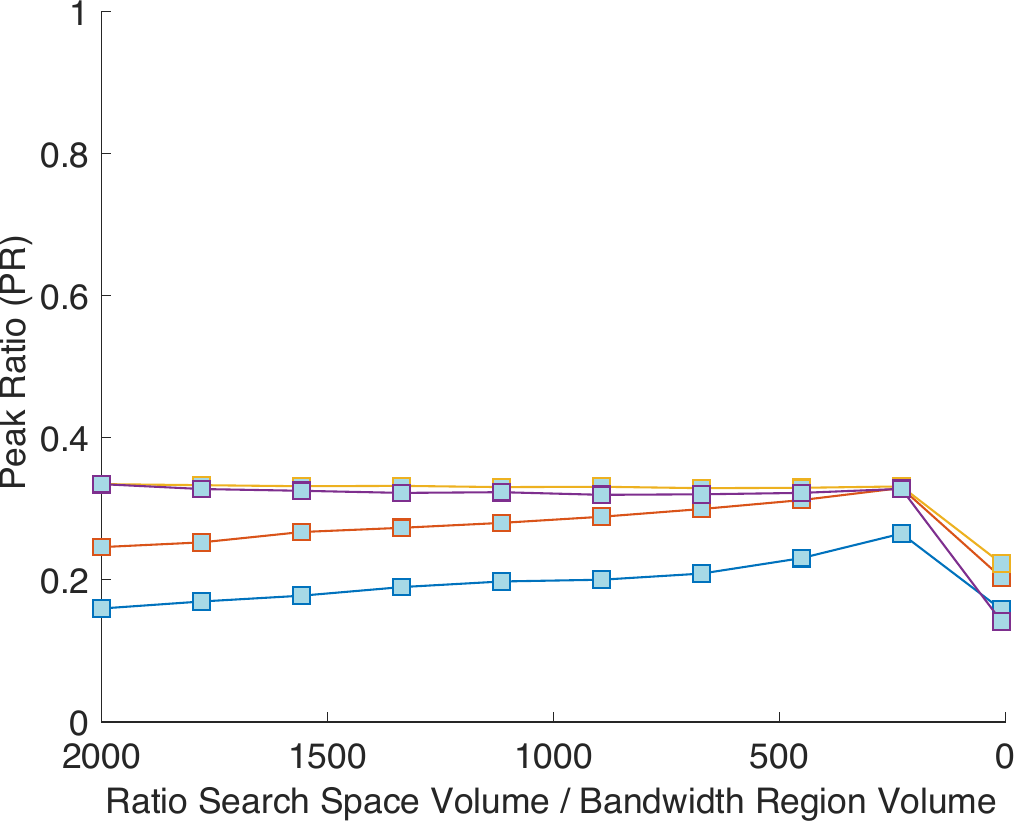}}

    \subfigure[\protect\url{}\label{fig:f09_r}$F_9$]
    {\includegraphics[height=3.5cm]{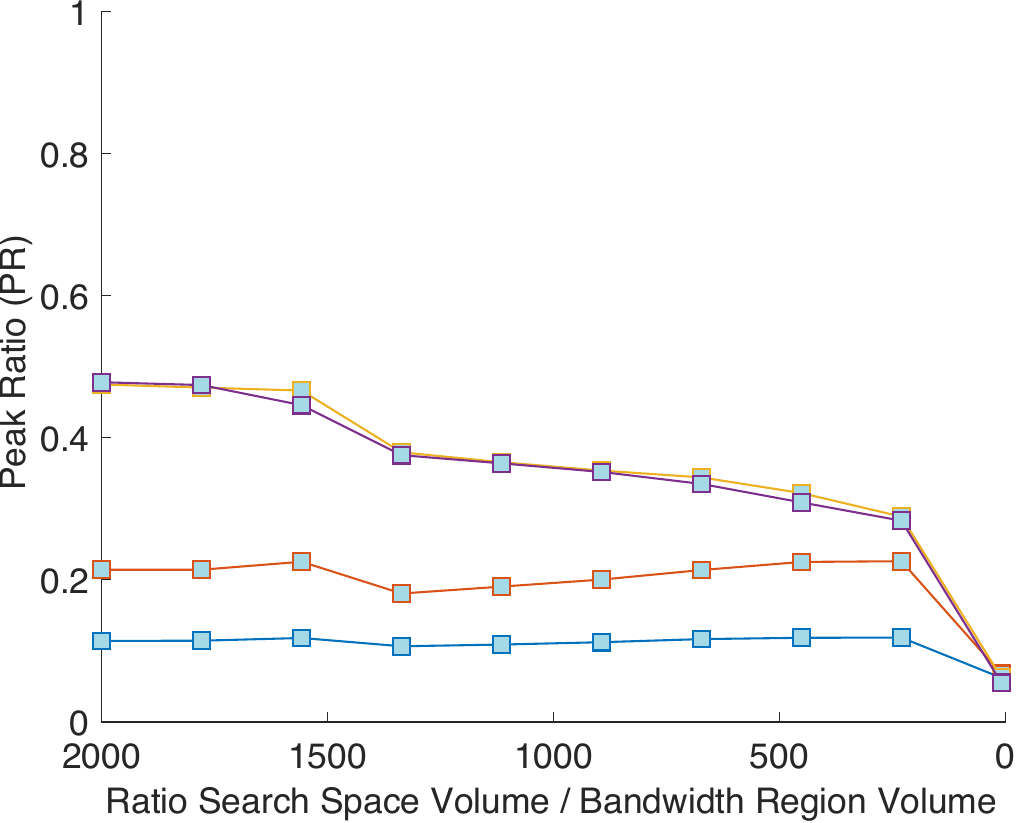}}
    \subfigure[\protect\url{}\label{fig:f10_r}$F_{10}$]
    {\includegraphics[height=3.5cm]{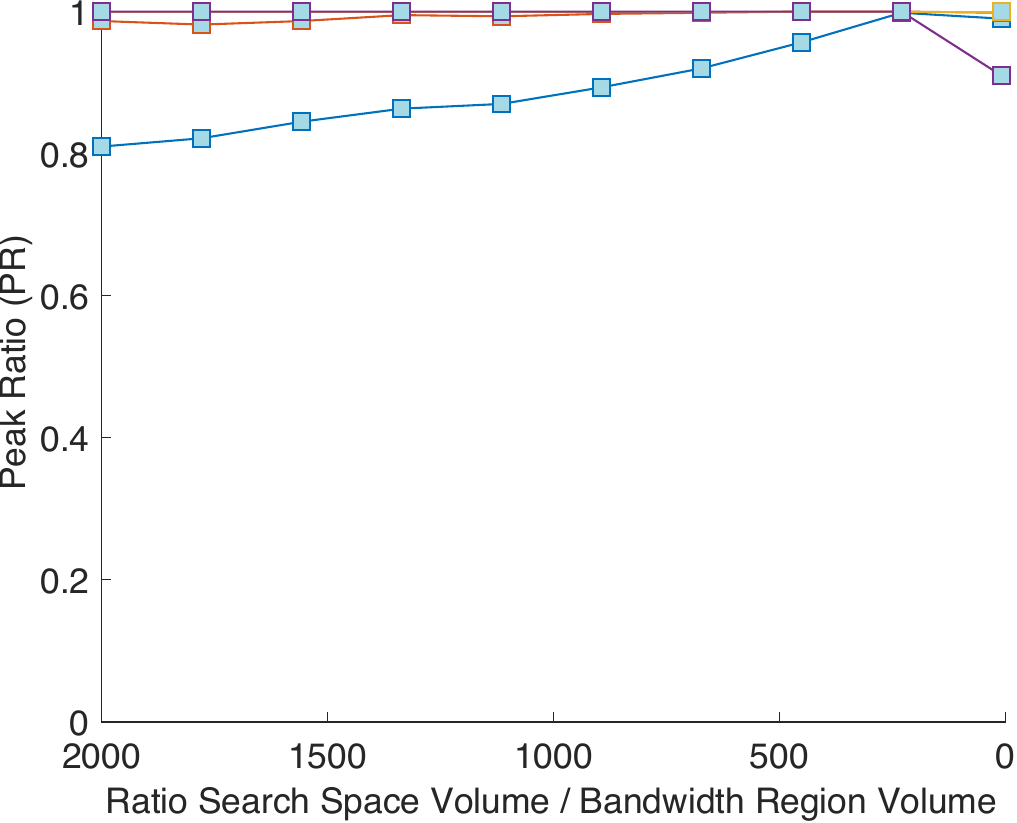}}
    \subfigure[\protect\url{}\label{fig:f11_r}$F_{11}$]
    {\includegraphics[height=3.5cm]{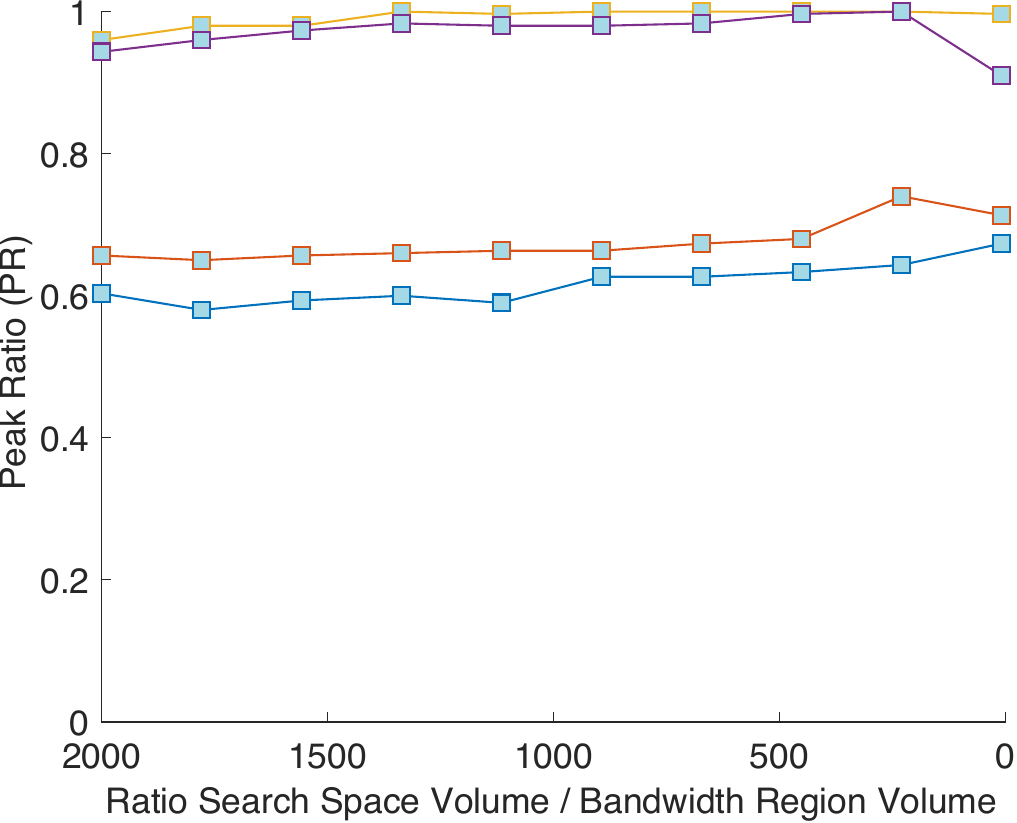}}
    \subfigure[\protect\url{}\label{fig:f12_r}$F_{12}$]
    {\includegraphics[height=3.5cm]{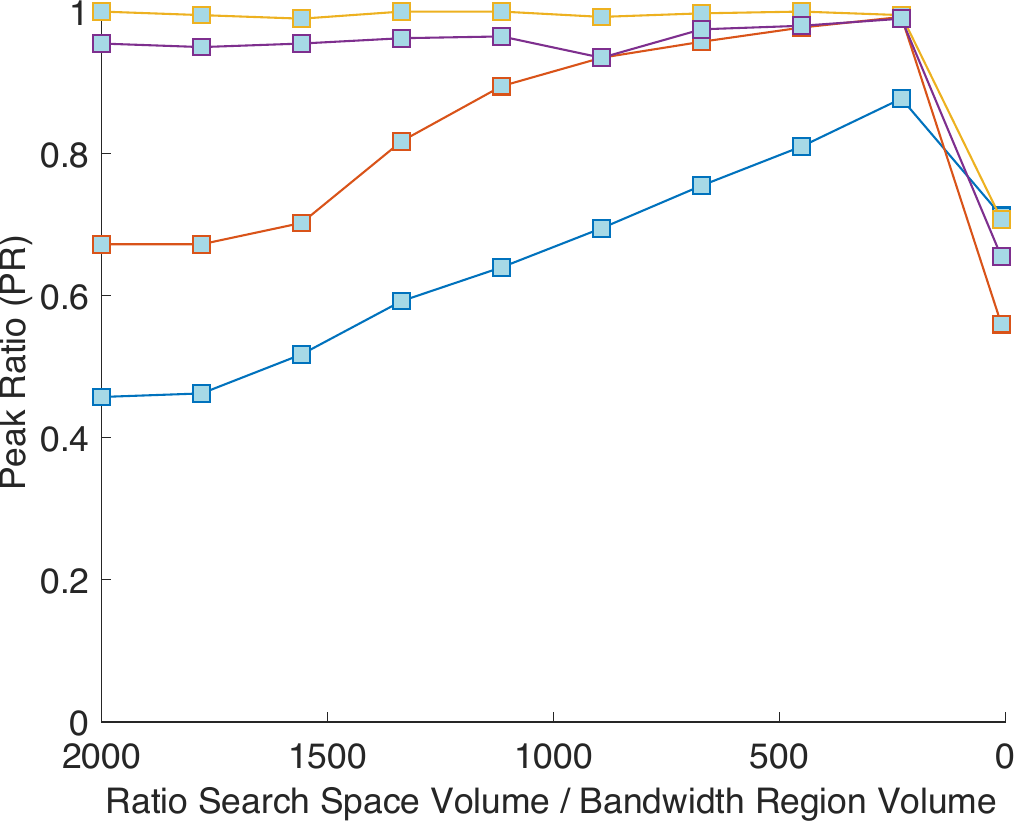}}

    \subfigure[\protect\url{}\label{fig:f13_r}$F_{13}$]
    {\includegraphics[height=3.5cm]{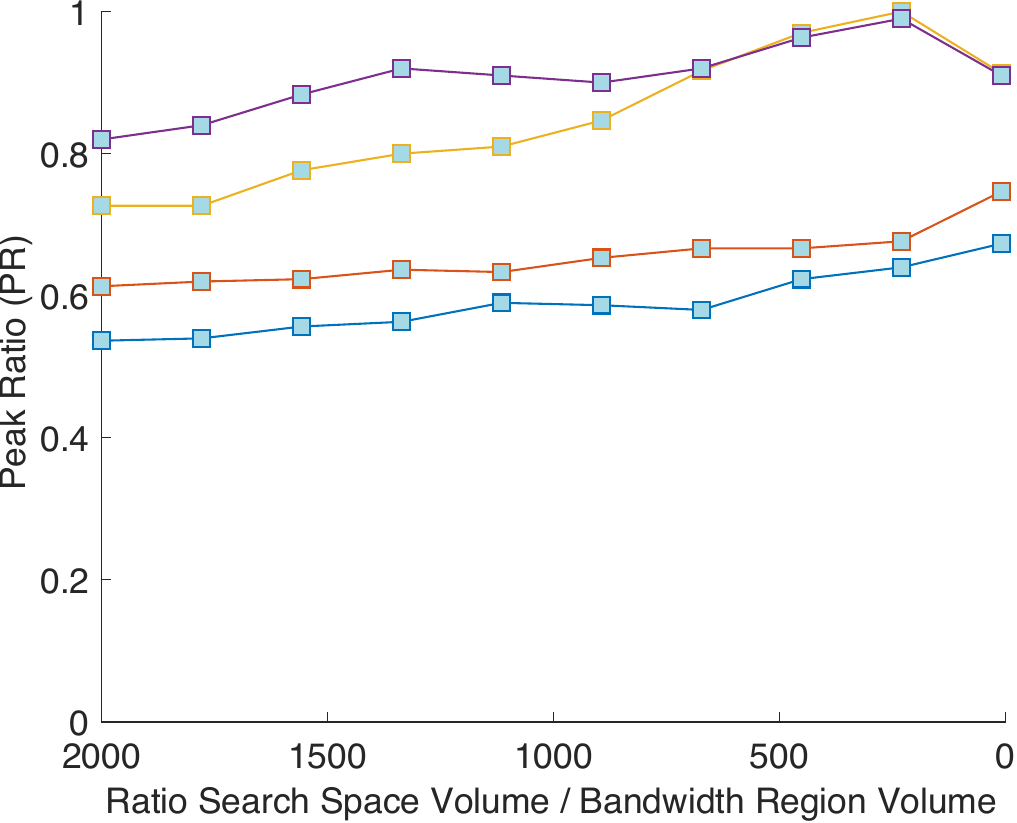}}
    \subfigure[\protect\url{}\label{fig:f14_r}$F_{14}$]
    {\includegraphics[height=3.5cm]{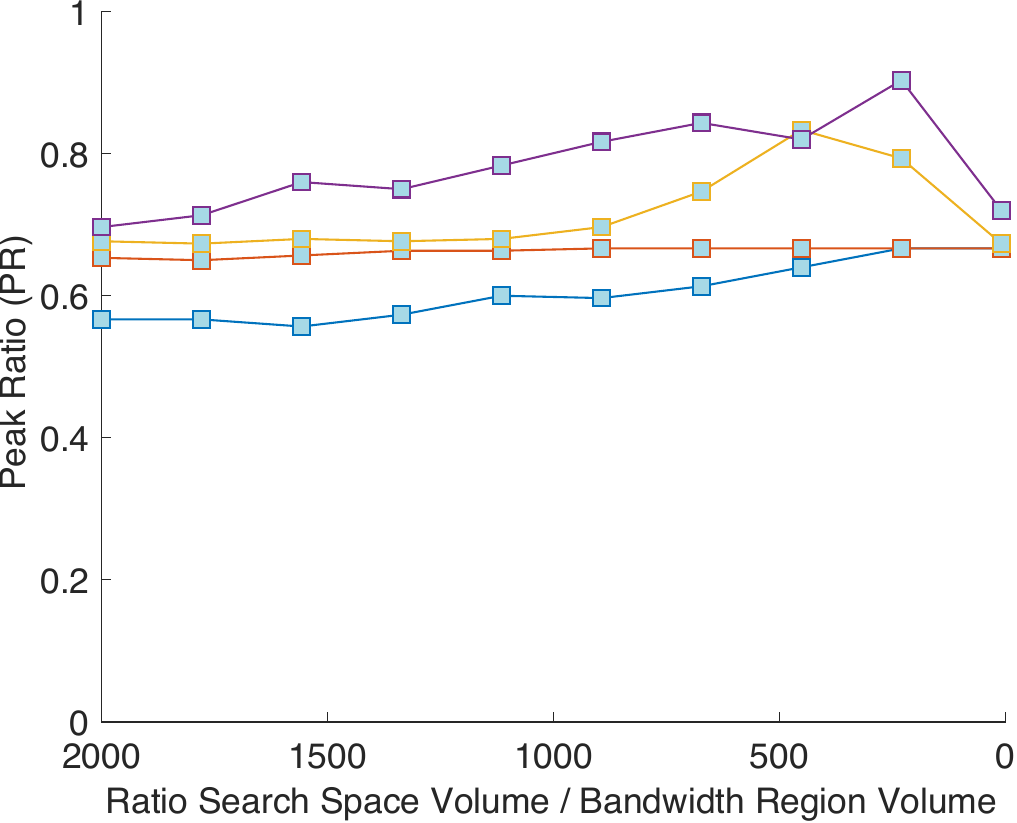}}
    \subfigure[\protect\url{}\label{fig:f15_r}$F_{15}$]
    {\includegraphics[height=3.5cm]{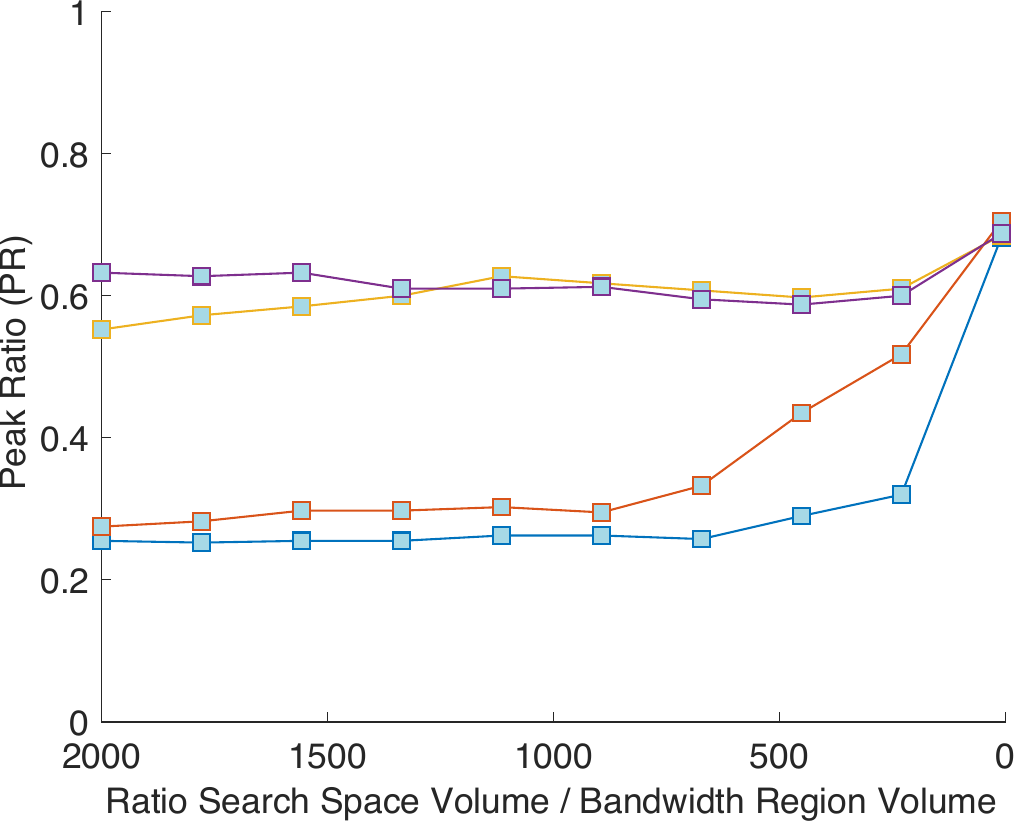}}
    \subfigure[\protect\url{}\label{fig:f16_r}$F_{16}$]
    {\includegraphics[height=3.5cm]{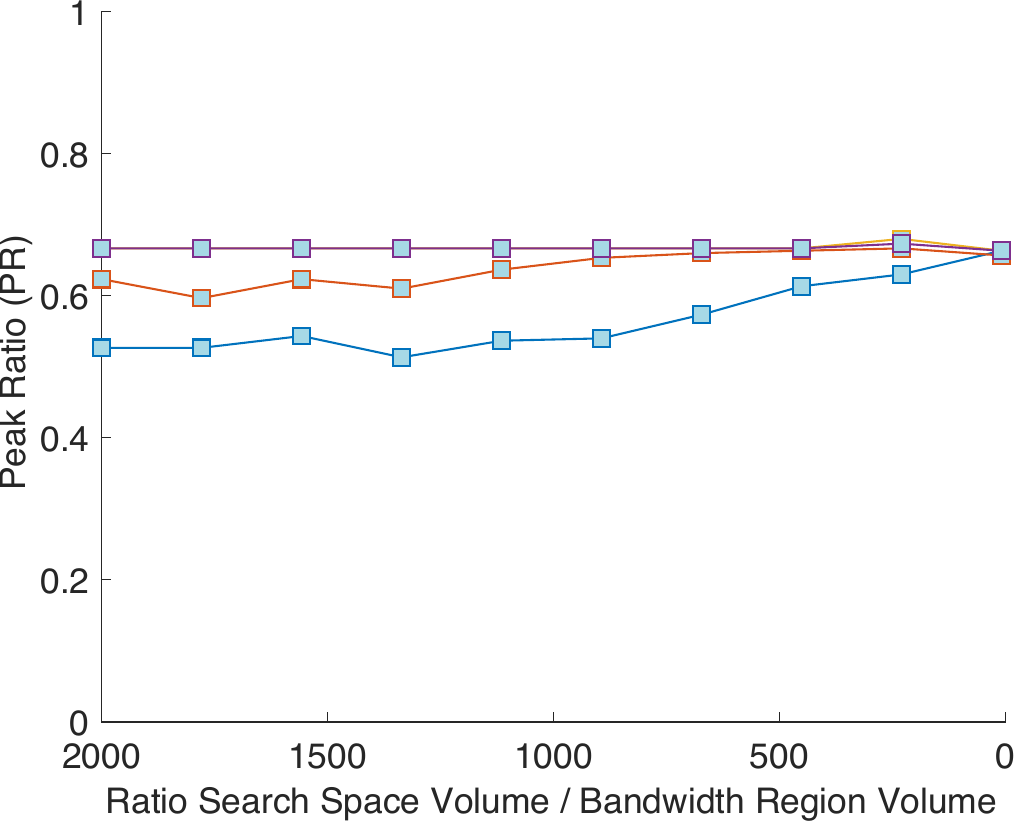}}

    \subfigure[\protect\url{}\label{fig:f17_r}$F_{17}$]
    {\includegraphics[height=3.5cm]{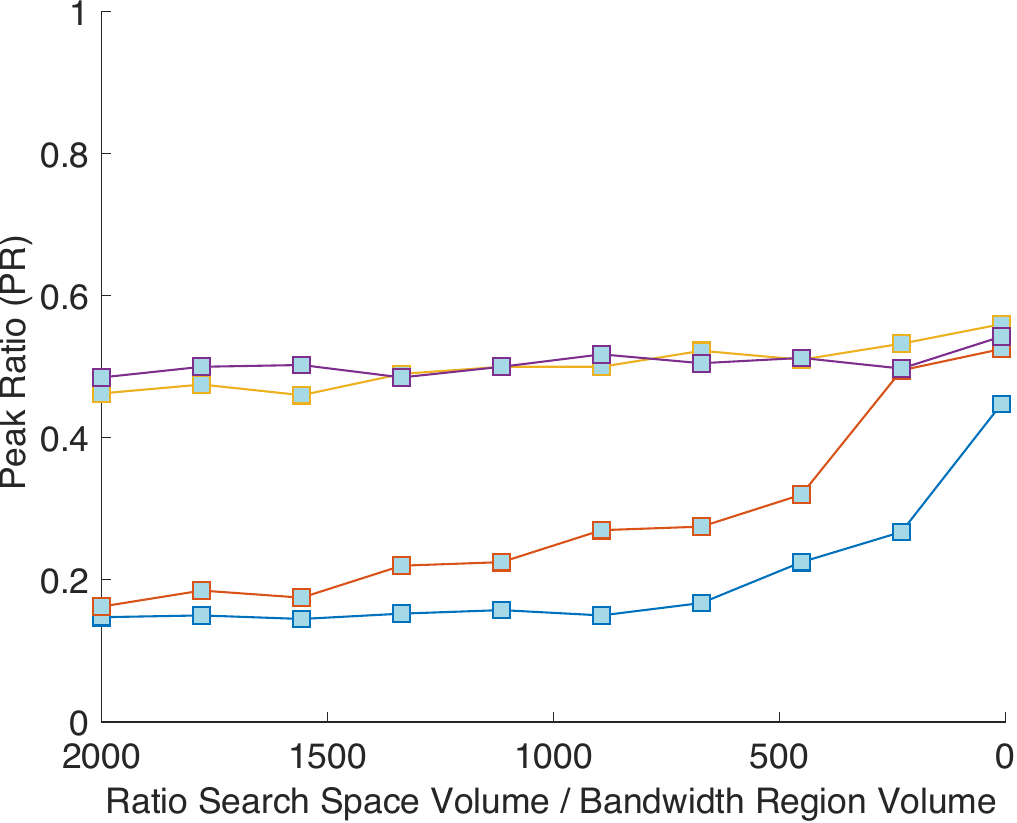}}
    \subfigure[\protect\url{}\label{fig:f18_r}$F_{18}$]
    {\includegraphics[height=3.5cm]{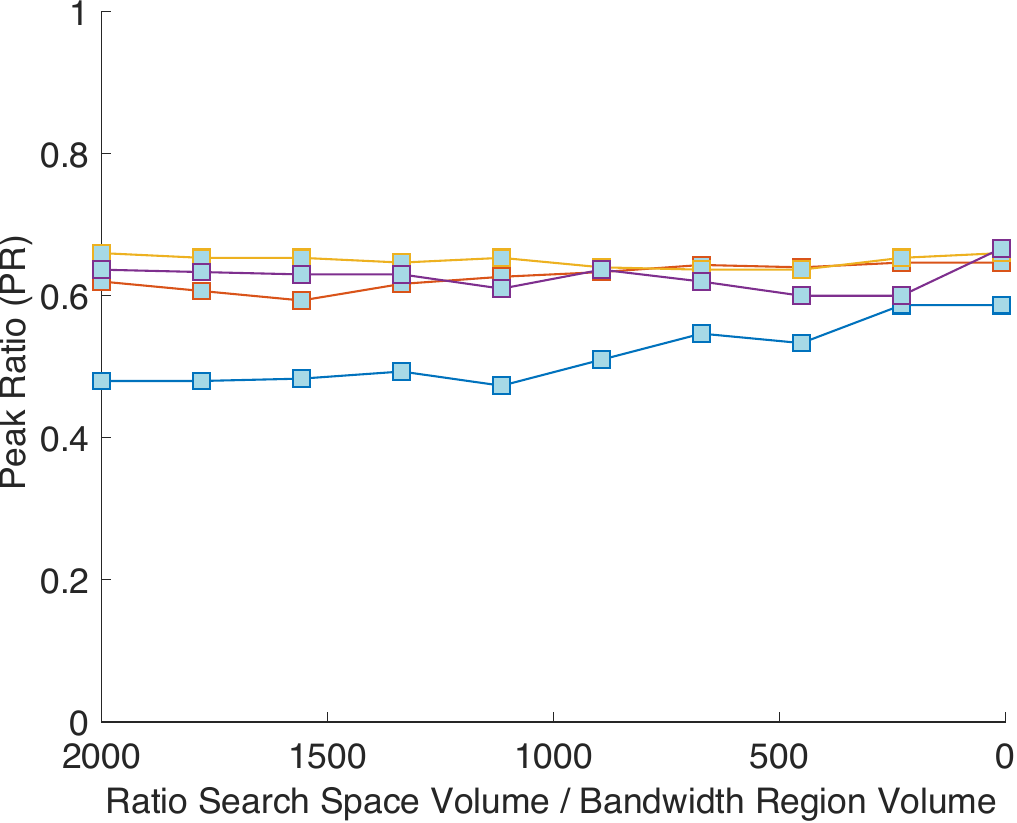}}
    \subfigure[\protect\url{}\label{fig:f19_r}$F_{19}$]
    {\includegraphics[height=3.5cm]{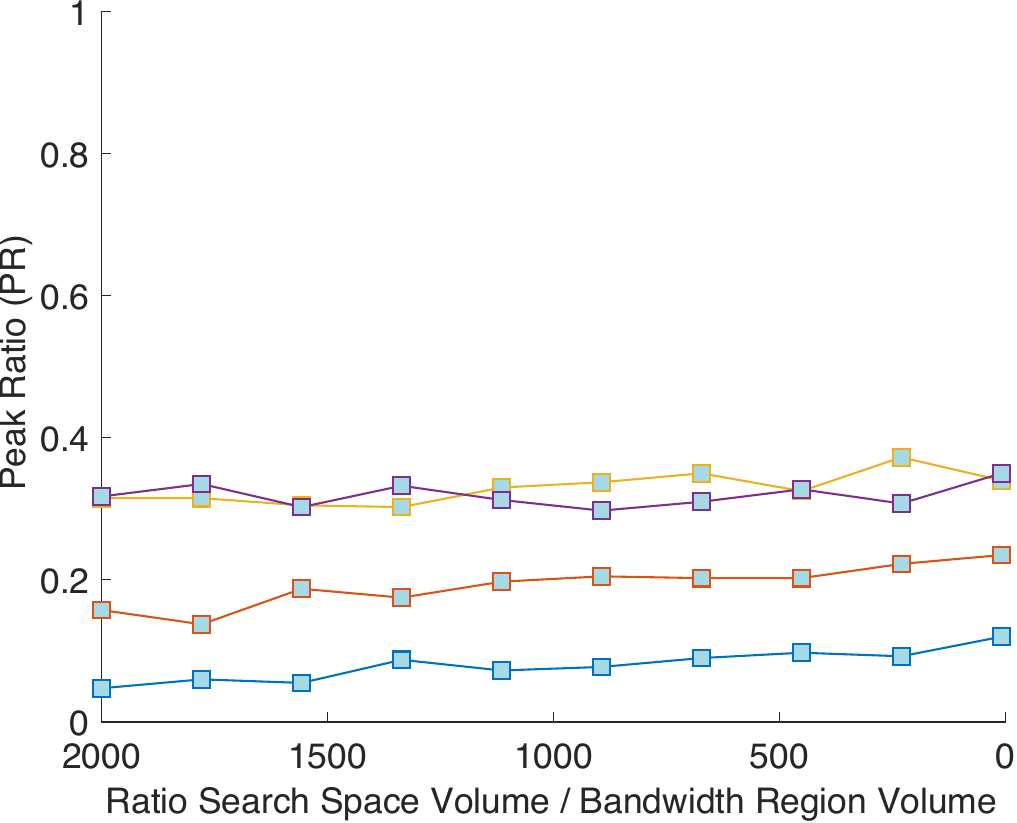}}
    \subfigure[\protect\url{}\label{fig:f20_r}$F_{20}$]
    {\includegraphics[height=3.5cm]{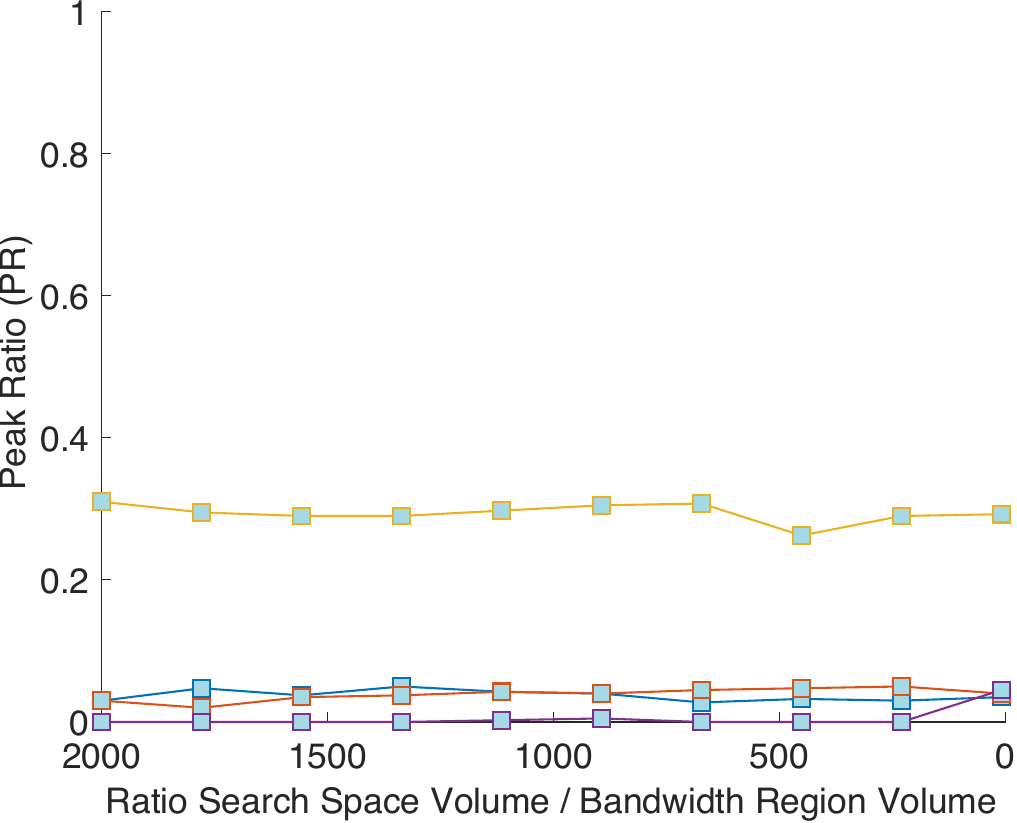}}
    
    \caption{Changes in the PR values (at the accuracy level $\varepsilon=1.0E-04$) with different Population Size (n) and the ratio between Clustering Kernel Bandwidth Volume and Search Space Volume on CEC'2013 benchmark set.}
    \label{fig:pr_values_r}
\end{figure*}

\begin{figure*}[h!]
    \centering
    \subfigure[\protect\url{}\label{fig:f01_r2}$F_1$]
    {\includegraphics[height=3.5cm]{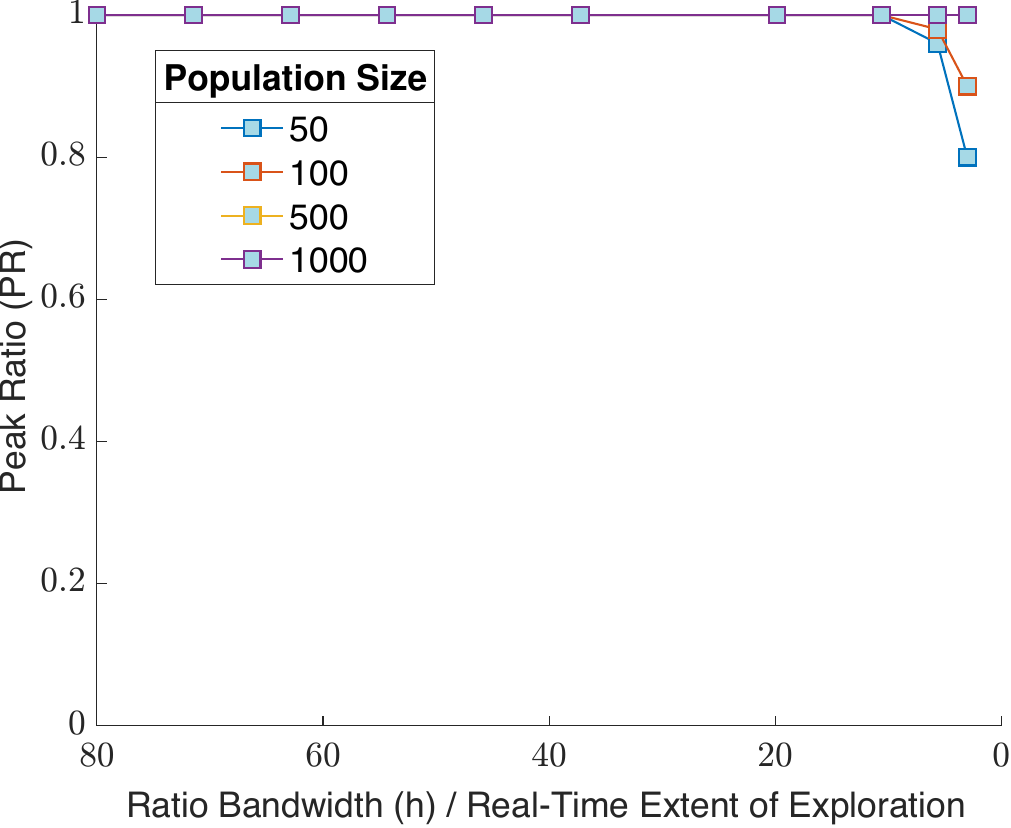}}
    \subfigure[\protect\url{}\label{fig:f02_r2}$F_2$]
    {\includegraphics[height=3.5cm]{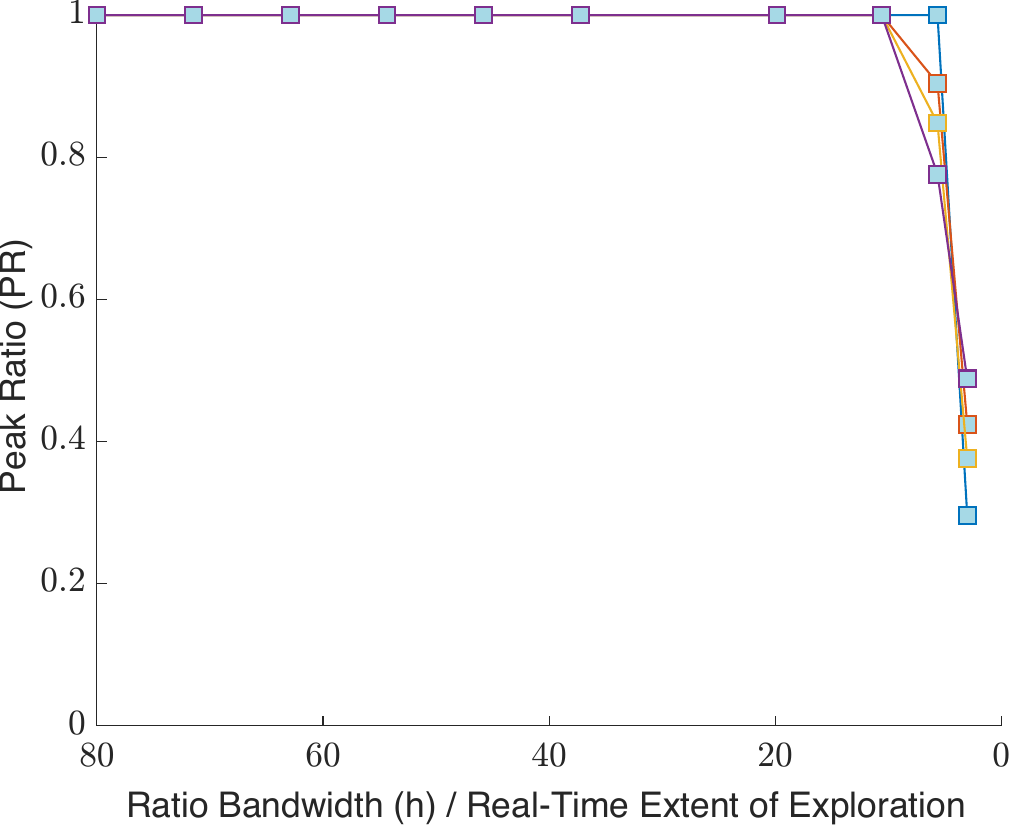}}
    \subfigure[\protect\url{}\label{fig:f03_r2}$F_3$]
    {\includegraphics[height=3.5cm]{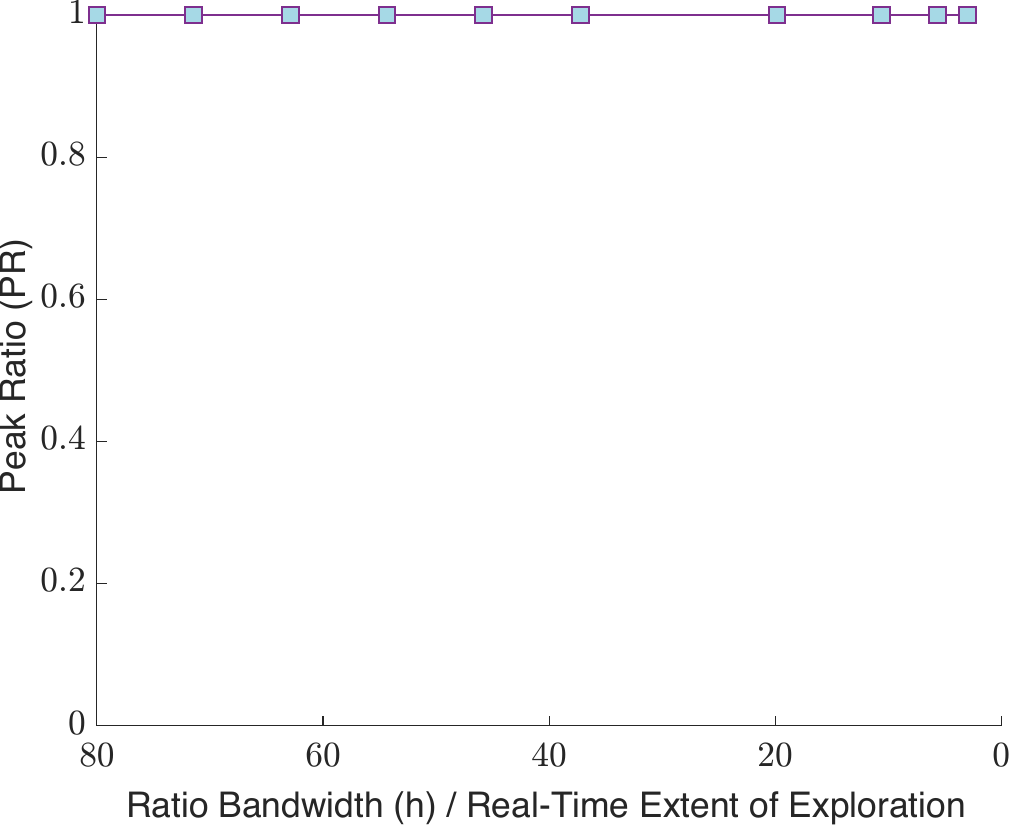}}
    \subfigure[\protect\url{}\label{fig:f04_r2}$F_4$]
    {\includegraphics[height=3.5cm]{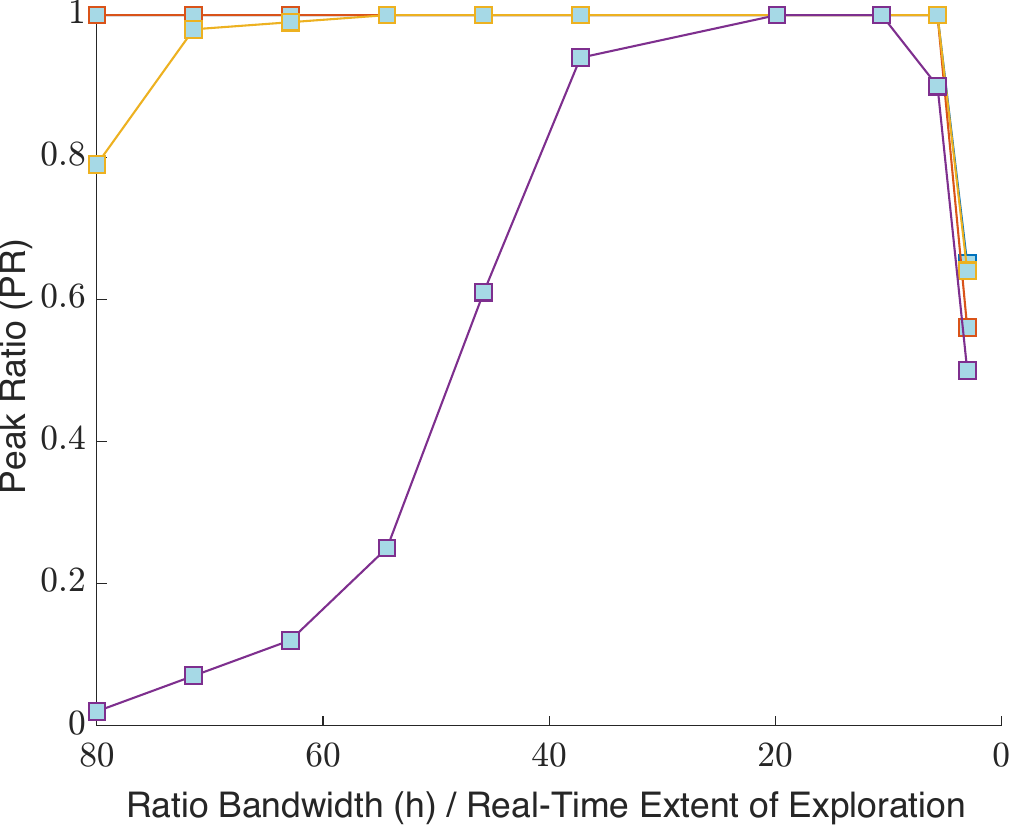}}

    \subfigure[\protect\url{}\label{fig:f05_r2}$F_5$]
    {\includegraphics[height=3.5cm]{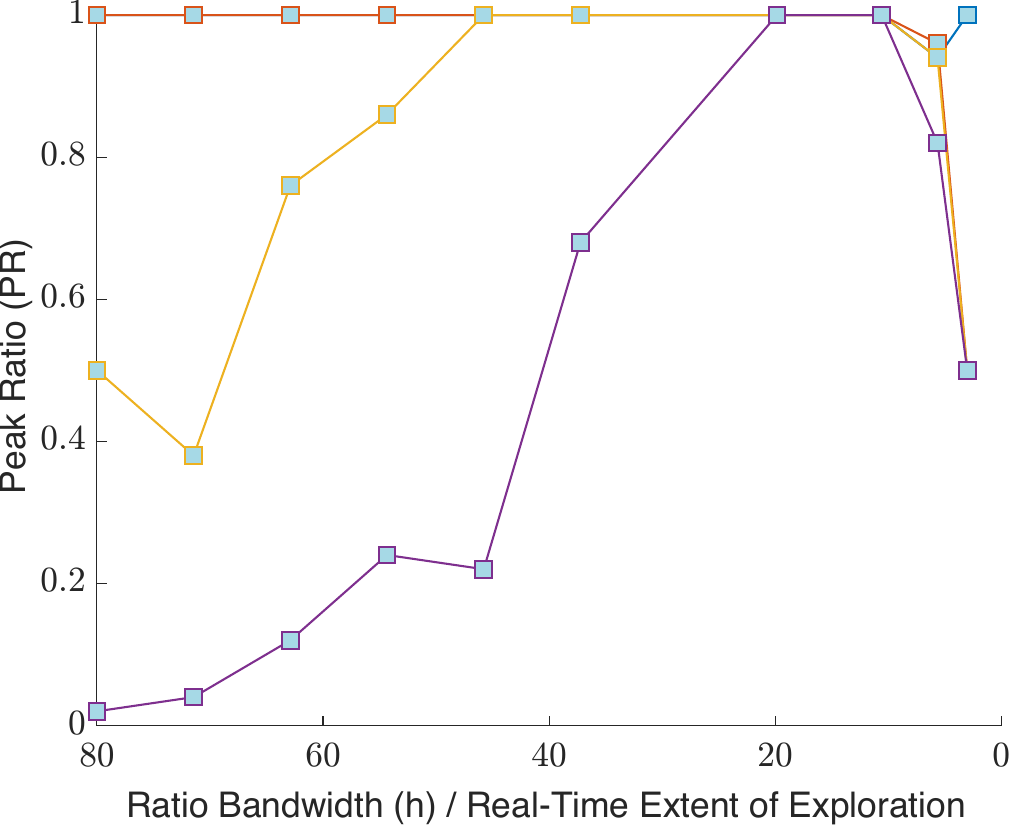}}
    \subfigure[\protect\url{}\label{fig:f06_r2}$F_6$]
    {\includegraphics[height=3.5cm]{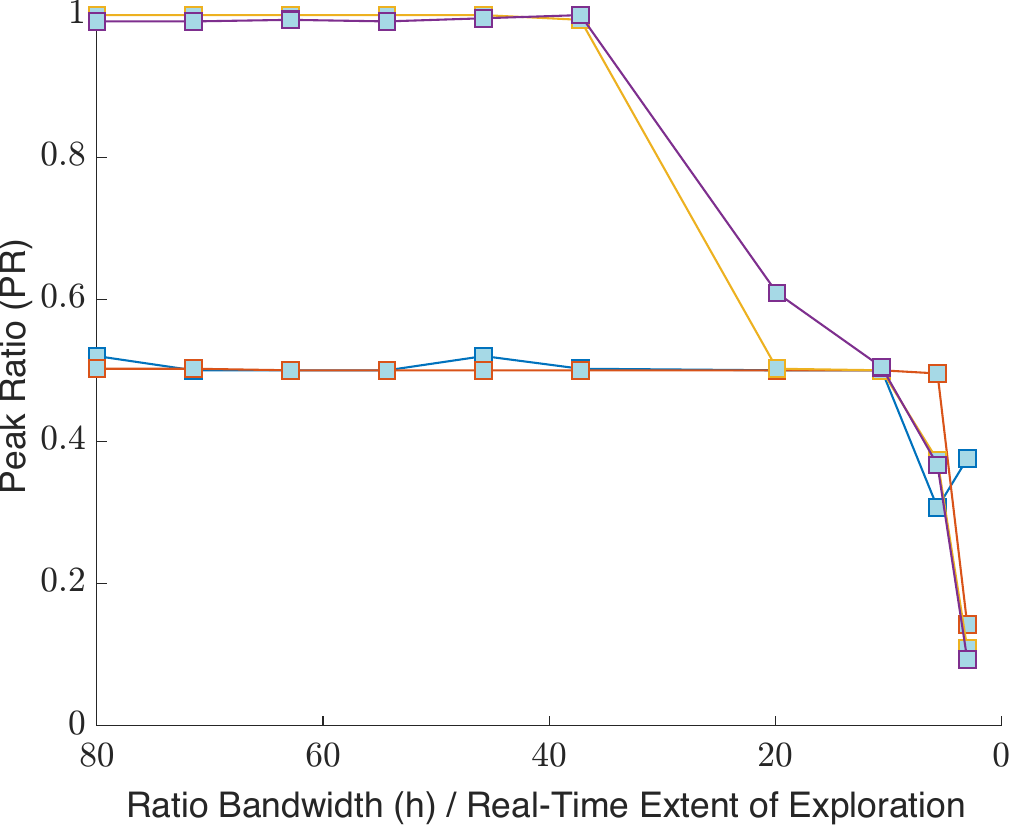}}
    \subfigure[\protect\url{}\label{fig:f07_r2}$F_7$]
    {\includegraphics[height=3.5cm]{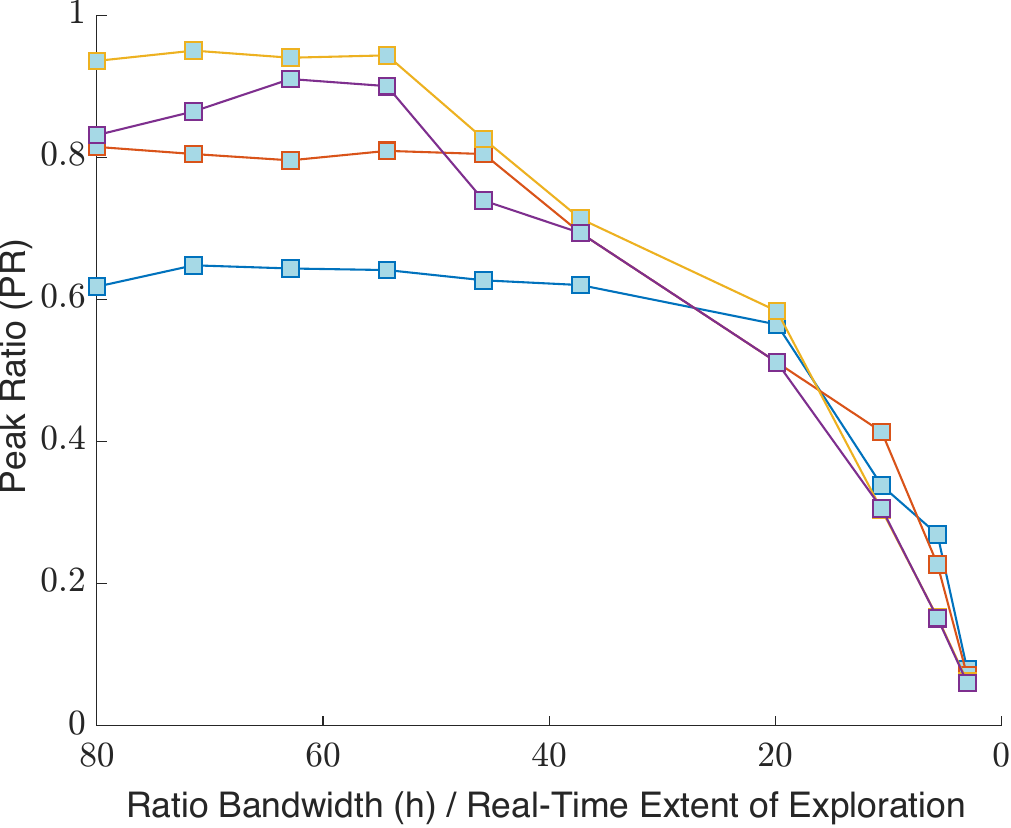}}
    \subfigure[\protect\url{}\label{fig:f08_r2}$F_8$]
    {\includegraphics[height=3.5cm]{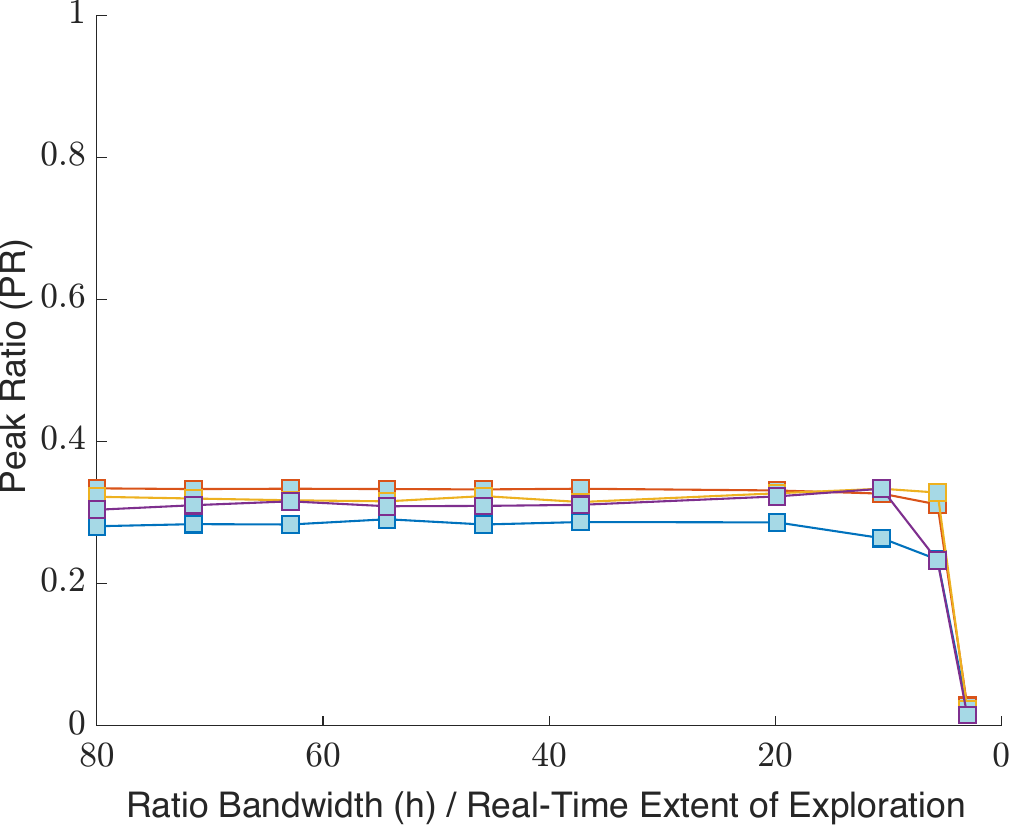}}

    \subfigure[\protect\url{}\label{fig:f09_r2}$F_9$]
    {\includegraphics[height=3.5cm]{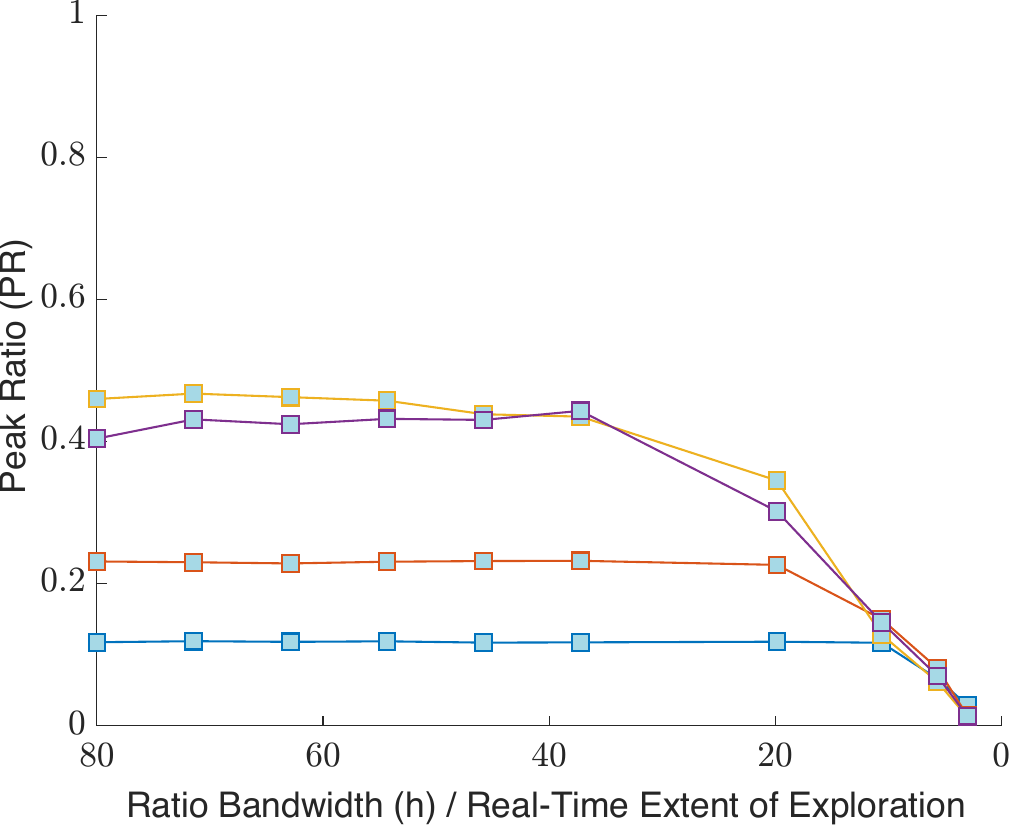}}
    \subfigure[\protect\url{}\label{fig:f10_r2}$F_{10}$]
    {\includegraphics[height=3.5cm]{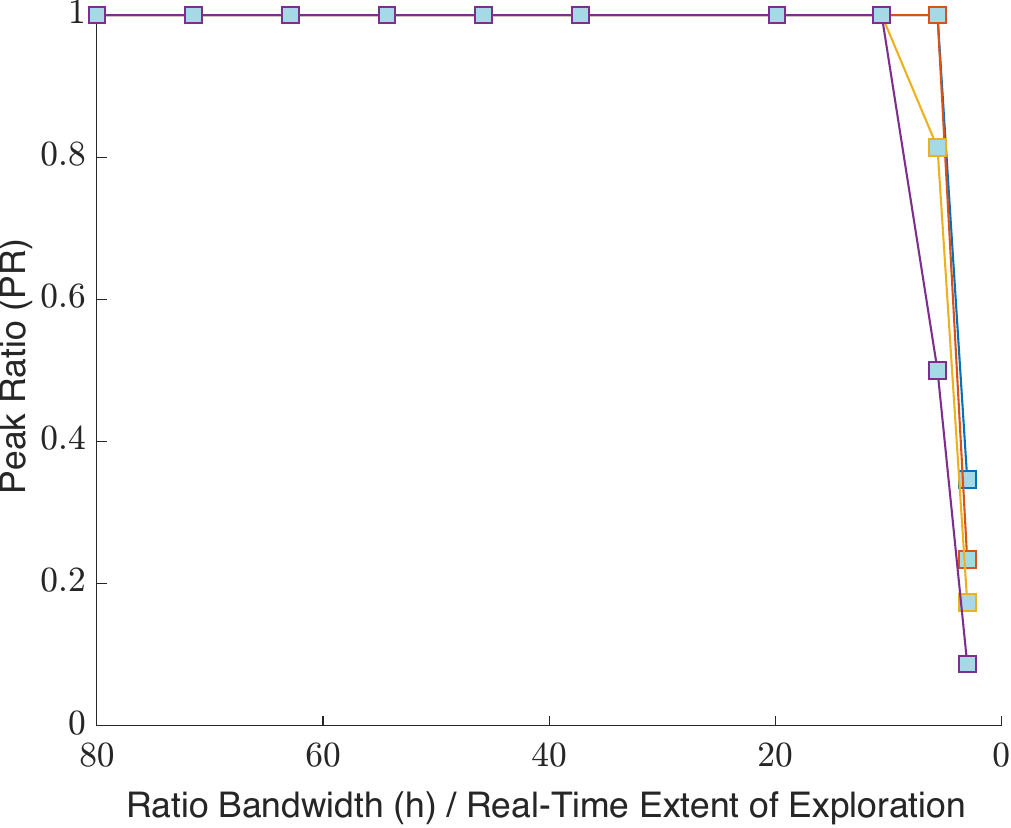}}
    \subfigure[\protect\url{}\label{fig:f11_r2}$F_{11}$]
    {\includegraphics[height=3.5cm]{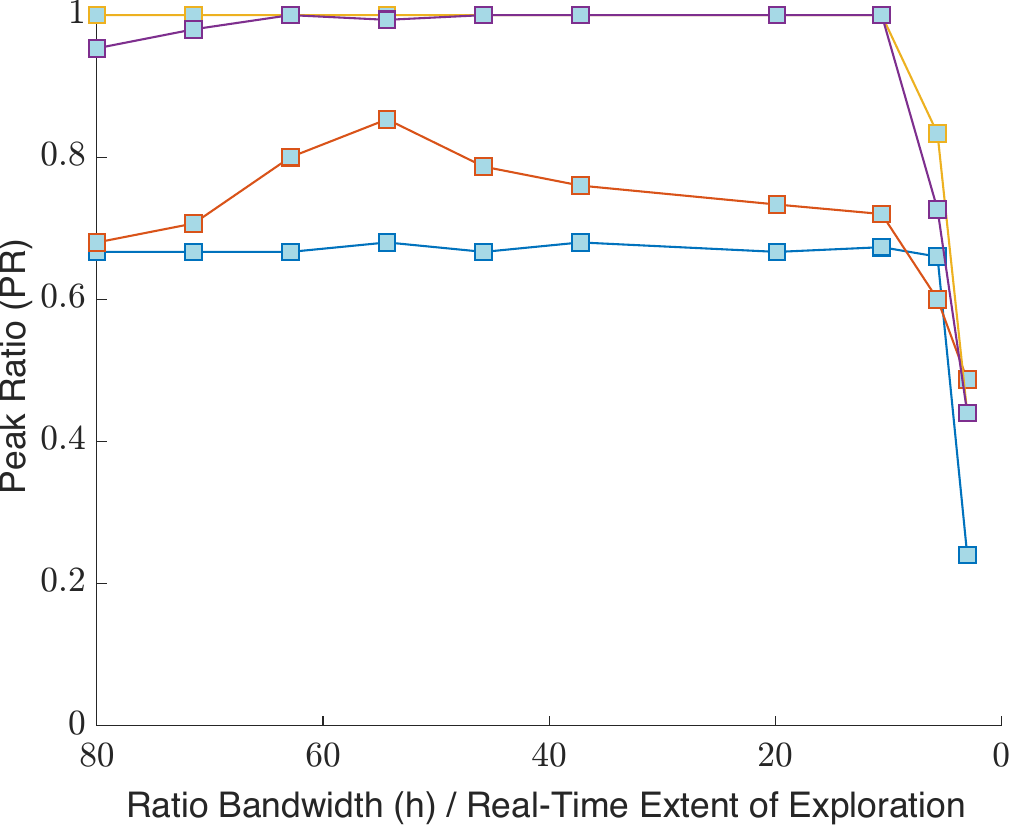}}
    \subfigure[\protect\url{}\label{fig:f12_r2}$F_{12}$]
    {\includegraphics[height=3.5cm]{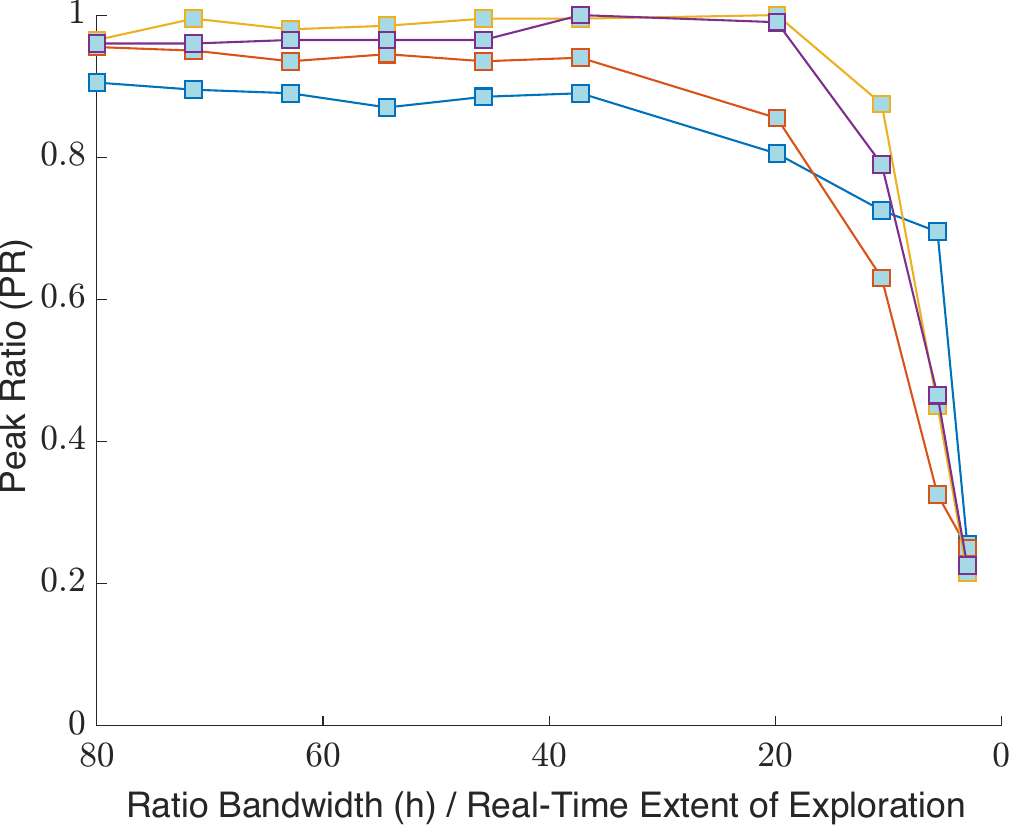}}

    \subfigure[\protect\url{}\label{fig:f13_r2}$F_{13}$]
    {\includegraphics[height=3.5cm]{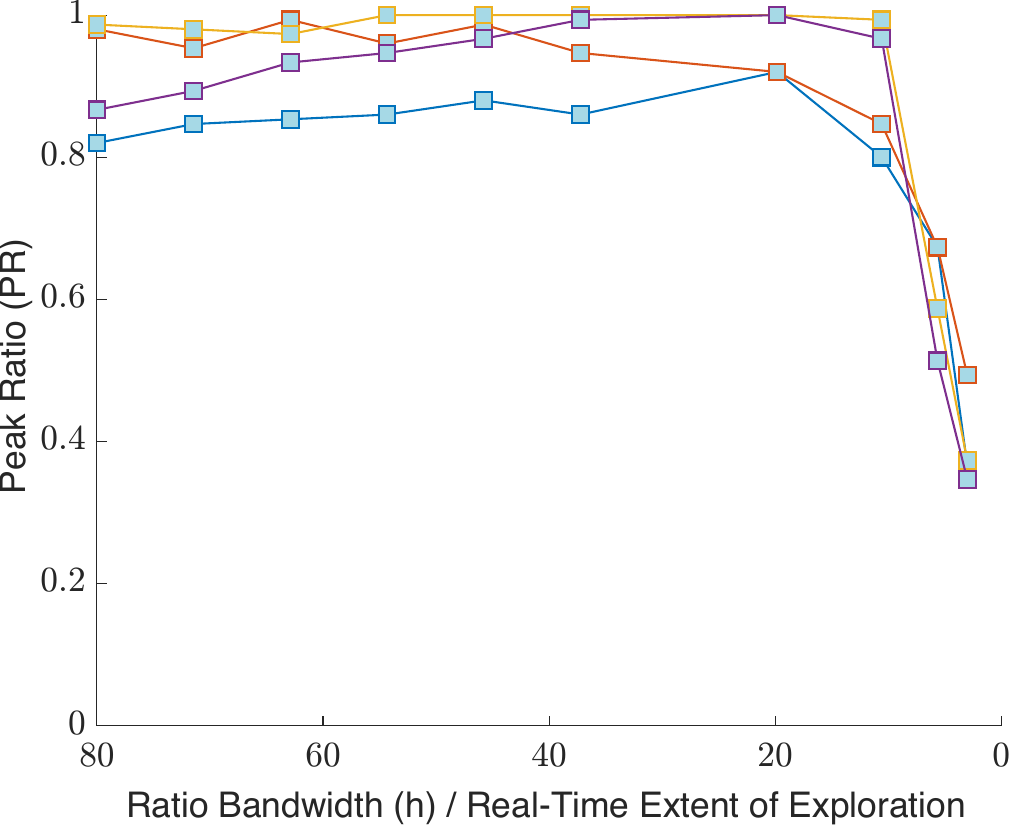}}
    \subfigure[\protect\url{}\label{fig:f14_r2}$F_{14}$]
    {\includegraphics[height=3.5cm]{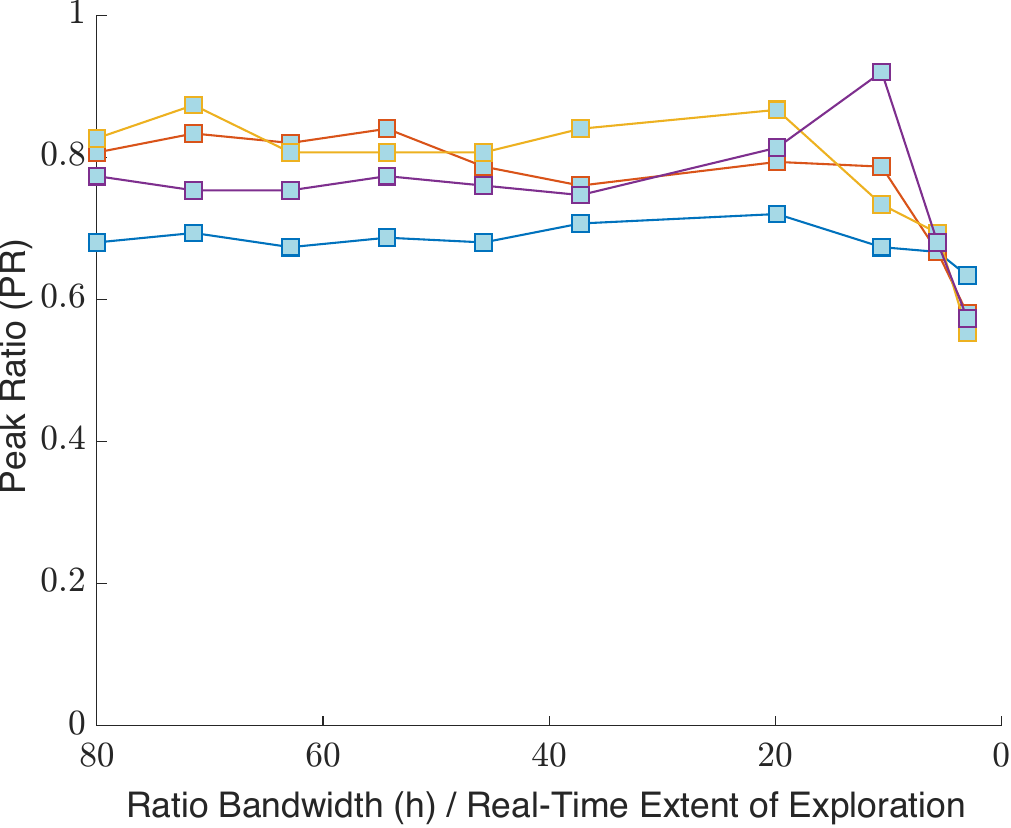}}
    \subfigure[\protect\url{}\label{fig:f15_r2}$F_{15}$]
    {\includegraphics[height=3.5cm]{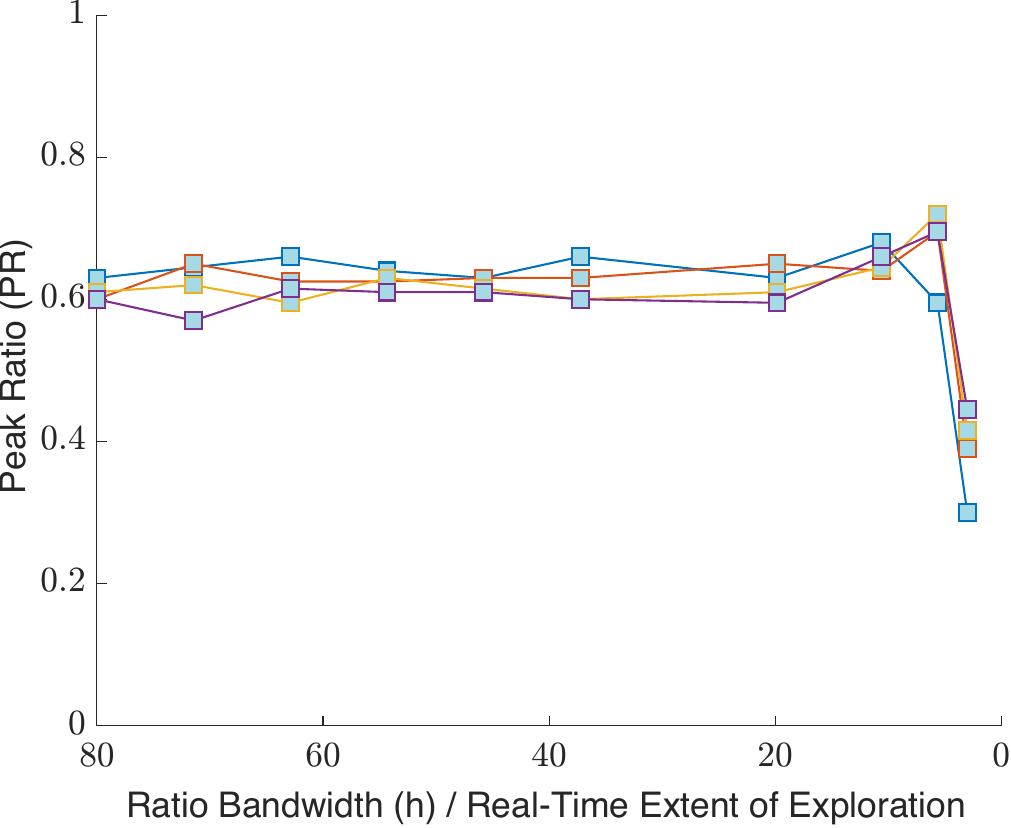}}
    \subfigure[\protect\url{}\label{fig:f16_r2}$F_{16}$]
    {\includegraphics[height=3.5cm]{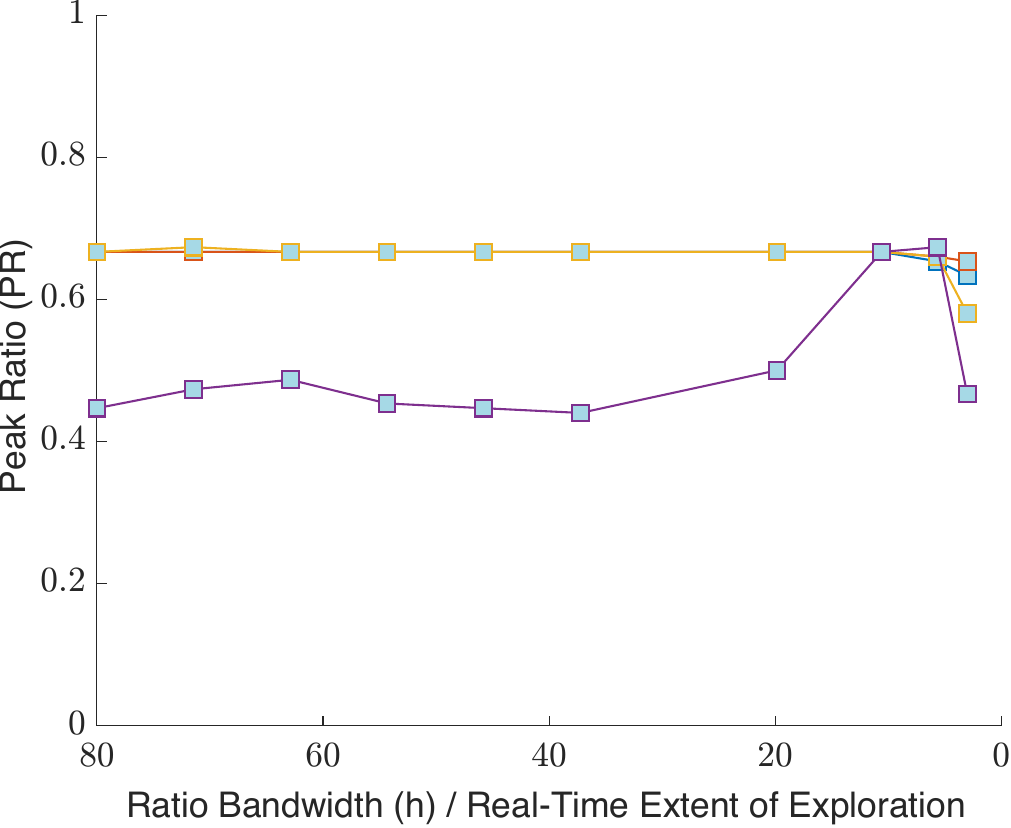}}

    \subfigure[\protect\url{}\label{fig:f17_r2}$F_{17}$]
    {\includegraphics[height=3.5cm]{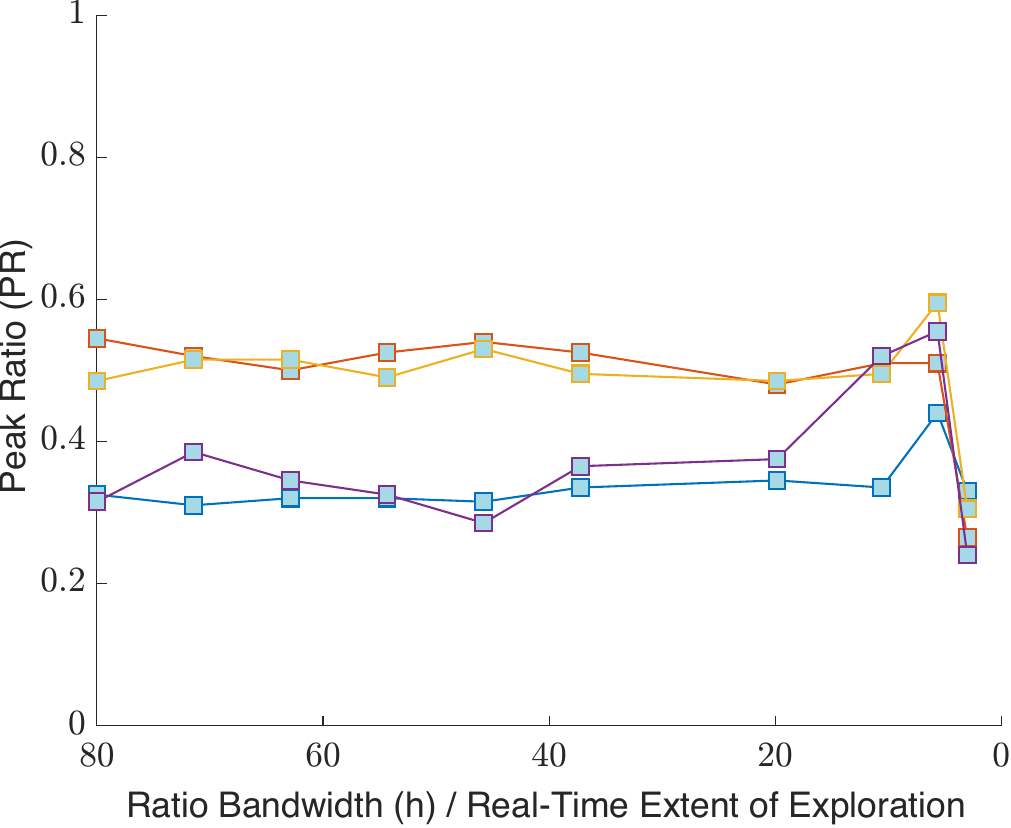}}
    \subfigure[\protect\url{}\label{fig:f18_r2}$F_{18}$]
    {\includegraphics[height=3.5cm]{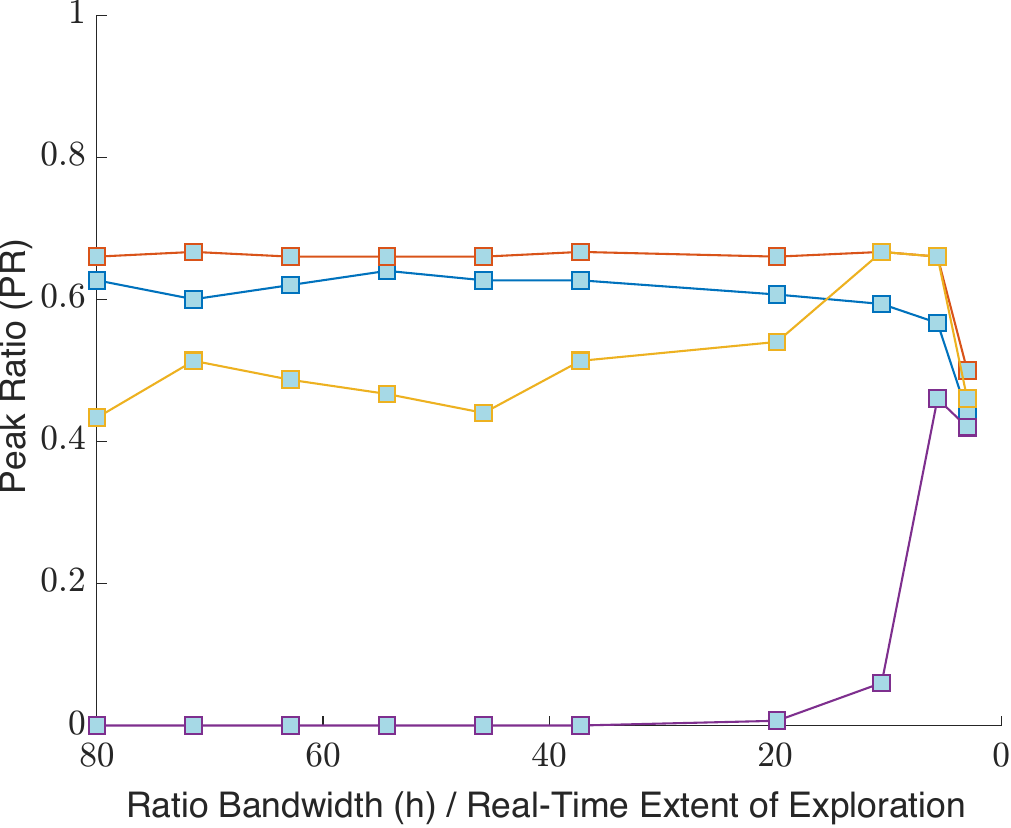}}
    \subfigure[\protect\url{}\label{fig:f19_r2}$F_{19}$]
    {\includegraphics[height=3.5cm]{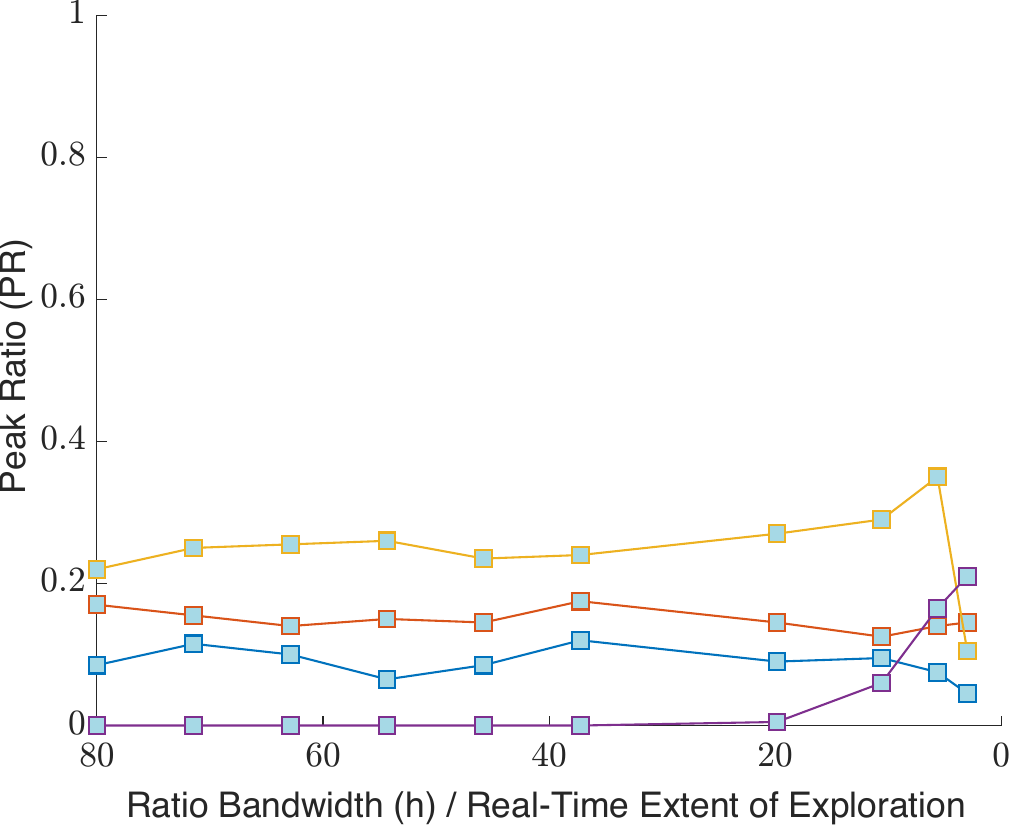}}
    \subfigure[\protect\url{}\label{fig:f20_r2}$F_{20}$]
    {\includegraphics[height=3.5cm]{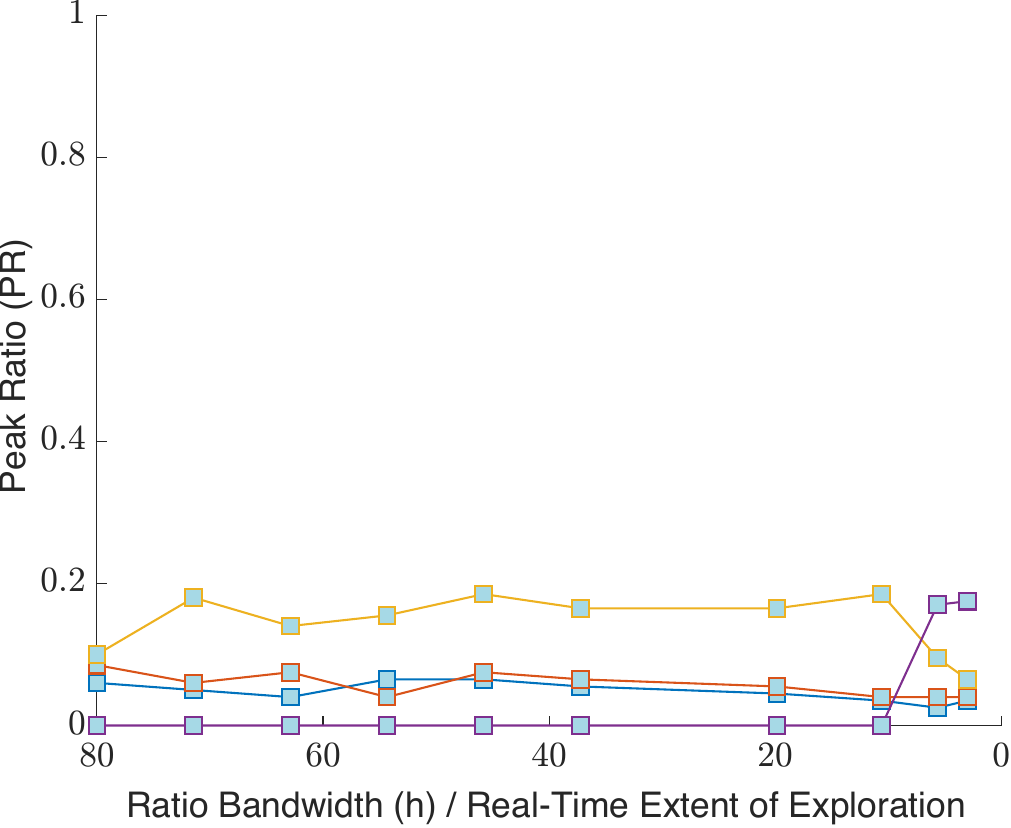}}
    
    \caption{Changes in the PR values (at the accuracy level $\varepsilon=1.0E-04$) with different Population Size (n) and Clustering Kernel Bandwidth in relation to the real-time extent of exploration (at each iteration) on CEC'2013 benchmark set.}
    \label{fig:pr_values_r2}
\end{figure*}

To observe how the values of the parameters affect the results of MGP-BBBC, we compared the best result for each function against each combination with the Wilcoxon rank-sum test at the significance level $\alpha = 0.05$ (results are reported in the supplementary material under Tables S.XXVII and S.XXVIII for accuracy $\varepsilon = 1.0E-03$, S.XXX and S.XXXI for $\varepsilon = 1.0E-04$, and S.XXXIII and S.XXXIV for $\varepsilon = 1.0E-05$). The tables indicate the number of settings that performed significantly worse ($-$) or non-significantly different ($\approx$) than the best setting in terms of both population size (rows) and the ratio between search space volume and bandwidth region volume (columns). The trend for each problem is graphically depicted in Fig.~\ref{fig:pr_values_r} for accuracy $\varepsilon = 1.0E-04$ only. 

We adopted Friedman's test with the Bonferroni-Dunn procedure~\cite{demvsar2006statistical} to test the robustness of MGP-BBBC on the 780 combinations between population size and ratio (results are reported in the supplementary material under Table S.XXIX for accuracy $\varepsilon = 1.0E-03$, S.XXXII for $\varepsilon = 1.0E-04$, and S.XXXV for $\varepsilon = 1.0E-05$). 
The p-value shows the result of the Bonferroni-Dunn procedure; ``Y'' indicates that there is a significant difference (at the significance level $\alpha = 0.05$) between the corresponding pairwise settings, whereas ``N'' indicates no significant difference.
We report obtaining 524 (67\%) combinations being non-significantly different against 256 (33\%) combinations that are significantly different -- the same outcome is observable for each accuracy level  $\varepsilon=1.0E-03$, $\varepsilon=1.0E-04$, and $\varepsilon=1.0E-05$. Most of the significant differences were found for population size $n=500$ and $n=1000$ and for small versus large bandwidth values. These results are in line with the plots in Fig.~\ref{fig:pr_values_r}, which show more consistency than the first experimental results in Fig.~\ref{fig:pr_values} (each function with specifically tuned bandwidths values, Sec.~\ref{sec:paramAnalysis_specific}) and support the hypothesis of relating the clustering kernel bandwidth with the volume of the search space for setting parameters independently from the problem to be solved.

\subsection{Kernel Bandwidth Values to Real-Time Extent of Exploration}
\label{sec:paramAnalysis_exploration}

To relate the kernel bandwidth to the extent of exploration performed in real time, we calculated (at each iteration) the maximum Euclidean distance in the population -- i.e., the distance between the two furthest individuals in the population at each generation. This provides us with a dynamic self-tuned indicator for setting how much the bandwidth should change based on how spread the population is. 

Similar to the analysis in Sec.~\ref{sec:paramAnalysis_volume}, we evaluated MGP-BBBC by setting (at each iteration) the kernel bandwidth $h$ based on its ratio with the maximum Euclidean distance in the population, such that each function is tested on the same range of values -- i.e., the bandwidth is tuned based on the problem to solve. We set ten equispaced ratio values, from $80$ to $3$, and applied the value of bandwidth retrieved with Eqn.~(\ref{eq:bandwidth_fromEucDistRatio}). These ratio values were chosen empirically based on the values of Tables S.I--S.VI in the range of values having non-significant differences against the best-retrieved PR. This analysis produced 400 observations: 4 values for population size (50, 100, 500, and 1000), 10 values for ratio, and 20 functions in the CEC'2013 benchmark set. 

\begin{equation}
    \label{eq:bandwidth_fromEucDistRatio}
    \begin{aligned}
        h = \frac{\max_{i,j} \| \mathbf{x}_i - \mathbf{x}_j \|_2}{\text{ratio}}
    \end{aligned}
\end{equation}

To observe how the values of the parameters affect the results of MGP-BBBC, we compared the best result for each function against each combination with the Wilcoxon rank-sum test at the significance level $\alpha = 0.05$ (results are reported in the supplementary material under Tables S.XXXVI and S.XXXVII for accuracy $\varepsilon = 1.0E-03$, S.XXXIX and S.XL for $\varepsilon = 1.0E-04$, and S.XLII and S.XLIII for $\varepsilon = 1.0E-05$). The tables indicate the number of settings that performed significantly worse ($-$) or non-significantly different ($\approx$) than the best setting in terms of both population size (rows) and the ratio between search space volume and bandwidth region volume (columns). The trend for each problem is graphically depicted in Fig.~\ref{fig:pr_values_r2} for accuracy $\varepsilon = 1.0E-04$ only. 

We adopted Friedman's test with the Bonferroni-Dunn procedure~\cite{demvsar2006statistical} to test the robustness of MGP-BBBC on the 780 combinations between population size and ratio (results are reported in the supplementary material under Table S.XXXVIII for accuracy $\varepsilon = 1.0E-03$, S.XLI for $\varepsilon = 1.0E-04$, and S.XLIV for $\varepsilon = 1.0E-05$). 
The p-value shows the result of the Bonferroni-Dunn procedure; ``Y'' indicates that there is a significant difference (at the significance level $\alpha = 0.05$) between the corresponding pairwise settings, whereas ``N'' indicates no significant difference.
We report obtaining the same results as in the analysis with the search space volume of Sec.~\ref{sec:paramAnalysis_volume}: 524 (67\%) combinations being non-significantly different against 256 (33\%) combinations that are significantly different -- the same outcome is observable for each accuracy level  $\varepsilon=1.0E-03$, $\varepsilon=1.0E-04$, and $\varepsilon=1.0E-05$. Most of the significant differences were found for population size $n=500$ and $n=1000$ and for small versus large bandwidth values. These results are in line with the plots in Fig.~\ref{fig:pr_values_r2}, which show more consistency than the first experimental results in Fig.~\ref{fig:pr_values} and support the hypothesis of relating the clustering kernel bandwidth with the maximum Euclidean distance in the population for setting parameters independently from the problem to be solved. However, with respect to the settings with the search space volume (Sec.~\ref{sec:paramAnalysis_volume}, Fig.~\ref{fig:pr_values_r}), it seems more consistent to choose the value of the bandwidth based on the search space volume, although no statistical difference has been found between the two methods.

\section{Conclusion} \label{sec:conclusion}

In this work, we propose the MGP-BBBC algorithm for MMOPs. 
MGP-BBBC is the multimodal extension of the BBBC algorithm~\cite{erol2006new, gencc2010big}, and an improved version of k-BBBC~\cite{yenin2023multi} based on clustering. 
MGP-BBBC stores the elites in an archive and applies a survival stage to the archive and the newly generated offspring: both populations are filtered separately to remove elite individuals too close to each other, ensuring that elites around sharp peaks are not lost in favor of elites around wide peaks. Then, MGP-BBBC produces a new archive of elites, on which the big-crunch operator is applied to retrieve centers of mass through clustering: this operation promotes isolated individuals by assigning them more offspring than individuals having high niche count in their cluster. Lastly, the centers of mass go through the big-bang operator, which produces new offspring by dynamically balancing exploration and exploitation during different generations. This operation allows MGP-BBBC to correctly converge to the peaks with the desired level of accuracy.

We conducted experiments on twenty multimodal test functions from the CEC'2013 benchmark set. The results show that the overall performance of MGP-BBBC is better than eleven out of thirteen state-of-the-art multimodal optimization algorithms and performs competitively against the other two. 

One of the biggest drawbacks of MGP-BBBC is to rely on a clustering procedure that depends on a threshold (kernel bandwidth) -- i.e., how close two points should be to belong to the same cluster. When different peaks have different sizes of attraction regions (e.g., like in the Vincent function), this becomes a problem: small bandwidths will generate extra clusters around large peaks, whereas large bandwidths might miss small peaks. In the future, we will extend MGP-BBBC by testing other clustering methods that could potentially address this problem, and introduce taboo regions where offspring must not be generated~\cite{ahrari2021static} to reduce the density of clusters. Additionally, we will test MGP-BBBC on constrained MMOPs and use it in real-world applications, such as ﬁnding multiple configurations for soft growing robots to solve a specific task with alternative optimal designs~\cite{stroppa2024design}.

\section*{Acknowledgements}

The authors would like to thank Dr. Ali Ahrari for providing relevant insights on algorithm comparisons. We thank Ozan Nurcan, Emir Ozen, Erk Demirel, and Ozan Kutlar for helping us run experiments. Lastly, we would like to thank Emre Ozel for allowing us to use 121 computers in the Kadir Has University's Computer Labs for parallel computation: it took us only three weeks instead of three years.

\section*{Funding} 
This work is funded by TUBİTAK within the scope of the 2232-B International Fellowship for Early Stage Researchers Program number 121C145.

\section*{Competing interests}
The authors have no relevant financial or non-financial interests to disclose.

\section*{Authors' contributions}

Fabio Stroppa: development, funding acquisition, literature, validation, writing, revision.
Ahmet Astar: experiments and data collection.

\section*{Ethics approval}
This is an observational study. The Institutional Review Board of Kadir Has University has confirmed that no ethical approval is required.

\section*{Consent to participate}
Not required.

\section*{Consent for publication}
No individual person’s data in any form is disclosed in the paper.

\section*{Code or data availability} 

The code is publicly available on MathWorks File Exchange: \url{https://www.mathworks.com/matlabcentral/fileexchange/171184}.

The supplementary material can be downloaded at: \url{https://www.mediafire.com/file/dq9db2b4la40qb9/MGPBBBC_supplementary_material.pdf/file}.

 \bibliographystyle{elsarticle-num} 
 \bibliography{references}

\end{document}